\documentclass[english,british,twocolumn]{svjour3}
\usepackage[T1]{fontenc}
\usepackage[latin9]{inputenc}
\usepackage{xcolor}
\usepackage{array}
\usepackage{verbatim}
\usepackage{rotating}
\usepackage{units}
\usepackage{textcomp}
\usepackage{bbding}
\usepackage{url}
\usepackage{multirow}
\usepackage{amsmath}
\usepackage{graphicx}
\PassOptionsToPackage{normalem}{ulem}
\usepackage{ulem}

\makeatletter

\providecommand{\tabularnewline}{\\}

\RequirePackage{fix-cm}

\smartqed  

\usepackage{times}
\usepackage{epsfig}
\usepackage{graphicx}
\usepackage{amsmath}
\usepackage{amssymb}

\usepackage{placeins}

\usepackage{color}
\usepackage{colortbl}
\usepackage{rotating}
\definecolor{lightgray}{gray}{0.6}
\definecolor{verylightgray}{gray}{0.9}

\usepackage{caption} 
\usepackage{hyphenat}
\usepackage[british,english]{babel}

\usepackage[pagebackref=true,breaklinks=true,letterpaper=true,colorlinks,bookmarks=false]{hyperref}

\@ifundefined{showcaptionsetup}{}{%
 \PassOptionsToPackage{caption=false}{subfig}}
\usepackage{subfig}
\makeatother

\usepackage{babel}
\begin{document}
\makeatletter  
\renewcommand{\paragraph}{%
\@startsection{paragraph}{4}%
 {\z@}{0.5ex \@plus 1ex \@minus .2ex}{-0.5em}%
  {\normalfont \normalsize \bfseries}%
} 

\let\originalparagraph\paragraph 
\renewcommand{\paragraph}[2][.]{\originalparagraph{#2#1}}

\title{Lucid Data Dreaming for Video Object Segmentation}
\selectlanguage{english}%

\author{Anna Khoreva \and Rodrigo Benenson \and Eddy Ilg \and Thomas Brox
\and Bernt Schiele}

\authorrunning{A. Khoreva, R. Benenson, E. Ilg, T. Brox, B. Schiele}
\institute{Anna Khoreva \at Max Planck Institute for Informatics, Germany\\ \email{khoreva@mpi-inf.mpg.de} 
	\and Rodrigo Benenson \at Google\\ \email{benenson@google.com}
\and Eddy Ilg \at University of Freiburg, Germany \\
 \email{ilg@cs.uni-freiburg.com} 
 \and Thomas
Brox \at  University of Freiburg, Germany \\ \email{brox@cs.uni-freiburg.com} 
\and Bernt Schiele \at Max Planck Institute for Informatics, Germany\\
\email{schiele@mpi-inf.mpg.de}
}

\date{Received: date / Accepted: date}
\maketitle
\selectlanguage{british}%
\begin{abstract}
Convolutional networks reach top quality in pixel-level video object segmentation
but require a large amount of training data (\textcolor{black}{$1k\negmedspace\sim\negmedspace100k$})
to deliver such results. We propose a new training strategy which
achieves state-of-the-art results across three evaluation datasets
while using \textcolor{black}{$20\times\negmedspace\sim\negmedspace1000\times$} less
annotated data than competing methods. Our approach is suitable for
both single and multiple object segmentation.

Instead of using large training sets hoping to generalize across domains,
we generate in-domain training data using the provided annotation
on the first frame of each video to synthesize (``lucid dream''\footnote{In a lucid dream the sleeper is aware that he or she is dreaming and
is sometimes able to control the course of the dream.}) plausible future video frames. In-domain per-video training data
allows us to train high quality appearance- and motion-based models,
as well as tune the post-processing stage. This approach allows to
reach competitive results even when training from only a single annotated
frame, without ImageNet pre-training. Our results indicate that using
a larger training set is not automatically better, and that for the
video object segmentation task a smaller training set that is closer to the target
domain is more effective. This changes the mindset regarding how many
training samples and general ``objectness'' knowledge are required
for the video object segmentation task.
\end{abstract}
\vspace{-1em}

\section{Introduction}

\begin{figure}
\begin{centering}
\includegraphics[width=0.97\columnwidth]{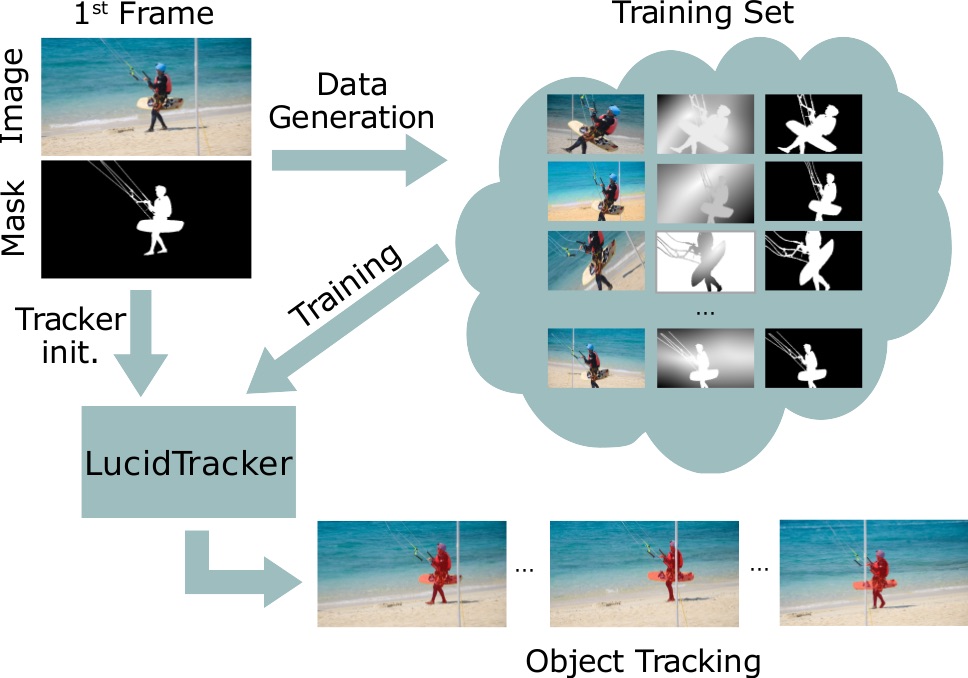}
\par\end{centering}
\caption{\label{fig:Teaser}Starting from scarce annotations we synthesize
in-domain data to train a specialized pixel-level video object segmenter for
each dataset or even each video sequence.}
\vspace{-1em}
\end{figure}

In the last years the field of localizing objects in videos has transitioned
from bounding box tracking \cite{Kristan2015Iccvw,Hadfield14d,Hadfield16d}
to pixel-level segmentation \cite{Li2013Iccv,Prest2012Cvpr,Perazzi2016Cvpr,Vojir2017TechreportVOT16Seg}.
Given a first frame labelled with the foreground object masks, one
aims to find the corresponding object pixels in future frames. Segmenting
objects at the pixel level enables a finer understanding of videos
and is helpful for tasks such as video editing, rotoscoping, and summarisation.

Top performing results are currently obtained using convolutional
networks (convnets) \cite{Jampani2016Arxiv,Caelles2017Cvpr,Khoreva2017CvprMaskTrack,Bertinetto2016Arxiv,Held2016Eccv,Nam2016Cvpr}.
Like most deep learning techniques, convnets for video object segmentation benefit from large amounts of training data. Current state-of-the-art
methods rely, for instance, on pixel accurate foreground/background
annotations of $\sim\negmedspace2k$ video frames \cite{Jampani2016Arxiv,Caelles2017Cvpr}
, $\sim\negmedspace10k$ images \cite{Khoreva2017CvprMaskTrack},
\textcolor{black}{or even more than $100k$ annotated samples for training \cite{Voigtlaender2017OnlineAO}}.
Labelling \textcolor{black}{images and videos} at the pixel level is a laborious task (compared
e.g. to drawing bounding boxes for detection), and creating a large
training set requires significant annotation effort.

In this work we aim to reduce the necessity for such large volumes
of training data. It is traditionally assumed that convnets require
large training sets to perform best. We show that for video object
segmentation having a larger training set is not automatically better
and that improved results can be obtained by using \textcolor{black}{$20\times\negmedspace\sim\negmedspace1000\times$
less training data than previous approaches \cite{Caelles2017Cvpr,Khoreva2017CvprMaskTrack,Voigtlaender2017OnlineAO}}.
The main insight of our work is that for video object segmentation
using few training frames ($1\negmedspace\sim\negmedspace100$) in
the target domain is more useful than using large training volumes
across domains \textcolor{black}{($1k\negmedspace\sim\negmedspace100k$)}. 

To ensure a sufficient amount of training data close to the target
domain, we develop a new technique for synthesizing training data
particularly tailored for the pixel-level video object segmentation scenario. We call this
data generation strategy ``\textit{lucid dreaming}'', where the
first frame and its annotation mask are used to generate plausible
future frames of the videos. The goal is to produce a large training
set of reasonably realistic images which capture the expected appearance
variations in future video frames, and thus is, by design, close to
the target domain.

Our approach is suitable for both single and multiple object segmentation in videos.
Enabled by the proposed data generation strategy and the efficient
use of optical flow, we are able to achieve high quality results while
using only $\sim\negmedspace100$ individual annotated training frames.
Moreover, in the extreme case with only a single annotated frame and
zero pre-training (i.e. without ImageNet pre-training), we still obtain
competitive video object segmentation results.

In summary, our contributions are the following:\\
1. We propose ``lucid data dreaming'', an automated approach to
synthesize training data for the convnet-based pixel-level video object segmentation that leads to top results for both single and multiple object
segmentation\footnote{Lucid data dreaming synthesis implementation is available at \url{https://www.mpi-inf.mpg.de/lucid-data-dreaming}.
}.\\
2. We conduct an extensive analysis to explore the factors contributing
to our good results.\\
3. We show that training a convnet for video object segmentation can be done
with only few annotated frames. We hope these results will affect
the trend towards even larger training sets, and popularize the design
of video segmentation convnets with lighter training needs.

With the results for multiple object segmentation we took the second place in the 2017 DAVIS Challenge on Video Object Segmentation \cite{Pont-Tuset_arXiv_2017}. 
A summary of the proposed approach was provided online \cite{DAVIS2017-2nd}. This paper significantly extends \cite{DAVIS2017-2nd} with in-depth discussions on the
method, more details of the formulation, its implementation, and its variants for single and multiple object segmentation in videos. 
It also offers a detailed ablation study and an error analysis as well as explores the impact of varying number of annotated training samples on the video segmentation quality.

\section{\label{subsec:Related-work}Related work}

\paragraph{Box tracking}

Classic work on video object tracking focused on bounding box tracking.
Many of the insights from these works have been re-used for video object segmentation.
Traditional box tracking smoothly updates across time a
linear model over hand-crafted features \cite{Henriques2012Eccv,breitenstein2009robust,Hadfield14d}.
Since then, convnets have been used as improved features \cite{Danelljan2015Iccvw,Ma-ICCV-2015,wang2015visual},
and eventually to drive the tracking itself \cite{Held2016Eccv,Bertinetto2016Arxiv,tao2016siamese,nam2016modeling,Nam2016Cvpr}.
Contrary to traditional box trackers (e.g. \cite{Henriques2012Eccv}), convnet-based approaches need additional data for pre-training and
learning the task.

\paragraph{Video object segmentation}

In this paper we focus on generating a foreground versus background
pixel-wise object labelling for each video frame starting from a first
manually annotated frame. Multiple strategies have been proposed to
solve this task.

\emph{Box-to-segment:} First a box-level track is built, and a space-time
grabcut-like approach is used to generate per frame segments {\cite{Xiao2016Cvpr}.}

\emph{Video saliency:} this group of methods extracts
the main foreground object pixel-level space-time tube. Both hand-crafted
models {\cite{Faktor2014Bmvc,Papazoglou2013Iccv}} or trained
convnets \textcolor{black}{\cite{TokmakovAS17,Jain2017ArxivFusionSeg,Song_2018_ECCV}}
have been considered. Because these methods ignore the first frame
annotation, they fail in videos where multiple salient objects move
(e.g. flock of penguins).

\emph{Space-time proposals:} these methods partition the space-time vo\-lume,
 and then the tube overlapping most with the first frame mask annotation
 is selected as tracking output \cite{Grundmann2010Cvpr,Perazzi2015Iccv,Chang2013tsp}.

\emph{Mask propagation}: Appearance similarity and motion smoothness
across time is used to propagate the first frame annotation across
the video \cite{Maerki2016Cvpr,Wang2017ArxivSTV,Tsai2016Cvpr}. These
methods usually leverage optical flow and long term trajectories.

\emph{Convnets:} following the trend in box tracking, recently
convnets have been proposed for video object segmentation. {\cite{Caelles2017Cvpr}}
trains a generic object saliency network, and fine-tunes it per-video
(using the first frame annotation) to make the output sensitive to
the specific object of interest. {\cite{Khoreva2017CvprMaskTrack}}
uses a similar strategy, but also feeds the mask from the previous
frame as guidance for the saliency network. 
\textcolor{black}{\cite{Voigtlaender2017OnlineAO} incorporates online adaptation of the network using the predictions from previous frames. \cite{Chandra2018DeepSR} extends the Gaussian-CRF approach to videos by exploiting spatio-temporal connections for pairwise terms and relying on unary terms from \cite{Voigtlaender2017OnlineAO}.}
Finally {\cite{Jampani2016Arxiv}}
mixes convnets with ideas of bilateral filtering. Our approach also
builds upon convnets.\\
What makes convnets particularly suitable for the task, is that they
can learn what are the common statistics of appearance and motion
patterns of objects, as well as what makes them distinctive from the
background, and exploit this knowledge when segmenting a particular
object. This aspect gives convnets an edge over traditional techniques
based on low-level hand-crafted features.

Our network architecture is similar to \cite{Caelles2017Cvpr,Khoreva2017CvprMaskTrack}.
Other than implementation details, there are three differentiating
factors. One, we use a different strategy for training: \textcolor{black}{\cite{Caelles2017Cvpr,Jampani2016Arxiv,Chandra2018DeepSR,Voigtlaender2017OnlineAO}}
rely on consecutive video training frames and \cite{Khoreva2017CvprMaskTrack}
uses an external saliency dataset, while our approach focuses on using
the first frame annotations provided with each targeted video benchmark
without relying on external annotations. Two, our approach exploits
optical flow better than these previous methods. Three, we describe
an extension to seamlessly handle segmentation of multiple objects.

\paragraph{Interactive video segmentation}

Interactive segmentation {\cite{Nagaraja2015Iccv,Jain2016Hcomp,SpinaF16,Wang2014Cviu}}
considers more diverse user inputs (e.g. strokes), and requires interactive
processing speed rather than providing maximal quality. Albeit our
technique can be adapted for varied inputs, we focus on maximizing
quality for the non-interactive case with no-additional hints along
the video.

\paragraph{Semantic labelling}

Like other convnets in this space \cite{Jampani2016Arxiv,Caelles2017Cvpr,Khoreva2017CvprMaskTrack},
our architecture builds upon the insights from the semantic labelling
networks \cite{ZhaoSQWJ16,LinMS016,WuSH16e,PixelNet}. Because of
this, the flurry of recent developments should directly translate
into better video object segmentation results. For the sake of comparison with previous
work, we build upon the well established VGG DeepLab architecture
\cite{Chen2016ArxivDeeplabv2}.

\paragraph{Synthetic data}

Like our approach, previous works have also explored synthesizing
training data. Synthetic renderings \cite{Mayer2016Cvpr}, video game
environment \cite{Richter_2016_ECCV}, mix-synthetic and real images
\cite{Varol0MMBLS17,chen2016synthesizing,Dosovitskiy15} have shown
promise, but require task-appropriate 3d models. Compositing real
world images pro{-}vides more realistic results, and has shown promise
for object detection \cite{georgakis2017synthesizing,tang2013learning},
text localization \cite{gupta2016synthetic} and pose estimation \cite{pishchulin2012articulated}. 

The closest work to ours is \cite{ParkR15}, which also generates
video-specific training data using the first frame annotations. They
use human skeleton annotations to improve pose estimation, while we
employ pixel-level mask annotations to improve video object segmentation.
\begin{figure*}[t]
\setlength{\tabcolsep}{0.1em} 
\renewcommand{\arraystretch}{0.1}

\hfill{}\hspace*{-0.5em}%
\begin{tabular}{c|cccc}
 &  &  &  & \tabularnewline
\begin{turn}{90}
\end{turn} &  & \includegraphics[width=0.32\textwidth]{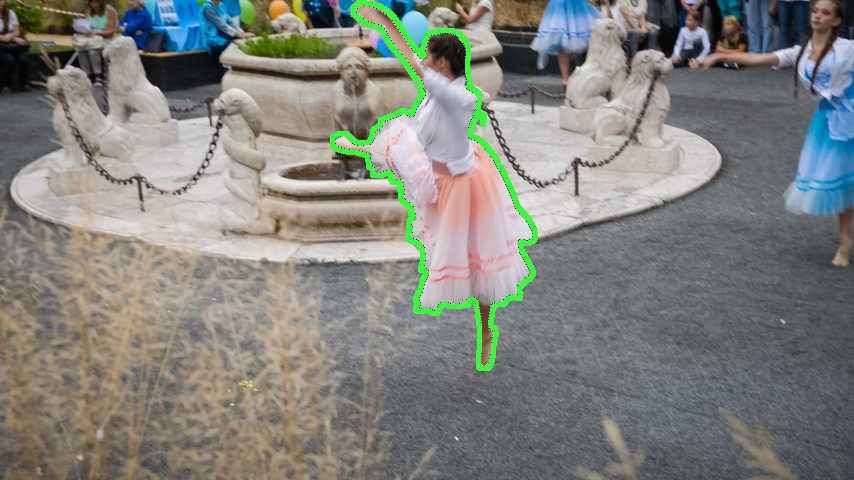} & \includegraphics[width=0.32\textwidth]{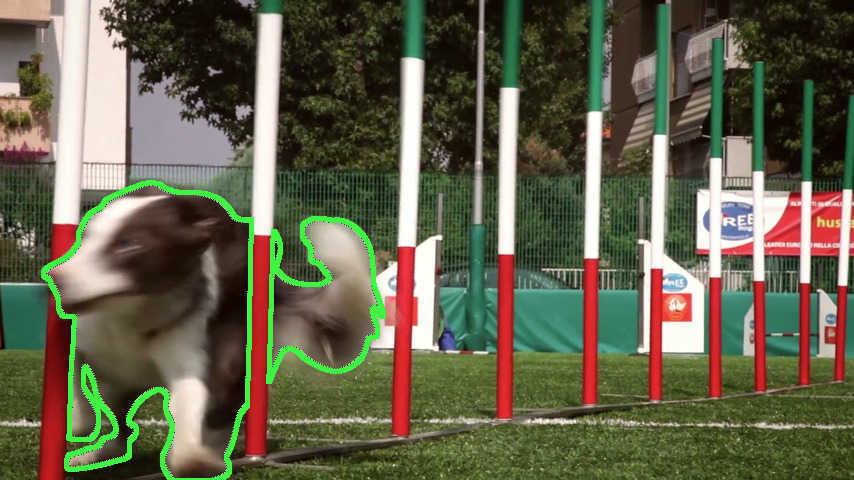} & \includegraphics[width=0.32\textwidth]{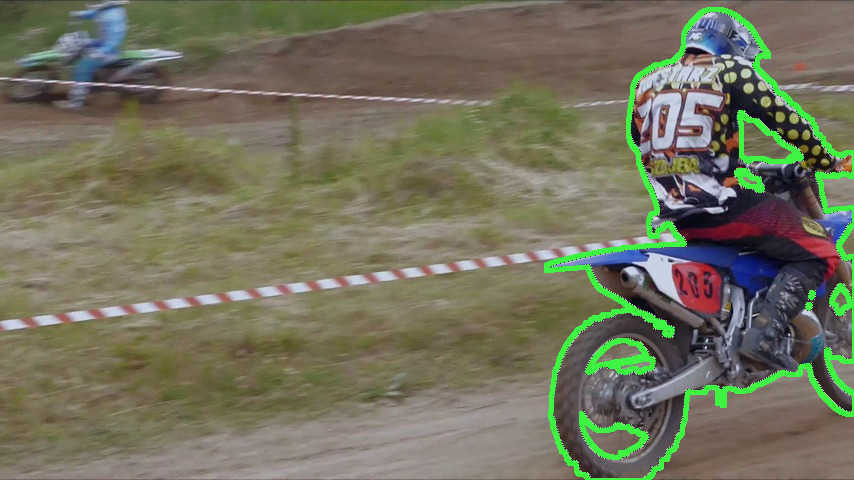}\tabularnewline
 &  &  &  & \tabularnewline
 &  & \multicolumn{3}{c}{$\mathcal{I}_{t}$}\tabularnewline
 &  &  &  & \tabularnewline
\begin{turn}{90}
{\footnotesize{}\hspace{1.5em}}\textcolor{black}{\small{}Inputs}
\end{turn} &  & \includegraphics[width=0.32\textwidth]{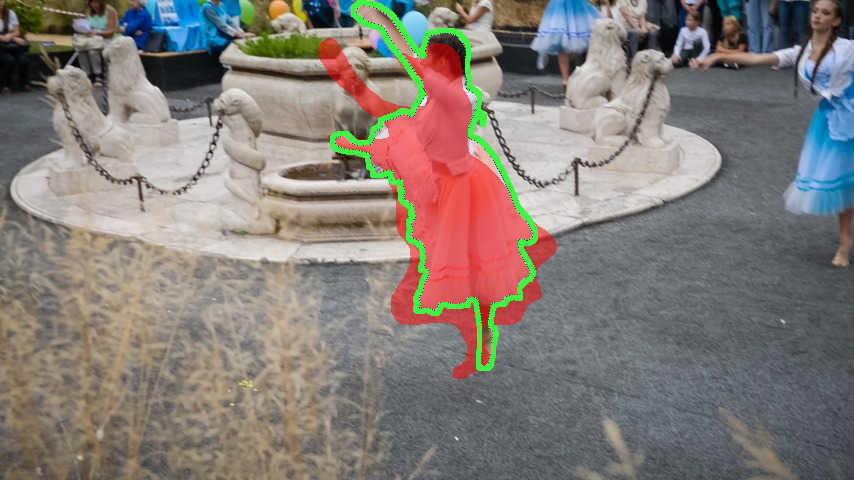} & \includegraphics[width=0.32\textwidth]{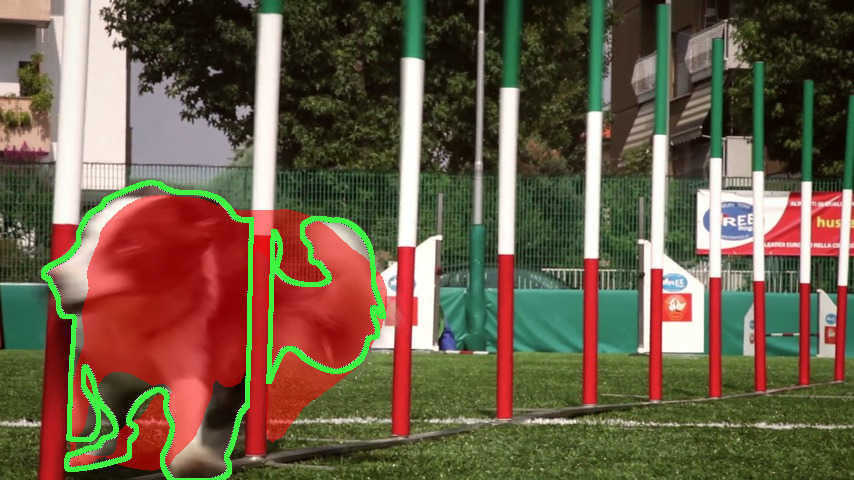} & \includegraphics[width=0.32\textwidth]{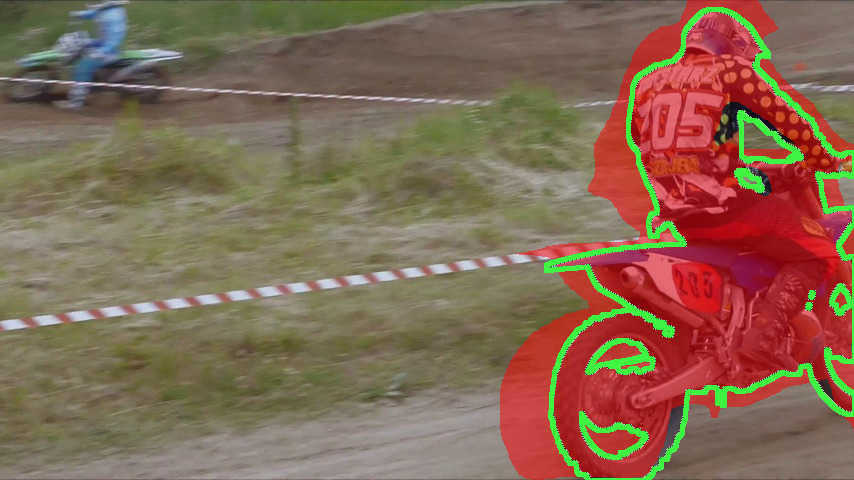}\tabularnewline
 &  &  &  & \tabularnewline
 &  & \multicolumn{3}{c}{$M_{t-1}$ (shown over $\mathcal{I}_{t}$)}\tabularnewline
 &  &  &  & \tabularnewline
 &  & \includegraphics[width=0.32\textwidth]{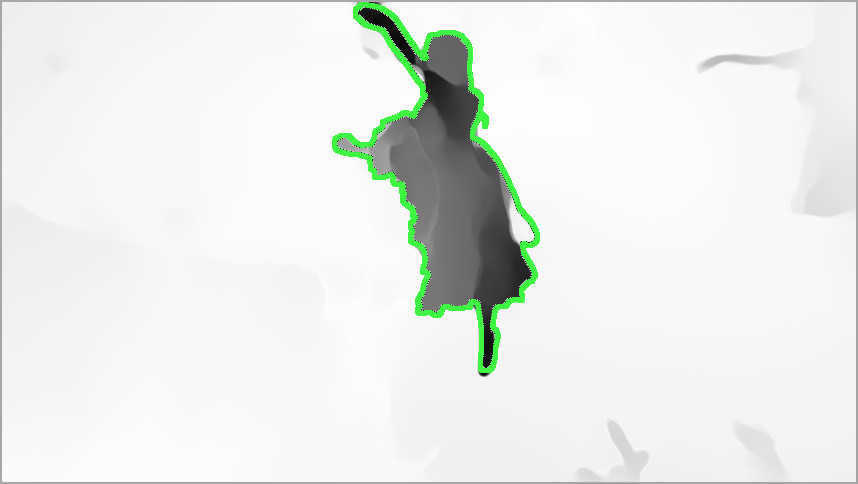} & \includegraphics[width=0.32\textwidth]{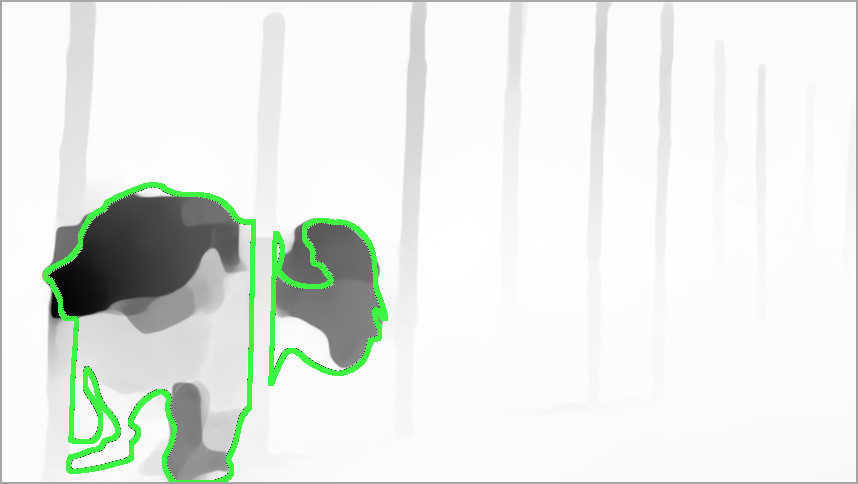} & \includegraphics[width=0.32\textwidth]{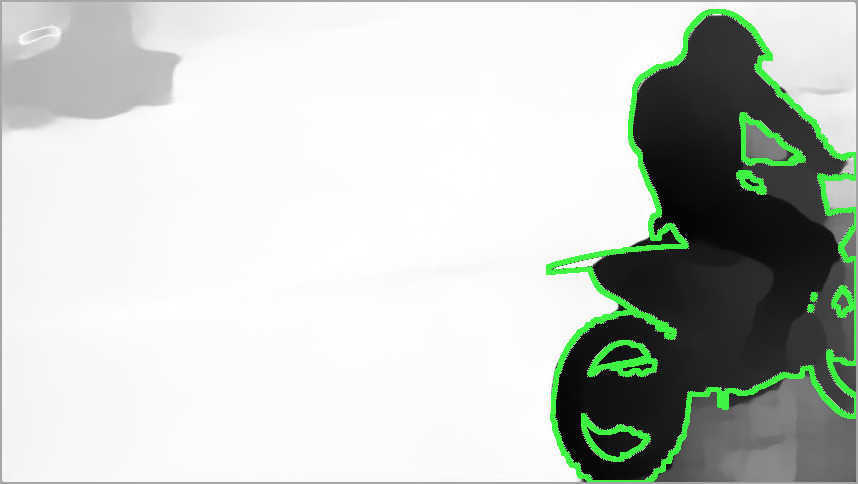}\tabularnewline
 &  &  &  & \tabularnewline
\multicolumn{1}{c}{} &  &  & $\left\Vert \mathcal{F}_{t}\right\Vert $ & \tabularnewline
\multicolumn{1}{c}{} &  &  & \smallskip{}
 & \tabularnewline
 &  &  &  & \tabularnewline
\begin{turn}{90}
{\footnotesize{}\hspace{1em}}\textcolor{black}{\small{}Output}
\end{turn} &  & \includegraphics[width=0.32\textwidth]{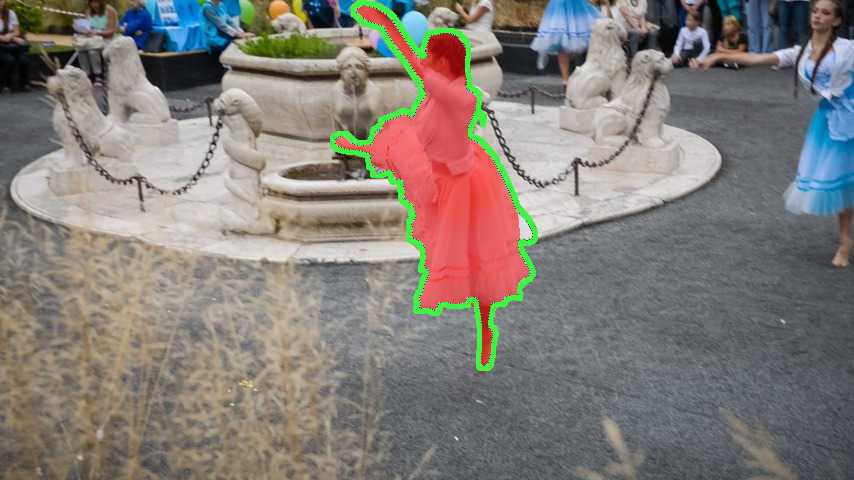} & \includegraphics[width=0.32\textwidth]{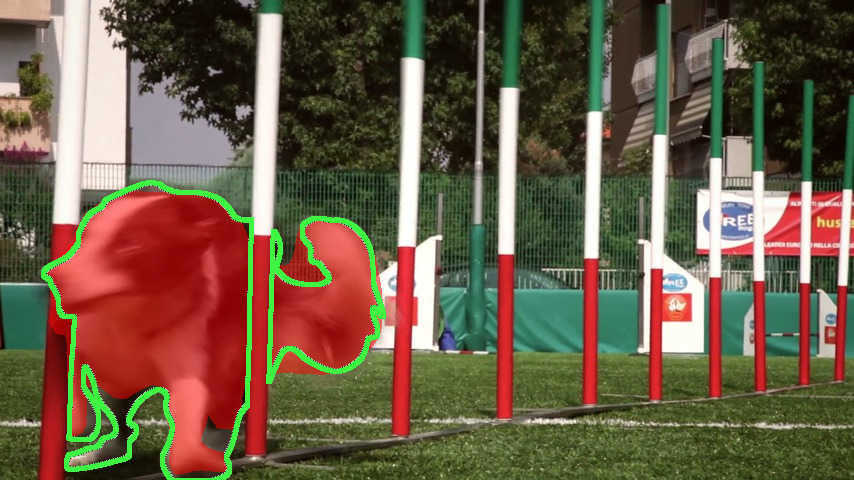} & \includegraphics[width=0.32\textwidth]{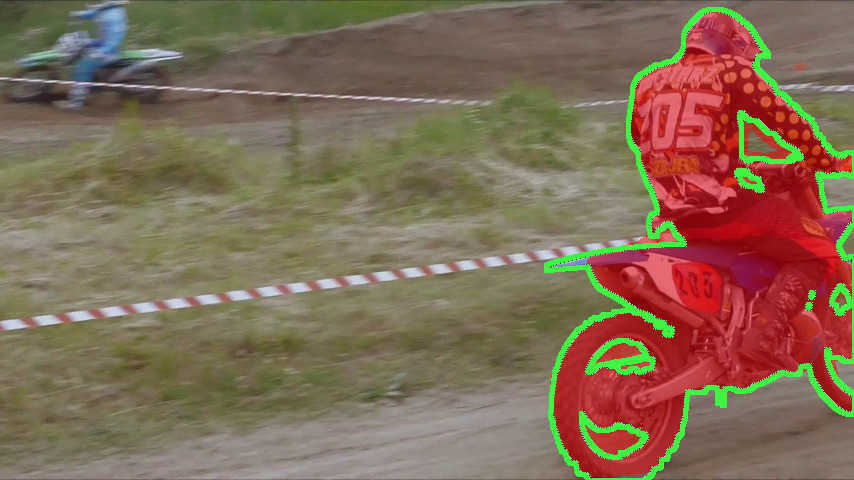}\tabularnewline
 &  &  &  & \tabularnewline
\multicolumn{1}{c}{} &  &  & $M_{t}$ & \tabularnewline
\end{tabular}\hfill{}

\caption{\label{fig:DataFlow}Data flow examples. $\mathcal{I}_{t},\,\left\Vert \mathcal{F}_{t}\right\Vert,\,M_{t-1}$
are the inputs, $M_{t}$ is the resulting output. Green boundaries
outline the ground truth segments. Red overlay indicates $M_{t-1},\,M_{t}$.}
\end{figure*}

\section{\label{sec:LucidTracker-architecture}LucidTracker }

Section \ref{subsec:Architecture} describes the network architecture
used, and how RGB and optical flow information are fused to predict
the next frame segmentation mask. Section \ref{subsec:Training-modalities}
discusses different training modalities employed with the proposed
video object segmentation system. In Section \ref{sec:Lucid-data-dreaming}
we discuss the training data generation, and sections \ref{sec:Single-object-results}/\ref{sec:Multiple-object-results}
report results for single/multiple object segmentation in videos.

\subsection{\label{subsec:Architecture}Architecture}

\paragraph{Approach}

We model video object segmentation as a mask refinement
task (mask: binary foreground/ background labelling of the image)
based on appearance and motion cues. From frame $t-1$ to frame $t$
the estimated mask $M_{t-1}$ is propagated to frame $t$, and the
new mask $M_{t}$ is computed as a function of the previous mask,
the new image $\mathcal{I}_{t}$, and the optical flow $\mathcal{F}_{t}$,
i.e. $M_{t}=f\left(\mathcal{I}_{t},\,\mathcal{F}_{t},\,M_{t-1}\right)$.
Since objects have a tendency to move smoothly through space in time,
there are little changes from frame to frame and mask $M_{t-1}$ can
be seen as a rough estimate of $M_{t}$. Thus we require our trained
convnet to learn to refine rough masks into accurate masks. Fusing
the complementary image $\mathcal{I}_{t}$ and motion flow $\mathcal{F}_{t}$
enables to exploits the information inherent to video and enables
the model to segment well both static and moving objects.

Note that this approach is incremental, does a single forward pass
over the video, and keeps no explicit model of the object appearance
at frame $t$. In some experiments we adapt the model $f$ per video,
using the annotated first frame $\mathcal{I}_{0},\,M_{0}$. However,
in contrast to traditional techniques \cite{Henriques2012Eccv}, this
model is not updated while we process the video frames, thus the only
state evolving along the video is the mask $M_{t-1}$ itself.

\paragraph{First frame}

In the video object segmentation task of our interest the mask for the first frame $M_{0}$
is given. This is the standard protocol of the benchmarks considered
in sections \ref{sec:Single-object-results} \& \ref{sec:Multiple-object-results}.
If only a bounding box is available on the first frame, then the mask
could be estimated using grabcut-like techniques \cite{Rother2004SiggraphGrabcut,Tang2016Eccv}.

\paragraph{RGB image $\mathcal{I}$}

Typically a semantic labeller generates pixel-wise labels based on
the input image (e.g. $M=g\left(\mathcal{I}\right)$). We use an augmented
semantic labeller with an input layer modified to accept 4 channels
(RGB + previous mask) so as to generate outputs based on the previous
mask estimate, e.g. $M_{t}=f_{\mathcal{I}}\left(\mathcal{I}_{t},\,M_{t-1}\right)$.
Our approach is general and can leverage any existing semantic labelling
architecture. We select the DeepLabv2 architecture with VGG base network
\cite{Chen2016ArxivDeeplabv2}, which is comparable to \cite{Jampani2016Arxiv,Caelles2017Cvpr,Khoreva2017CvprMaskTrack};
FusionSeg {\cite{Jain2017ArxivFusionSeg}} uses ResNet.

\begin{figure}[t]
\begin{centering}
\vspace{0em}
\subfloat[\label{fig:architecture}Two streams architecture, where image $\mathcal{I}_{t}$
and optical flow information $\left\Vert \mathcal{F}_{t}\right\Vert $
are used to update mask $M_{t-1}$ into $M_{t}$. See equation \ref{eq:two-streams}.]{

\includegraphics[width=0.97\columnwidth,height=0.18\textheight]{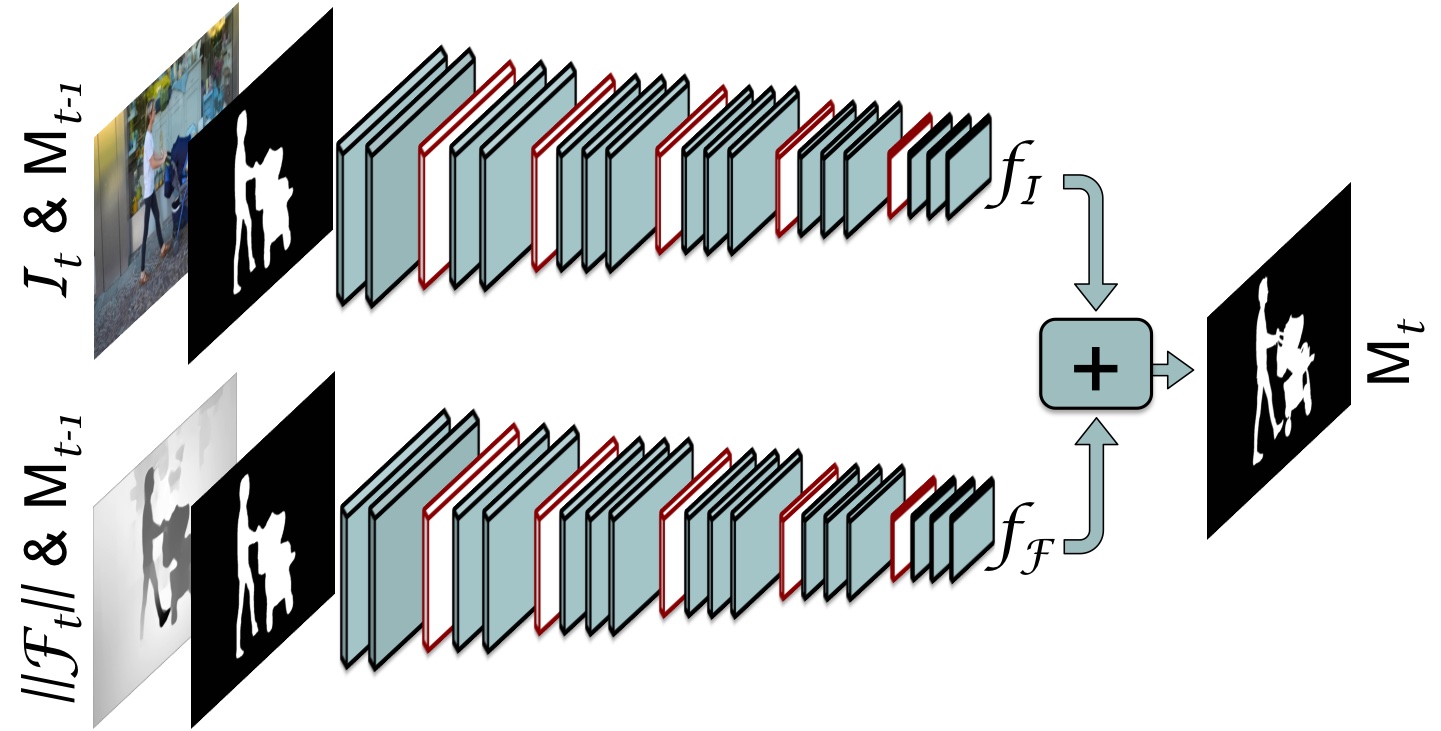}

}
\par\end{centering}
\subfloat[\label{fig:one-stream-architecture}One stream architecture, where
5 input channels: image $\mathcal{I}_{t}$, optical flow information
$\left\Vert \mathcal{F}_{t}\right\Vert $ and mask $M_{t-1}$ are
used to estimate mask $M_{t}$.]{\includegraphics[width=0.97\columnwidth]{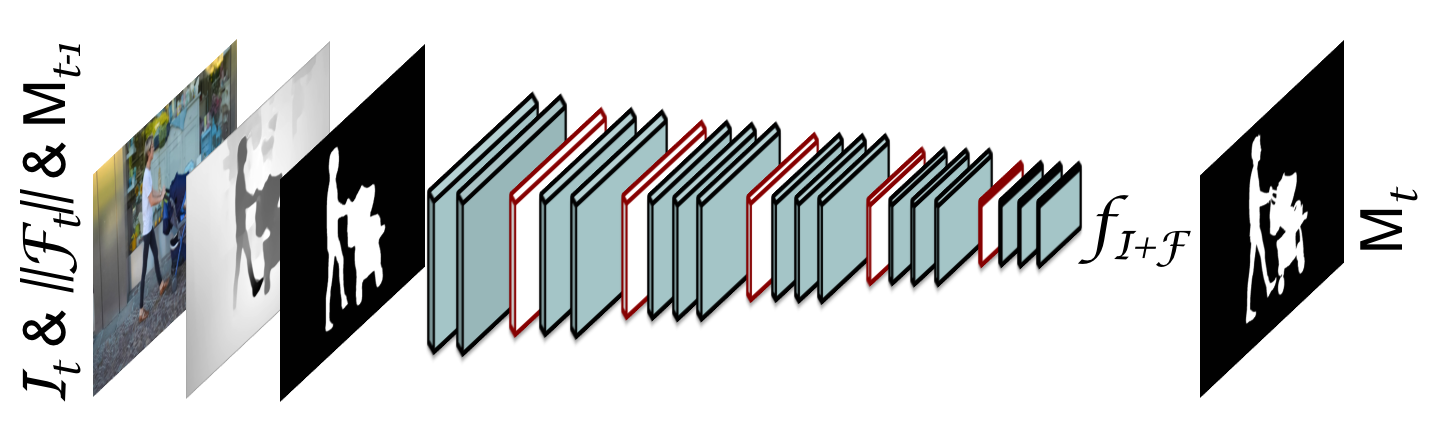}

}

\caption{\label{fig:Two-and-single-stream-architectures}Overview of the proposed
one and two streams architectures. See Section \ref{subsec:Architecture}. }
\vspace{0em}
\end{figure}

\paragraph{Optical flow $\mathcal{F}$}

We use flow in two complementary ways. First, to obtain a better initial
estimate of $M_{t}$ we warp $M_{t-1}$ using the flow $\mathcal{F}_{t}$:
$M_{t}=f_{\mathcal{I}}\left(\mathcal{I}_{t},\,w(M_{t-1},\mathcal{\,F}_{t})\right)$;
we call this \textquotedbl{}mask warping\textquotedbl{}. Second, we
use flow as a direct source of information about the mask $M_{t}$.
As can be seen in Figure \ref{fig:DataFlow}, when the object is moving
relative to background, the flow magnitude $\left\Vert \mathcal{F}_{t}\right\Vert $
provides a very reasonable estimate of the mask $M_{t}$. We thus
consider using a convnet specifically for mask estimation from flow:
$M_{t}=f_{\mathcal{F}}\left(\left\Vert \mathcal{F}_{t}\right\Vert ,\,w(M_{t-1},\mathcal{\,F}_{t})\right)$,
and merge it with the image-only version by naive averaging 
\begin{equation}
M_{t}=0.5\cdot f_{\mathcal{I}}\left(\mathcal{I}_{t},\,\ldots\right)+0.5\cdot f_{\mathcal{F}}\left(\left\Vert \mathcal{F}_{t}\right\Vert ,\,\ldots\right).\label{eq:two-streams}
\end{equation}

We use the state-of-the-art optical flow estimation method FlowNet2.0
\cite{IlgCVPR17}, which itself is a convnet that computes $\mathcal{F}_{t}=h\left(\mathcal{I}_{t-1},\,\mathcal{I}_{t}\right)$
and is trained on synthetic renderings of flying objects \cite{Mayer2016Cvpr}.
For the optical flow magnitude computation we subtract the median
motion for each frame, average the magnitude of the forward and backward
flow and scale the values per-frame to $[0;255]$, bringing it to
the same range as RGB channels.

The loss function is the sum of cross-entropy terms over each pixel
in the output map (all pixels are equally weighted). In our experiments
$f_{\mathcal{I}}$ and $f_{\mathcal{F}}$ are trained independently,
via some of the modalities listed in Section \ref{subsec:Training-modalities}.
Our two streams architecture is illustrated in Figure \ref{fig:architecture}. 

We also explored expanding our network to accept 5 input channels
(RGB + previous mask + flow magnitude) in one stream: $M_{t}=f_{\mathcal{I+F}}\left(\mathcal{I}_{t},\,\left\Vert \mathcal{F}_{t}\right\Vert ,\,w(M_{t-1},\mathcal{\,F}_{t})\right)$,
but did not observe much difference in the performance compared to
naive averaging, see experiments in Section \ref{subsubsec:convnet-arch}.
Our one stream architecture is illustrated in Figure \ref{fig:one-stream-architecture}.
One stream network is more affordable to train and allows to easily
add extra input channels, e.g. providing additionally semantic information
about objects. 

\paragraph{Multiple objects}

The proposed framework can easily be extended to segmenting multiple objects simultaneously.
Instead of having one additional input channel for the previous frame mask
we provide the mask for each object instance in a separate channel, expanding
the network to accept $3+N$ input channels (RGB + $N$ object masks):
$M_{t}=f_{\mathcal{I}}\left(\mathcal{I}_{t},\,w(M_{t-1}^{1},\mathcal{\,F}_{t}),{\scriptstyle \ldots},\,w(M_{t-1}^{N},\mathcal{\,F}_{t})\right)$,
where $N$ is the number of objects annotated on the first frame.

\begin{figure}[t]
\begin{centering}
\vspace{0em}
\includegraphics[width=0.97\columnwidth,height=0.18\textheight]{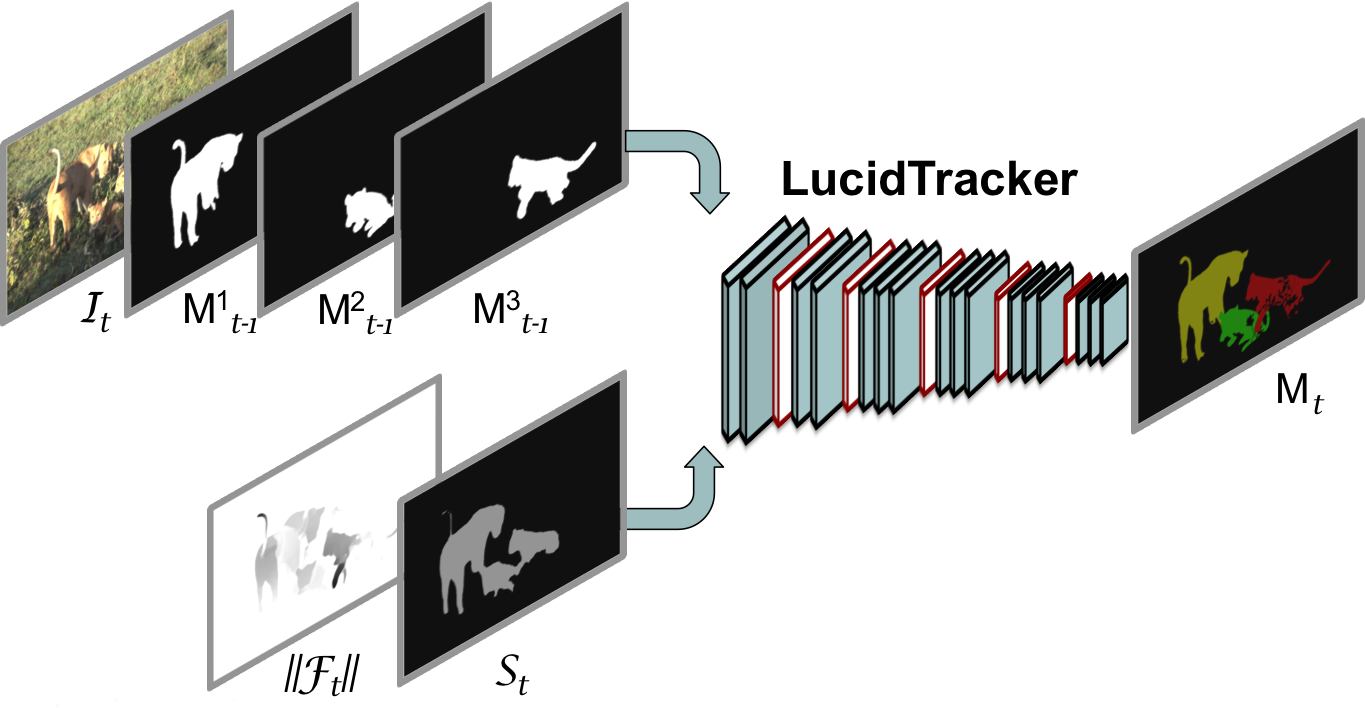}\caption{\label{fig:architecture-mult}Extension of LucidTracker to multiple
objects. The previous frame mask for each object is provided in a
separate channel. We additionally explore using optical flow $\mathcal{F}$
and semantic segmentation $\mathtt{\mathcal{S}}$ as additional inputs.
See Section \ref{subsec:Architecture}. }
\par\end{centering}
\vspace{0em}
\end{figure}

For multiple object segmentation we employ a one-stream architecture
for the experiments, using optical flow $\mathcal{F}$ and semantic
segmentation $\mathtt{\mathcal{S}}$ as additional input channels:\linebreak{}
$M_{t}{\scriptstyle {\scriptstyle =}}f_{\mathcal{I{\scriptscriptstyle +}F{\scriptscriptstyle +}\mathcal{S}}}\left(\mathcal{I}_{t},\,\left\Vert \mathcal{F}_{t}\right\Vert ,\,\mathcal{\mathcal{S}}_{t},\,w(M_{t-1}^{1},\mathcal{\,F}_{t}),{\scriptstyle \ldots},\,w(M_{t-1}^{N},\mathcal{\,F}_{t})\right)$.
This allows to leverage the appearance model with semantic priors
and motion information. See Figure \ref{fig:architecture-mult} for
an illustration. 

\textcolor{black}{The one-stream network is trained with multi-class cross entropy loss and is able to segment multiple objects simultaneously, sharing the feature computation for different instances. This allows to avoid a linear increase of the cost with the number of objects. 
In our preliminary results using a single architecture also pro{-}vides better
results than segmenting multiple objects separately, one at a time;
and avoids the need to design a merging strategy amongst overlapping
tracks.}

\paragraph{Semantic labels $\mathcal{S}$}

To compute the pixel-level semantic labelling $\mathcal{\mathcal{S}}_{t}=h\left(\mathcal{I}_{t}\right)$
we use the state-of-the-art convnet PSPNet \cite{ZhaoSQWJ16}, trained
on Pascal VOC12 \cite{Everingham15}. Pascal VOC12 annotates 20 categories,
yet we want to track any type of objects. $\mathcal{\mathcal{S}}_{t}$
can also provide information about unknown category instances by describing
them as a spatial mixture of known ones (e.g. a sea lion might looks
like a dog torso, and the head of cat). As long as the predictions
are consistent through time, $\mathcal{\mathcal{S}}_{t}$ will provide
a useful cue for segmentation. Note that we only use $\mathcal{\mathcal{S}}_{t}$
for the multi-object segmentation challenge, discussed in Section \ref{sec:Multiple-object-results}.
In the same way as for the optical flow we scale $\mathcal{\mathcal{S}}_{t}$
to bring all the channels to the same range.

We additionally experiment with ensembles of different variants, that
allows to make the system more robust to the challenges inherent in
videos. For our main results on the multiple object segmentation task we
consider an ensemble of four models: $M_{t}{\scriptstyle {\scriptstyle =}}0.25\cdot\left(f_{\mathcal{I{\scriptscriptstyle +}F{\scriptscriptstyle +}\mathcal{S}}}+f_{\mathcal{I{\scriptscriptstyle +}F}}+f_{\mathcal{I{\scriptscriptstyle +}S}}+f_{\mathcal{I}}\right)$,
where we merge the outputs of the models by naive averaging. See Section
\ref{sec:Multiple-object-results} for more details. 

\paragraph{Temporal coherency}
To improve the temporal coherency of the proposed video object segmentation framework we introduce an additional step into the system.
Before providing as input the previous frame mask warped with the optical flow $w(M_{t-1},\mathcal{\,F}_{t})$, we look at frame $t-2$ to remove inconsistencies between the predicted masks $M_{t-1}$ and $M_{t-2}$. 
In particular, we split the mask $M_{t-1}$ into connected components and remove all components from $M_{t-1}$ which do not overlap with $M_{t-2}$. 
This way we remove possibly spurious blobs generated by our model in $M_{t-1}$.
Afterwards we warp the ``pruned'' mask $\widetilde{{M}}_{t-1}$ with the optical flow and use $w(\widetilde{M}_{t-1},\mathcal{\,F}_{t})$ as an input to the network. 
This step is applied only during inference, it mitigates error propagation issues, as well as help generating more temporally coherent results.

\paragraph{Post-processing}

As a final stage of our pipeline, we refine per-frame $t$ the generated
mask $M_{t}$ using DenseCRF \cite{Kraehenbuehl2011Nips}. This adjusts
small image details that the network might not be able to handle.
It is known by practitioners that DenseCRF is quite sensitive to its
parameters and can easily worsen results. We will use our lucid dreams
to handle per-dataset CRF-tuning too, see Section \ref{subsec:Training-modalities}.

We refer to our full $f_{\mathcal{I{\scriptscriptstyle +}F}}$ system
as $\mathtt{LucidTracker}$, and as $\mathtt{LucidTracker^{-}}$ when
no temporal coherency or post-processing steps are used. The usage of $S_{t}$ or model ensemble
will be explicitly stated.

\subsection{\label{subsec:Training-modalities}Training modalities}

Multiple modalities are available to train a tracker. \textbf{Training-free}
approaches (e.g. BVS {\cite{Maerki2016Cvpr}}, SVT {\cite{Wang2017ArxivSTV}})
are fully hand-crafted systems with hand-tuned parameters, and thus
do not require training data. They can be used as-is over different
datasets. Supervised methods can also be trained to generate a \textbf{dataset-agnostic}
model that can be applied over different datasets. Instead of using
a fixed model for all cases, it is also possible to obtain specialized
\textbf{per-dataset} models, either via self-supervision \cite{Wang2015IccvUnsupervised,Pathak2016Arxiv,Yu2016ArxivFlow,Zhu2017ArxivFlow}
or by using the first frame annotation of each video in the dataset
as training/tuning set. Finally, inspired by traditional box tracking
techniques, we also consider adapting the model weights to the specific
video at hand, thus obtaining \textbf{per-video} models. Section \ref{sec:Single-object-results}
reports new results over these four training modalities (training-free,
dataset-agnostic, per-dataset, and per-video). 

Our $\mathtt{LucidTracker}$ obtains best results when first pre-trained
on ImageNet, then trained per-dataset using all data from first frame
annotations together, and finally fine-tuned per-video for each evaluated
sequence. The post-processing DenseCRF stage is automatically tuned
per-dataset. The experimental section \ref{sec:Single-object-results}
details the effect of these training stages. Surprisingly, we can
obtain reasonable performance even when training from only a single
annotated frame (without ImageNet pre-training, i.e. zero pre-training);
this results goes against the intuition that convnets require large
training data to provide good results.

Unless otherwise stated, we fine-tune per-video models relying solely
on the first frame $\mathcal{I}_{0}$ and its annotation $M_{0}$.
This is in contrast to traditional techniques \cite{Henriques2012Eccv,breitenstein2009robust,Hadfield14d}
which would update the appearance model at each frame $\mathcal{I}_{t}$.

\section{\label{sec:Lucid-data-dreaming}Lucid data dreaming}

To train the function $f$ one would think of using ground truth data
for $M_{t-1}$ and $M_{t}$ (like \cite{Bertinetto2016Arxiv,Caelles2017Cvpr,Held2016Eccv}),
however such data is expensive to annotate and rare. {\cite{Caelles2017Cvpr}}
thus trains on a set of $30$ videos ($\sim\negmedspace2k$ frames)
and requires the model to transfer across multiple tests sets. {\cite{Khoreva2017CvprMaskTrack}}
side-steps the need for consecutive frames by generating synthetic
masks $M_{t-1}$ from a saliency dataset of $\sim\negmedspace10k$
images with their corresponding mask $M_{t}$ . We propose a new data
generation strategy to reach better results using only $\sim\negmedspace100$
individual training frames. 

\begin{figure*}
\setlength{\tabcolsep}{0.1em} 
\renewcommand{\arraystretch}{1}

\vspace{0em}
\hfill{}%
\begin{tabular}{cc|cccc}
\multirow{2}{*}{%
\begin{tabular}{c}
\includegraphics[width=0.24\textwidth]{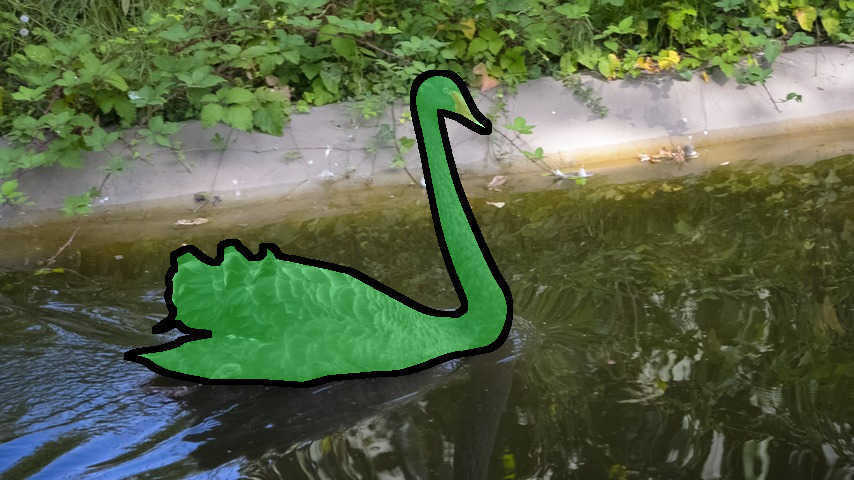}\tabularnewline
blackswan\tabularnewline
\end{tabular}} &  &  & \includegraphics[width=0.24\textwidth]{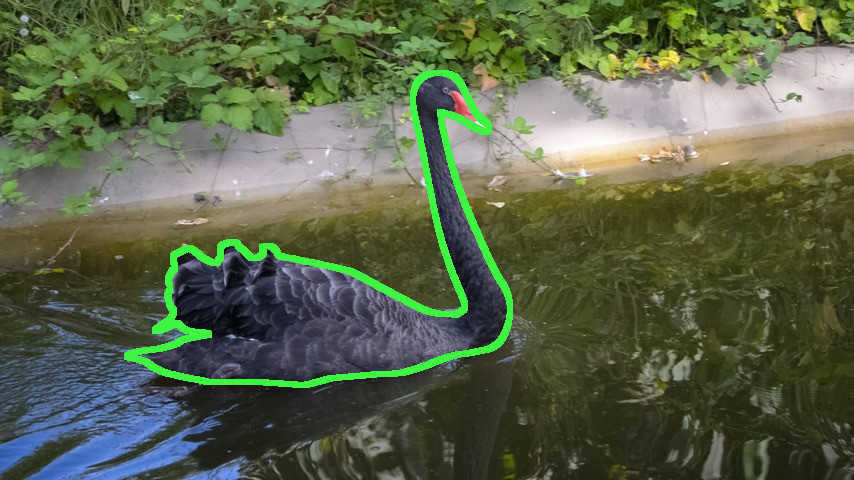} & \includegraphics[width=0.24\textwidth]{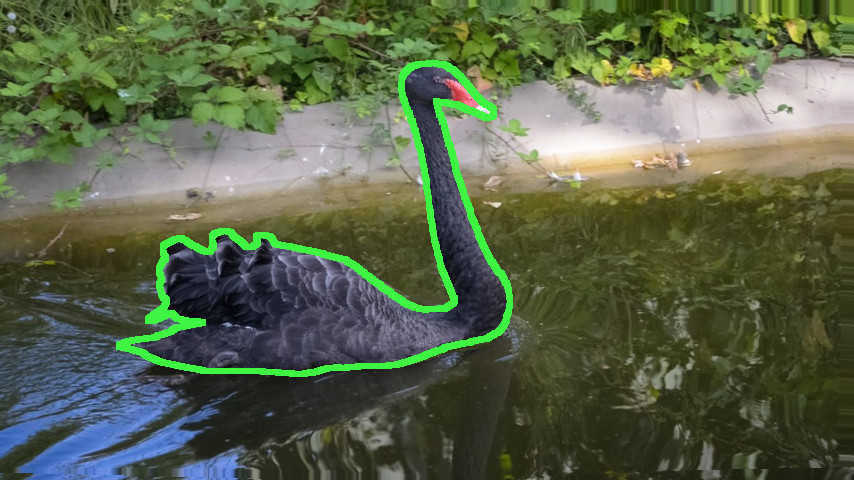} & \includegraphics[width=0.24\textwidth]{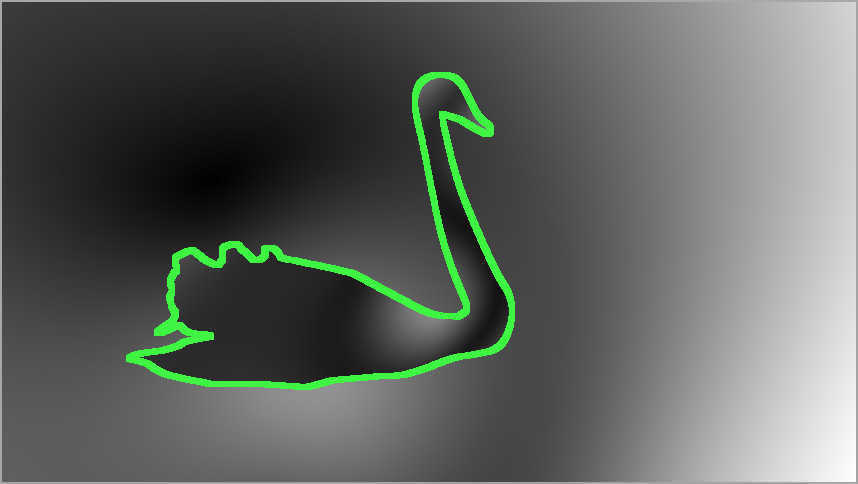}\tabularnewline
 &  &  & \includegraphics[width=0.24\textwidth]{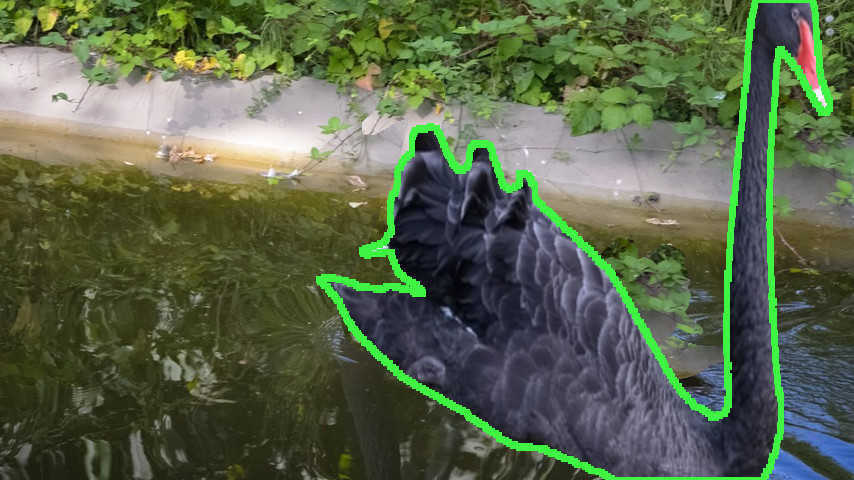} & \includegraphics[width=0.24\textwidth]{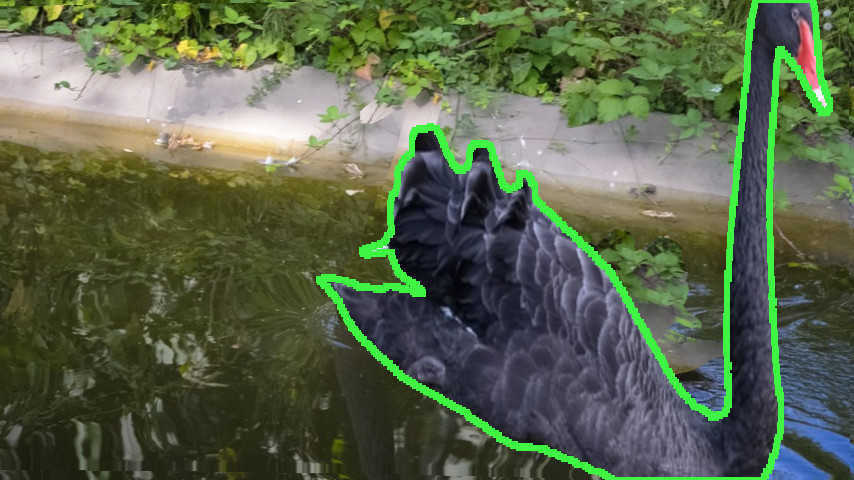} & \includegraphics[width=0.24\textwidth]{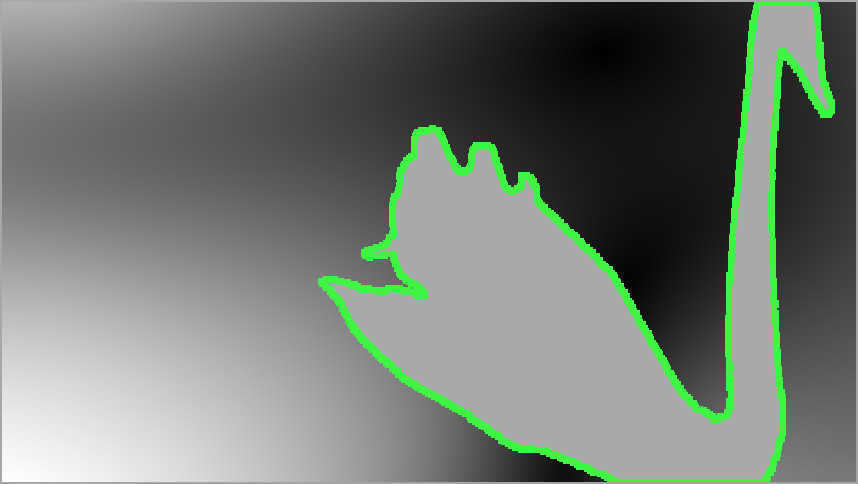}\tabularnewline
\multicolumn{6}{c}{\vspace{0cm}
}\tabularnewline
\multirow{2}{*}{%
\begin{tabular}{c}
\includegraphics[width=0.24\textwidth]{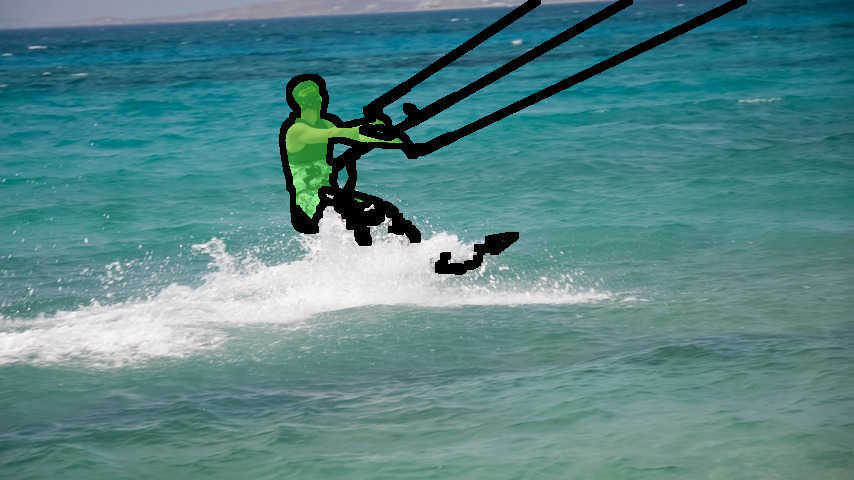}\tabularnewline
kite-surf\tabularnewline
\end{tabular}} &  &  & \includegraphics[width=0.24\textwidth]{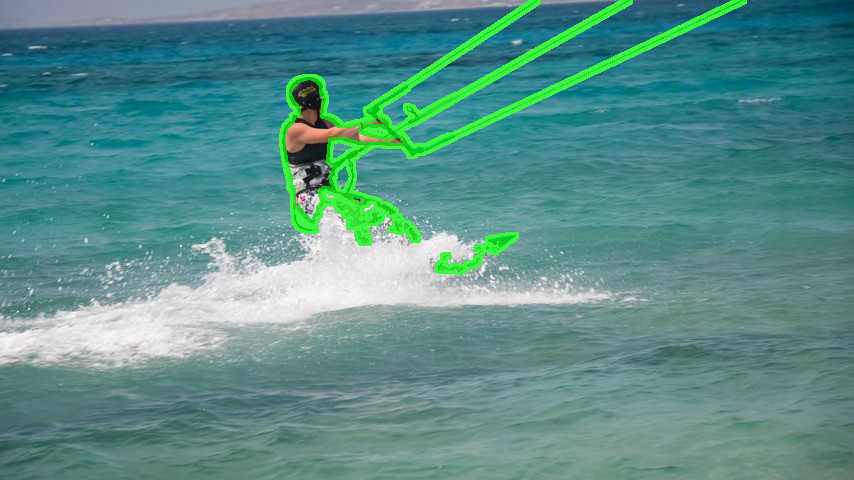} & \includegraphics[width=0.24\textwidth]{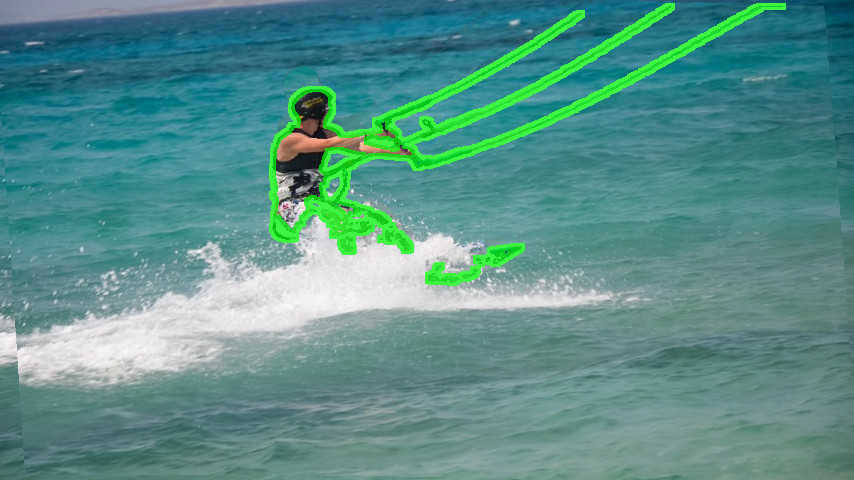} & \includegraphics[width=0.24\textwidth]{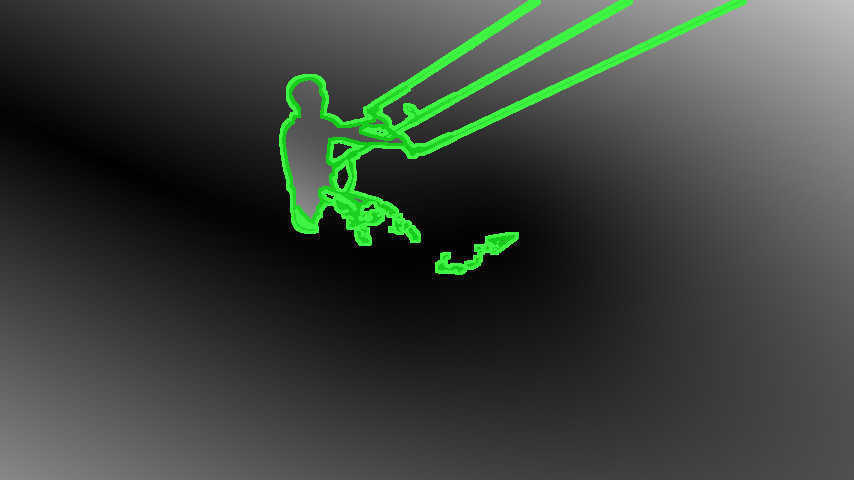}\tabularnewline
 &  &  & \includegraphics[width=0.24\textwidth]{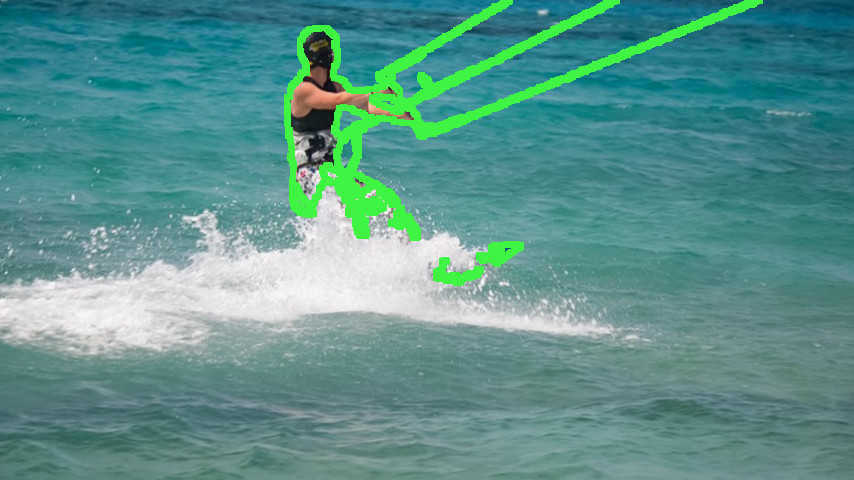} & \includegraphics[width=0.24\textwidth]{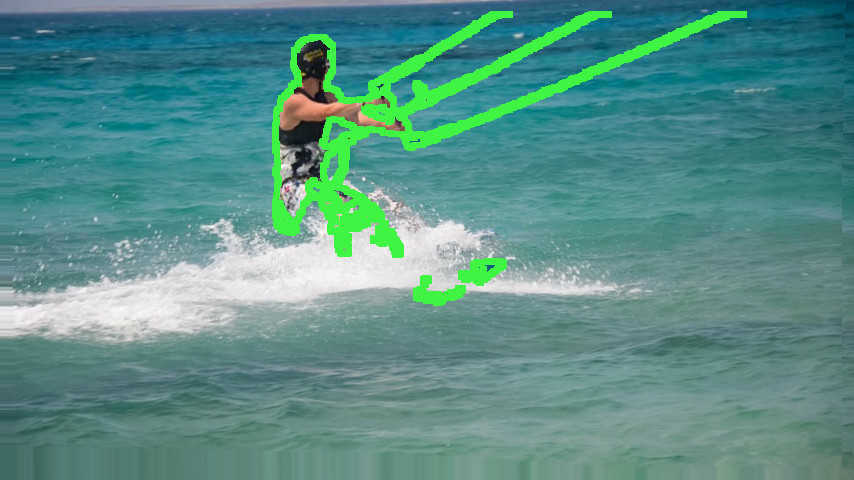} & \includegraphics[width=0.24\textwidth]{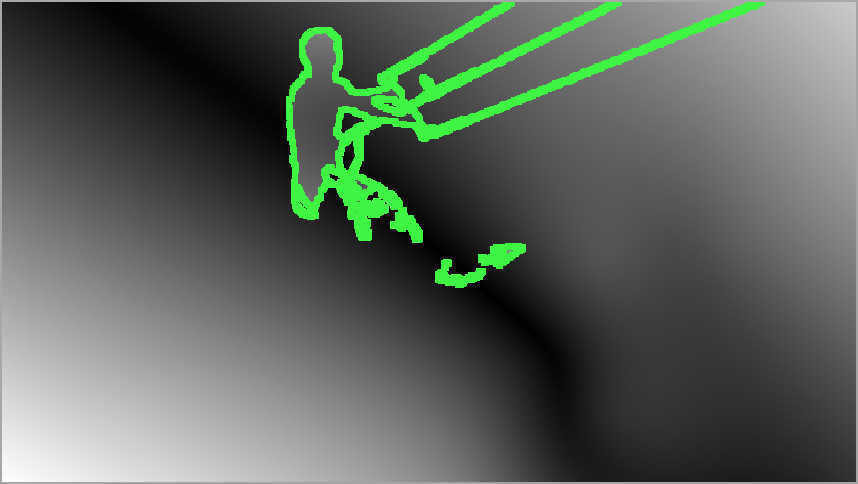}\tabularnewline
\multicolumn{6}{c}{\vspace{0cm}
}\tabularnewline
\multirow{2}{*}{%
\begin{tabular}{c}
\includegraphics[width=0.24\textwidth]{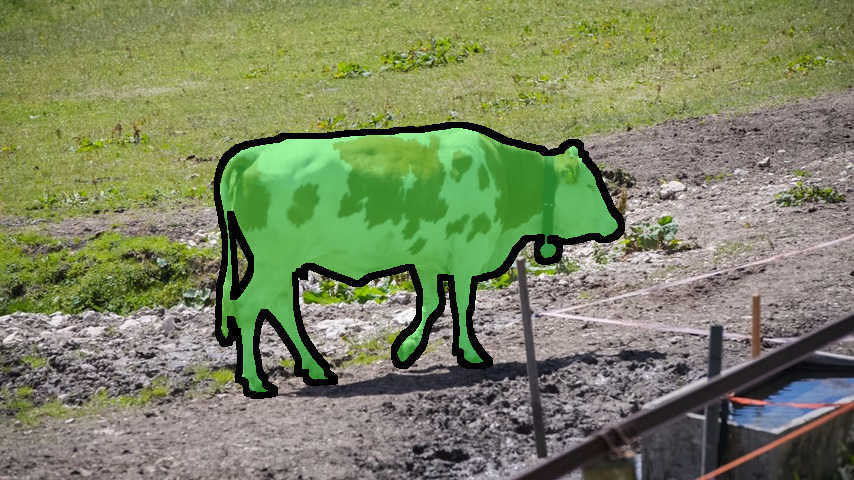}\tabularnewline
cows\tabularnewline
\end{tabular}} &  &  & \includegraphics[width=0.24\textwidth]{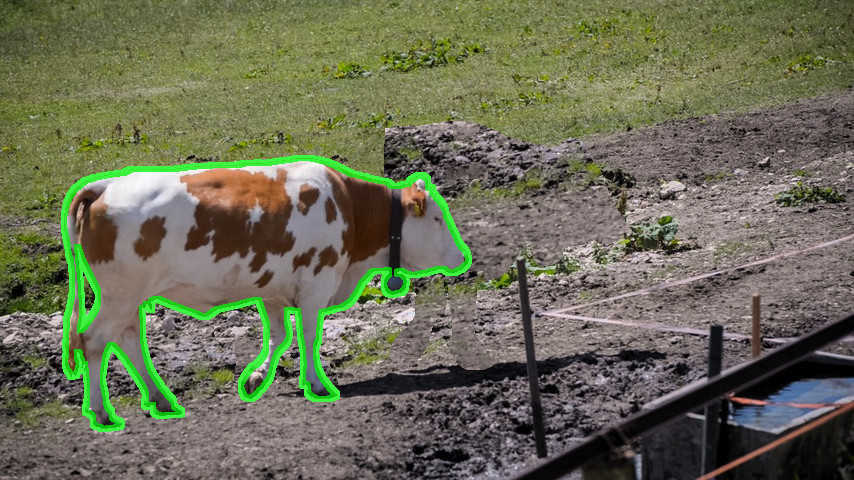} & \includegraphics[width=0.24\textwidth]{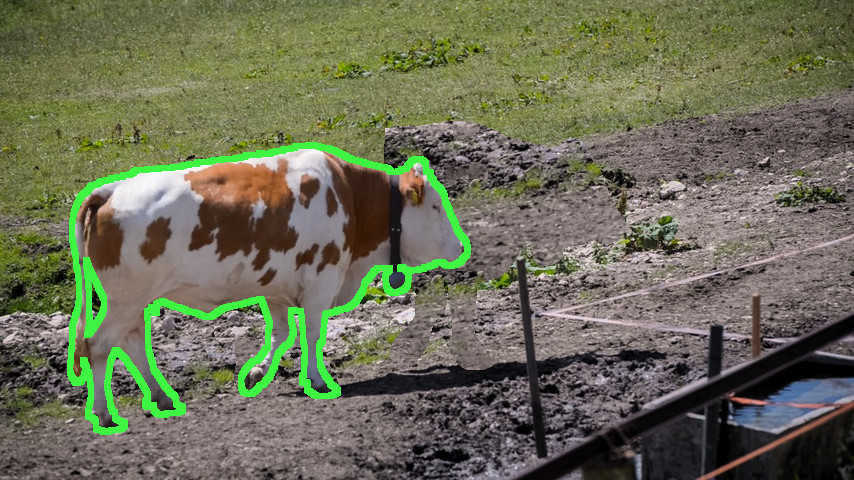} & \includegraphics[width=0.24\textwidth]{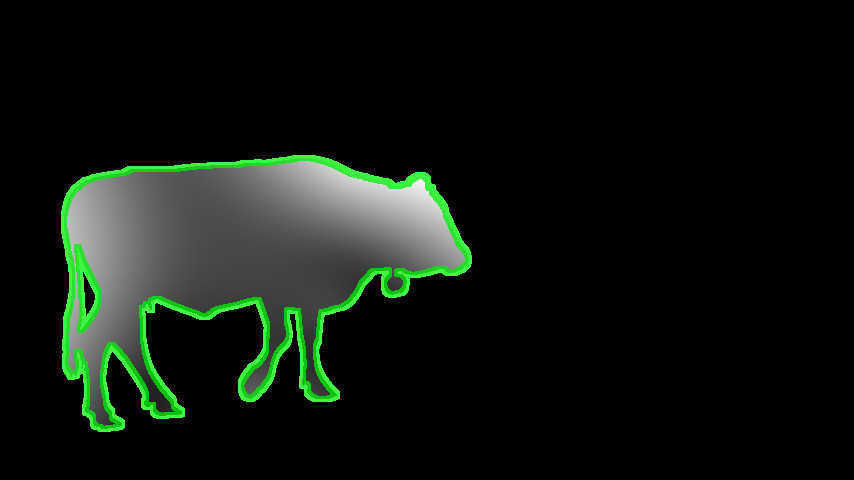}\tabularnewline
 &  &  & \includegraphics[width=0.24\textwidth]{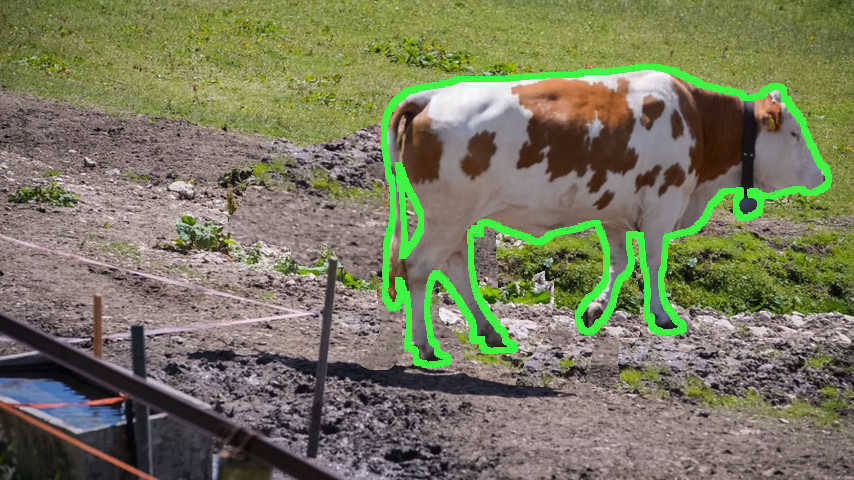} & \includegraphics[width=0.24\textwidth]{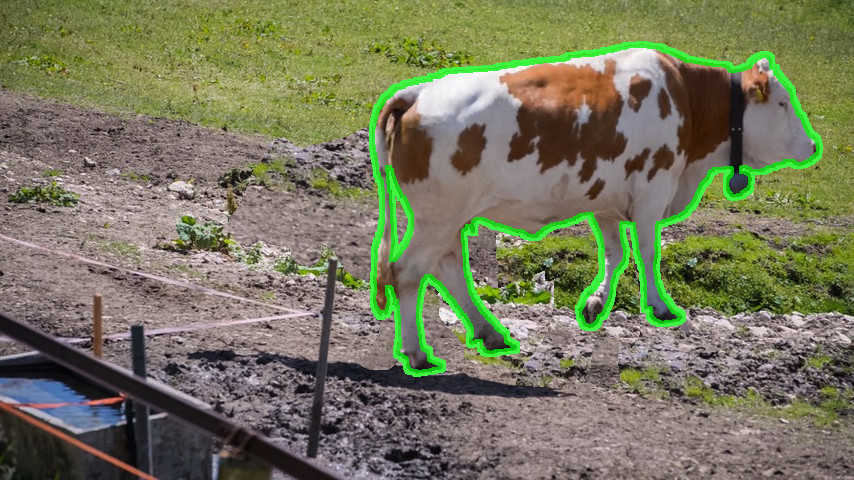} & \includegraphics[width=0.24\textwidth]{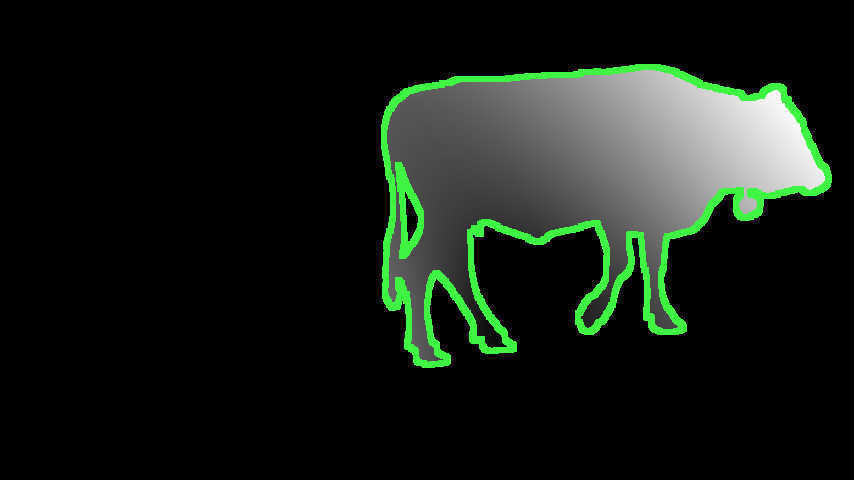}\tabularnewline
\multicolumn{6}{c}{\vspace{0cm}
}\tabularnewline
\multirow{2}{*}{%
\begin{tabular}{c}
\includegraphics[width=0.24\textwidth]{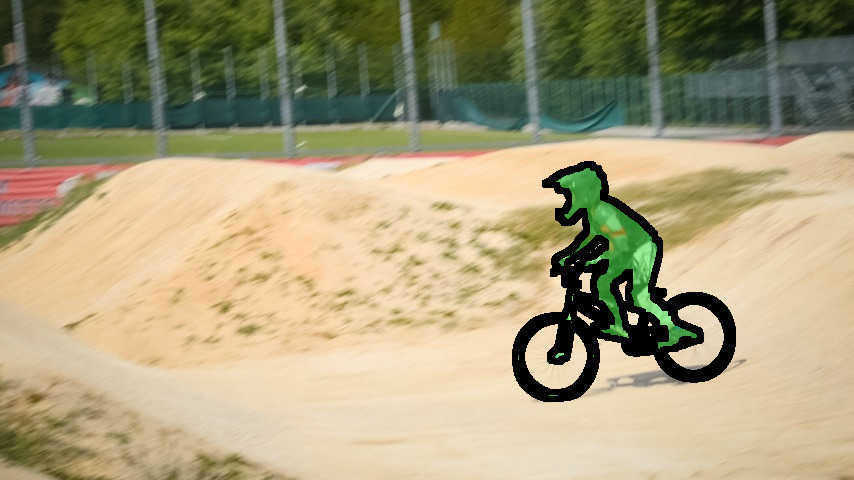}\tabularnewline
bmx-bumps\tabularnewline
\end{tabular}} &  &  & \includegraphics[width=0.24\textwidth]{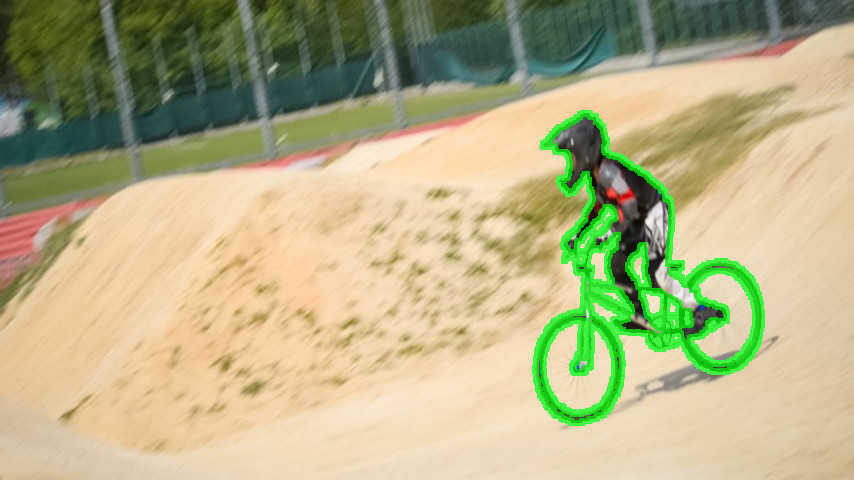} & \includegraphics[width=0.24\textwidth]{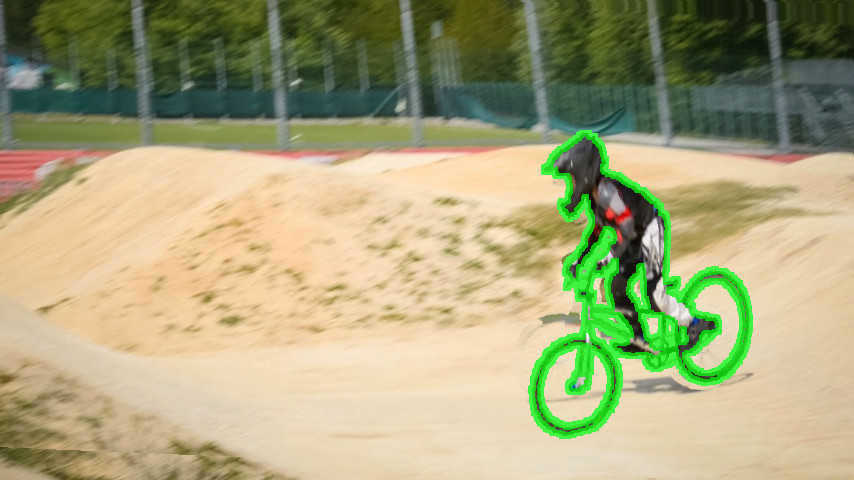} & \includegraphics[width=0.24\textwidth]{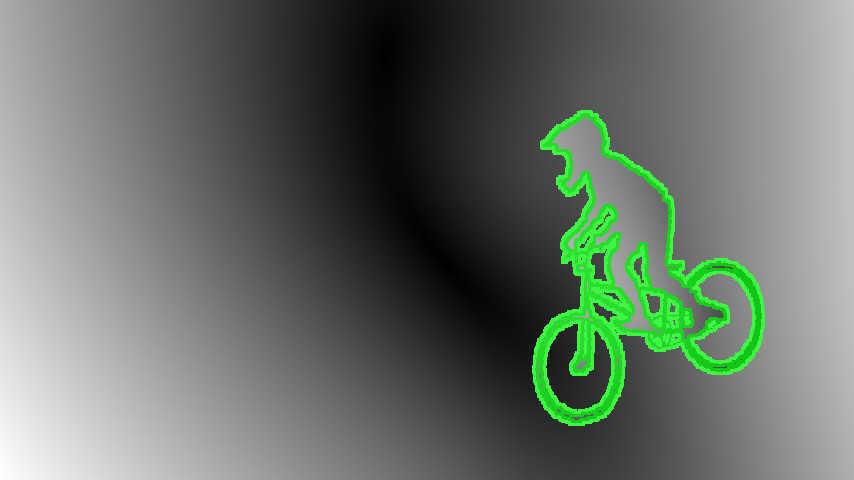}\tabularnewline
 &  &  & \includegraphics[width=0.24\textwidth]{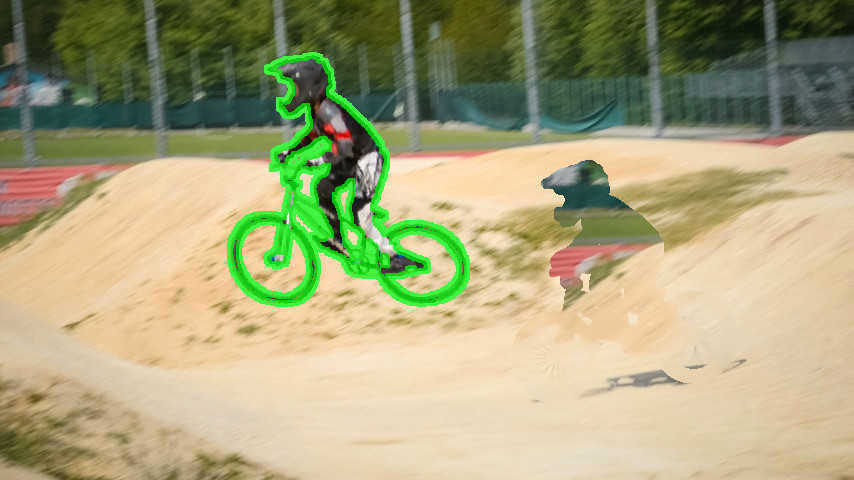} & \includegraphics[width=0.24\textwidth]{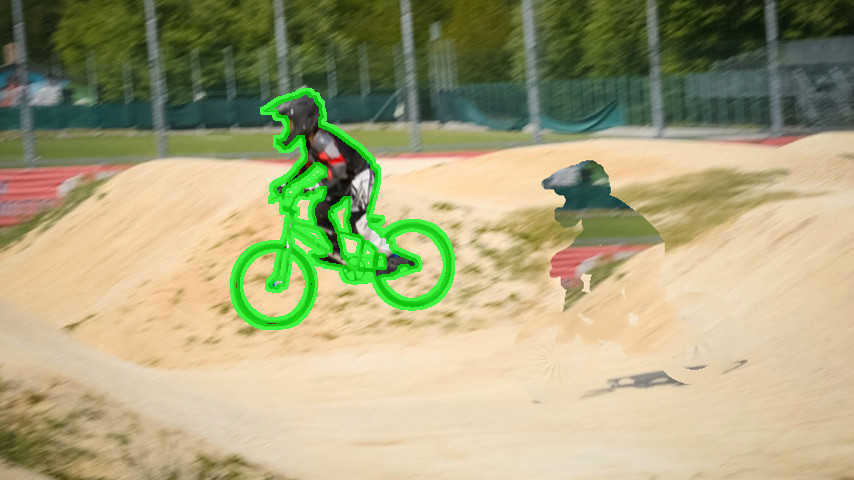} & \includegraphics[width=0.24\textwidth]{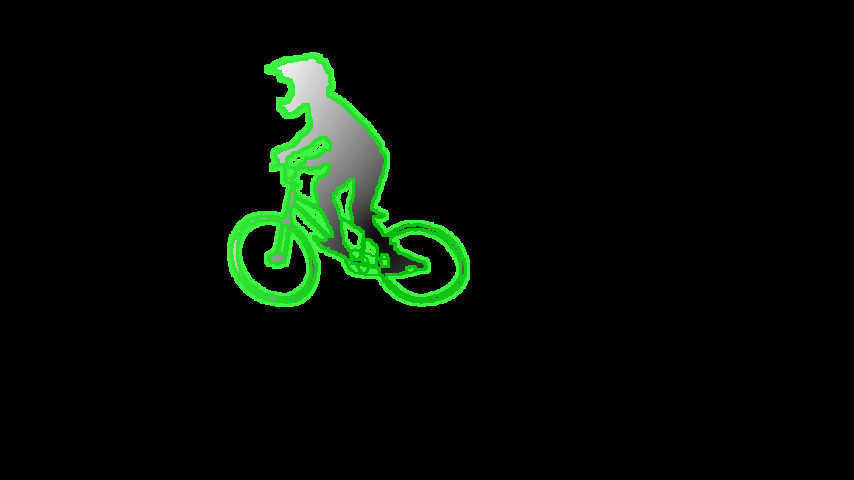}\tabularnewline
\multicolumn{6}{c}{\vspace{0cm}
}\tabularnewline
\begin{tabular}{c}
Original image $\mathcal{I}_{0}$ and\tabularnewline
mask annotation $M_{0}$\tabularnewline
\end{tabular}  & \multicolumn{1}{c}{} &  & Generated image $\mathcal{I}_{\tau-1}$ & Generated image $\mathcal{I}_{\tau}$ & %
\begin{tabular}{c}
Generated flow\tabularnewline
magnitude $\left\Vert \mathcal{F}_{\tau}\right\Vert $\tabularnewline
\end{tabular} \tabularnewline
\multicolumn{6}{c}{\vspace{0em}
}\tabularnewline
\end{tabular}\hfill{}

\caption{\label{fig:data-augmentation}\label{fig:data-augmentation-2}Lucid
data dreaming examples. From one annotated frame we generate pairs
of images ($\mathcal{I}_{\tau-1},\,\mathcal{I}_{\tau}$) that are
plausible future video frames, with known optical flow ($\mathcal{F}_{\tau}$)
and masks (green boundaries). Note the inpainted background and foreground/background
deformations.}
\end{figure*}

Ideally training data should be as similar as possible to the test
data, even subtle differences may affect quality (e.g. training on
static images for testing on videos under-performs \cite{Tang_NIPS2012}).
To ensure our training data is in-domain, we propose to generate it
by synthesizing samples from the provided annotated frame (first frame)
in each target video. This is akin to ``lucid dreaming'' as we intentionally
``dream'' the desired data by creating sample images that are plausible
hypothetical future frames of the video. The outcome of this process
is a large set of frame pairs in the target domain ($2.5k$ pairs
per annotation) with known optical flow and mask annotations, see
Figure \ref{fig:data-augmentation}. 

\paragraph{Synthesis process}

The target domain for a tracker is the set of future frames of the
given video. Traditional data augmentation via small image perturbation
is insufficient to cover the expect variations across time, thus a
task specific strategy is needed. Across the video the tracked object
might change in illumination, deform, translate, be occluded, show
different point of views, and evolve on top of a dynamic background.
All of these aspects should be captured when synthesizing future frames.
We achieve this by cutting-out the foreground object, in-painting
the background, perturbing both foreground and background, and finally
recomposing the scene. This process is applied twice with randomly
sampled transformation parameters, resulting in a pair of frames ($\mathcal{I}_{\tau-1},\,\mathcal{I}_{\tau}$)
with known pixel-level ground-truth mask annotations ($M_{\tau-1},\,M_{\tau}$),
optical flow $\mathcal{F}_{\tau}$, and occlusion regions. The object
position in $\mathcal{I}_{\tau}$ is uniformly sampled, but the changes
between $\mathcal{I}_{\tau-1},\,\mathcal{I}_{\tau}$ are kept small
to mimic the usual evolution between consecutive frames.\\
In more details, starting from an annotated image:\\
\emph{1. Illumination changes:} we globally modify the image by randomly
altering saturation S and value V (from HSV colour space) via $x'=a\cdot x^{b}+c$,
where $a\in1\pm0.05$, $b\in1\pm0.3$, and $c\in\pm0.07$.\\
\emph{2. Fg/Bg split:} the foreground object is removed from the image
$\mathcal{I}_{0}$ and a background image is created by inpainting
the cut-out area \cite{Criminisi2004}. \\
\emph{3. Object motion:} we simulate motion and shape deformations
by applying global translation as well as affine and non-rigid deformations
to the foreground object. For $\mathcal{I}_{\tau-1}$ the object is
placed at any location within the image with a uniform distribution,
and in $\mathcal{I}_{\tau}$ with a translation of $\pm10\%$ of the
object size relative to $\tau-1$. In both frames we apply random
rotation $\pm30^{\circ}$, scaling $\pm15\%$ and thin-plate splines
deformations \cite{Bookstein1989Pami} of $\pm10\%$ of the object
size.\\
\emph{4. Camera motion:} We additionally transform the background
using affine deformations to simulate camera view changes. We apply
here random translation, rotation, and scaling within the same ranges
as for the foreground object.\\
\emph{5. Fg/Bg merge:} finally ($\mathcal{I}_{\tau-1},\,\mathcal{I}_{\tau}$)
are composed by blending the perturbed foreground with the perturbed
background using Poisson matting \cite{Sun2004}. Using the known
transformation parameters we also synthesize ground-truth pixel-level
mask annotations ($M_{\tau-1},\,M_{\tau}$) and optical flow $\mathcal{F}_{\tau}$.\\
Figure \ref{fig:data-augmentation} shows example results. Albeit
our approach does not capture appearance changes due to point of view,
occlusions, nor shadows, we see that already this rough modelling
is effective to train our segmentation models.

The number of synthesized images can be arbitrarily large. We generate
$2.5k$ pairs per annotated video frame. This training data is, by
design, in-domain with regard of the target video. The experimental
section \ref{sec:Single-object-results} shows that this strategy
is more effective than using thousands of manually annotated images
from close-by domains.

\begin{figure}
\setlength{\tabcolsep}{0.1em} 
\renewcommand{\arraystretch}{0}

\vspace{0em}
\hfill{}\subfloat[Original image $\mathcal{I}_{0}$ and mask annotation $M_{0}$]{\centering{}%
\begin{tabular}{cc}
\includegraphics[width=0.45\columnwidth]{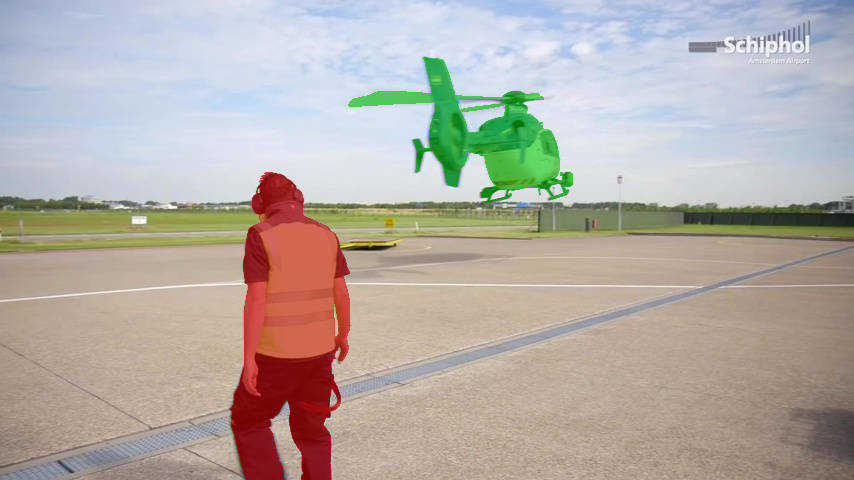} & \includegraphics[width=0.45\columnwidth]{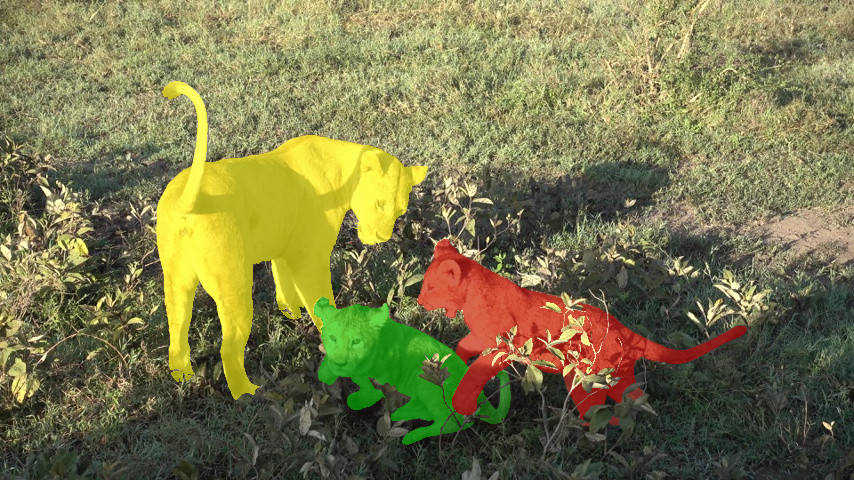}\tabularnewline
\end{tabular}}\hfill{}

\vspace{-1em}
\hfill{}\subfloat[Generated image $\mathcal{I}_{\tau}$ and mask $M_{\tau}$]{\centering{}%
\begin{tabular}{cc}
\includegraphics[width=0.45\columnwidth]{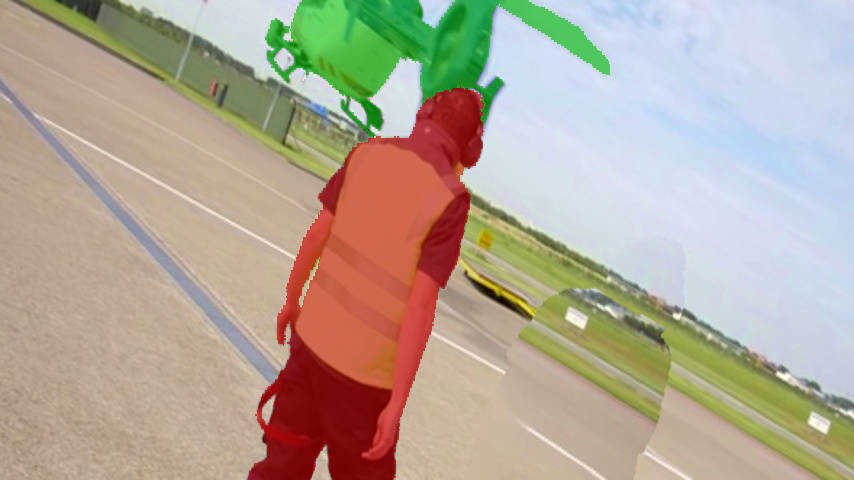} & \includegraphics[width=0.45\columnwidth]{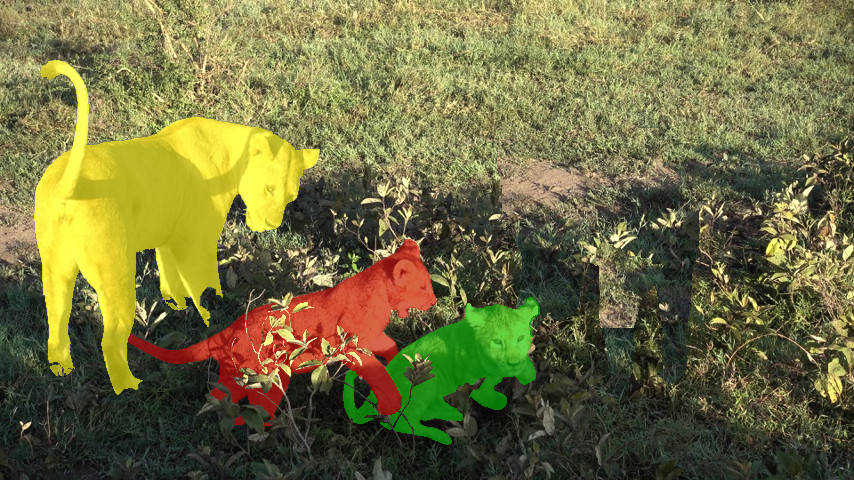}\tabularnewline
\end{tabular}}\hfill{}

\vspace{-1em}
\hfill{}\subfloat[Generated flow magnitude $\left\Vert \mathcal{F}_{\tau}\right\Vert $]{\centering{}%
\begin{tabular}{cc}
\includegraphics[width=0.45\columnwidth]{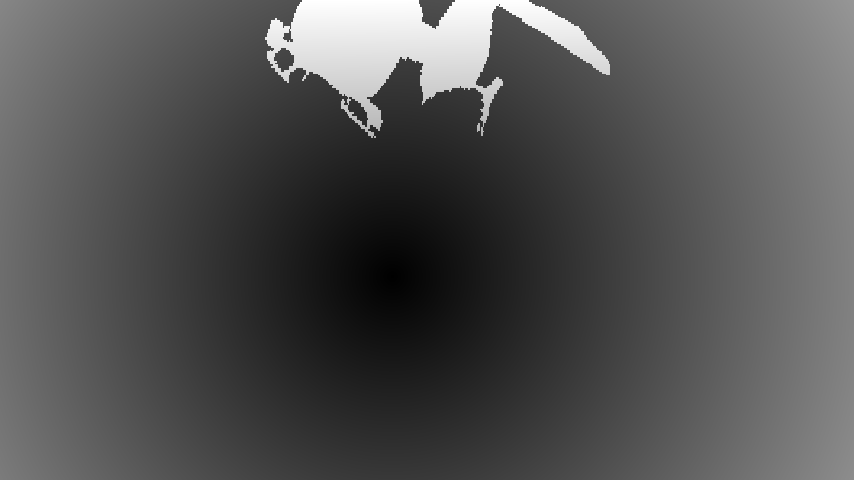} & \includegraphics[width=0.45\columnwidth]{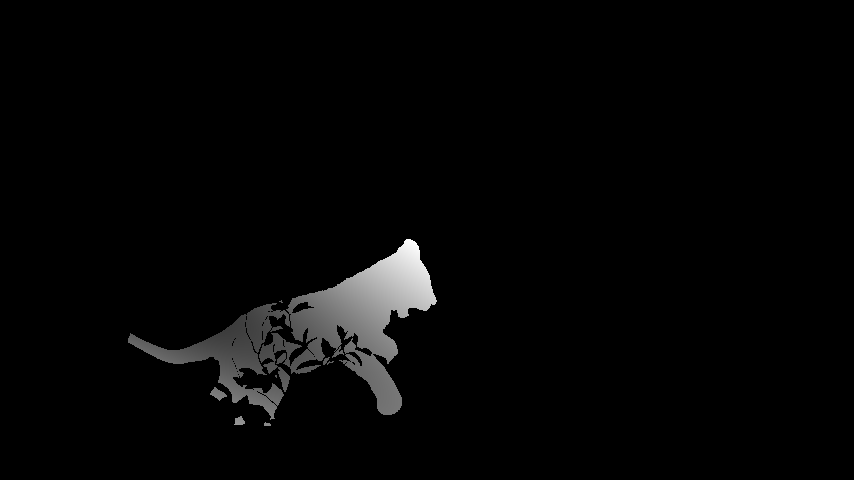}\tabularnewline
\end{tabular}}\hfill{}

\caption{\label{fig:data-augmentatoin-mult}Lucid data dreaming examples with
multiple objects. From one annotated frame we generate a plausible
future video frame ($\mathcal{I}_{\tau}$), with known optical flow
($\mathcal{F}_{\tau}$) and mask ($M_{\tau}$). }
\vspace{0em}
\end{figure}

The same strategy for data synthesis can be employed for multiple
object segmentation task. Instead of manipulating a single object we handle
multiple ones at the same time, applying independent transformations
to each of them. We model occlusion between objects by adding a random
depth ordering obtaining both partial and full occlusions in the training
set. Including occlusions in the lucid dreams allows to better handle
plausible interactions of objects in the future frames. See Figure
\ref{fig:data-augmentatoin-mult} for examples of the generated data.

\vspace{0em}

\section{\label{sec:Single-object-results}Single object segmentation results}

We present here a detailed empirical evaluation on three different
datasets for the single object segmentation task: given a first frame
labelled with the foreground object mask, the goal is to find the
corresponding object pixels in future frames. (Section \ref{sec:Multiple-object-results}
will discuss the multiple objects case.)

\subsection{\label{subsec:Experimental-setup}Experimental setup}

\paragraph{Datasets}

We evaluate our method on three video object segmentation datasets:
$\text{DAVIS}_{\text{16}}$ \cite{Perazzi2016Cvpr}, YouTubeObjects
\cite{Prest2012Cvpr,Jain2014Eccv}, and $\text{SegTrack}_{\text{v2}}$
\cite{Li2013Iccv}. The goal is to track an object through all video
frames given an object mask in the first frame. These three
datasets provide diverse challenges with a mix of high and low resolution
web videos, single or multiple salient objects per video, videos with
flocks of similar looking instances, longer ($\sim\negmedspace400$
frames) and shorter ($\sim\negmedspace10$ frames) sequences, as well
as the usual video segmentation challenges such as occlusion, fast motion, illumination,
view point changes, elastic deformation, etc. 

The $\text{DAVIS}_{\text{16}}$ \cite{Perazzi2016Cvpr} video segmentation
benchmark consists of 50 full-HD videos of diverse object categories
with all frames annotated with pixel-level accuracy, where one single
or two connected moving objects are separated from the background.
The number of frames in each video varies from 25 to 104.

YouTubeObjects \cite{Prest2012Cvpr,Jain2014Eccv} includes web videos
from 10 object categories. We use the subset of 126 video sequences
with mask annotations provided by \cite{Jain2014Eccv} for evaluation,
where one single object or a group of objects of the same category
are separated from the background. In contrast to $\text{DAVIS}_{\text{16}}$
these videos have a mix of static and moving objects. The number of
frames in each video ranges from 2 to 401.

$\text{SegTrack}_{\text{v2}}$\cite{Li2013Iccv} consists of 14 videos
with multiple object annotations for each frame. For videos with multiple
objects each object is treated as a separate problem, resulting in
24 sequences. The length of each video varies from 21 to 279 frames.
The images in this dataset have low resolution and some compression
artefacts, making it hard to track the object based on its appearance.

The main experimental work is done on $\text{DAVIS}_{\text{16}}$,
since it is the largest densely annotated dataset out of the three,
and provides high quality/high resolution data. The videos for this
dataset were chosen to represent diverse challenges, making it a good
experimental playground. 

We additionally report on the two other datasets as complementary
test set results.

\paragraph{Evaluation metric}

To measure the accuracy of video object segmentation we use the mean intersection-over-union
overlap (mIoU) between the per-frame ground truth object mask and
the predicted segmentation, averaged across all video sequences. We
have noticed disparate evaluation procedures used in previous work,
and we report here a unified evaluation across datasets. When possible,
we re-evaluated certain methods using results provided by their authors.
For all three datasets we follow the $\text{DAVIS}_{\text{16}}$ evaluation
protocol, excluding the first frame from evaluation and using all
other frames from the video sequences, independent of object presence
in the frame. 
\begin{figure*}
\begin{centering}
\setlength{\tabcolsep}{0em} 
\renewcommand{\arraystretch}{0.5}
\par\end{centering}
\begin{centering}
\vspace{0em}
\hfill{}%
\begin{tabular}{ccccccc}
\multirow{4}{*}{\begin{turn}{90}
\hspace{-5em} $\text{DAVIS}_{\text{16}}$\hspace{-5em}
\end{turn}} & {\footnotesize{}\hspace{-1em}}\includegraphics[width=0.15\textwidth]{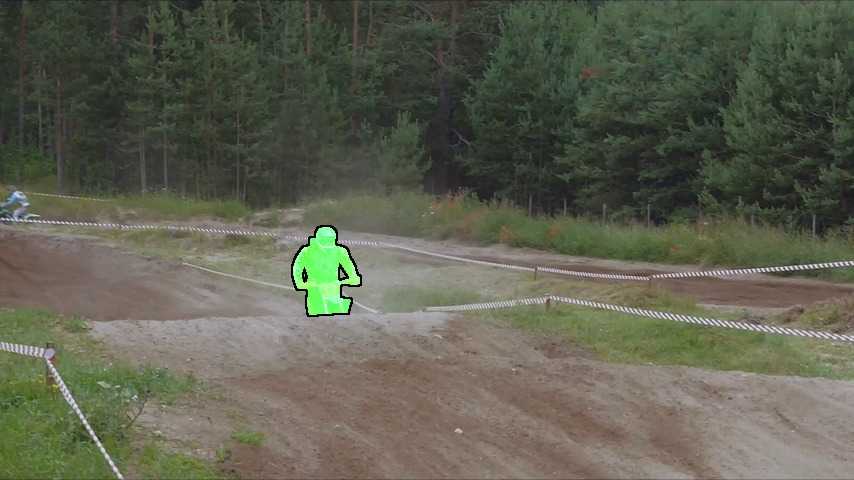} & {\footnotesize{}\hspace{-1em}}\includegraphics[width=0.15\textwidth]{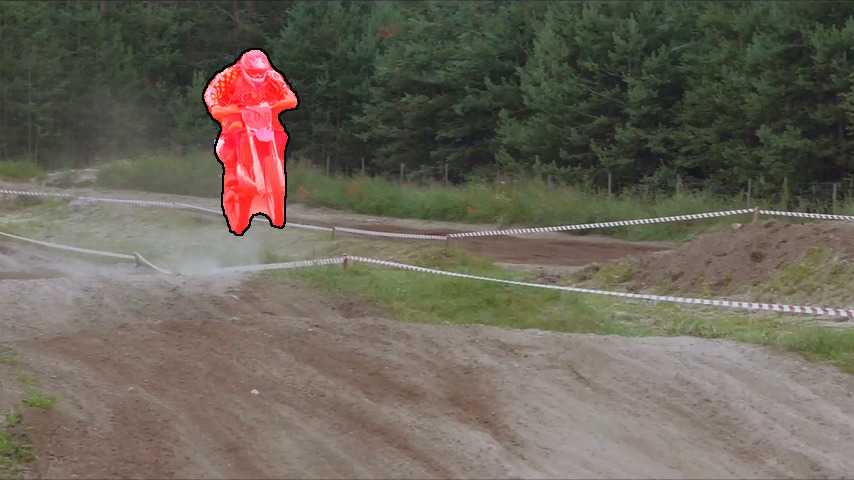} & {\footnotesize{}\hspace{-1em}}\includegraphics[width=0.15\textwidth]{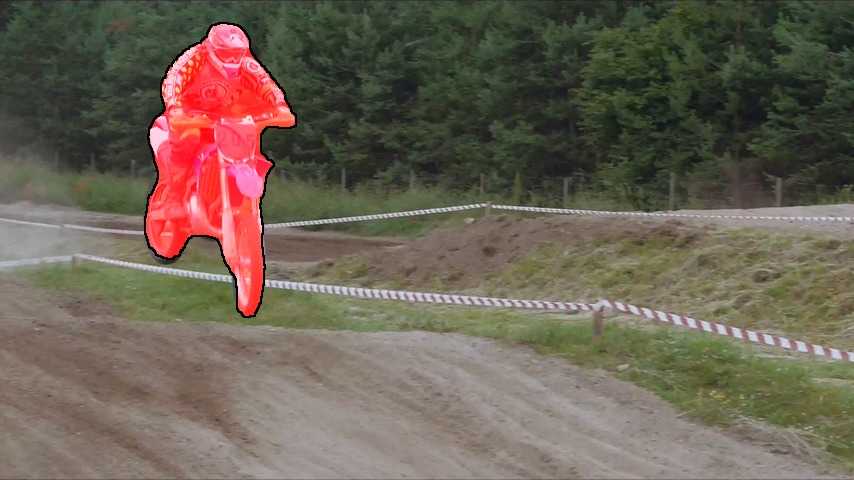} & {\footnotesize{}\hspace{-1em}}\includegraphics[width=0.15\textwidth]{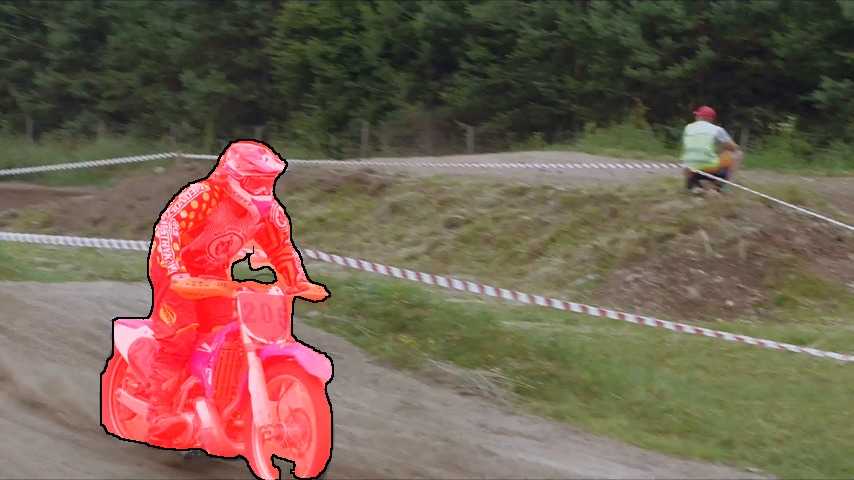} & {\footnotesize{}\hspace{-1em}}\includegraphics[width=0.15\textwidth]{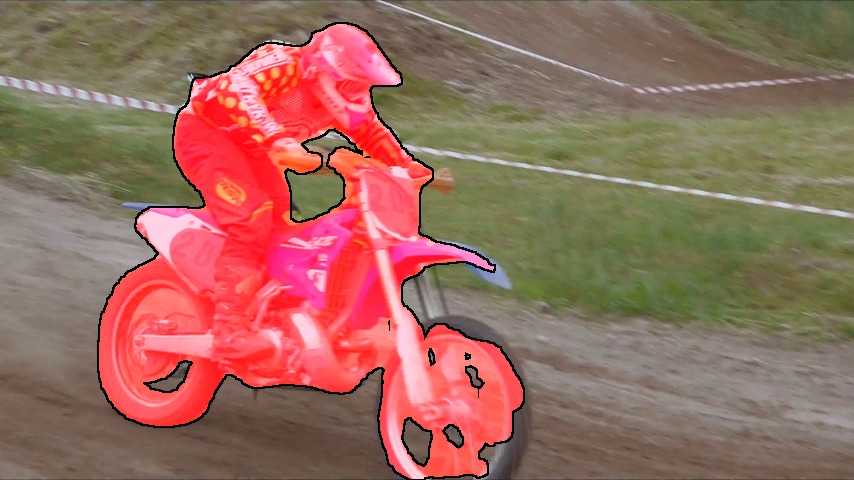} & {\footnotesize{}\hspace{-1em}}\includegraphics[width=0.15\textwidth]{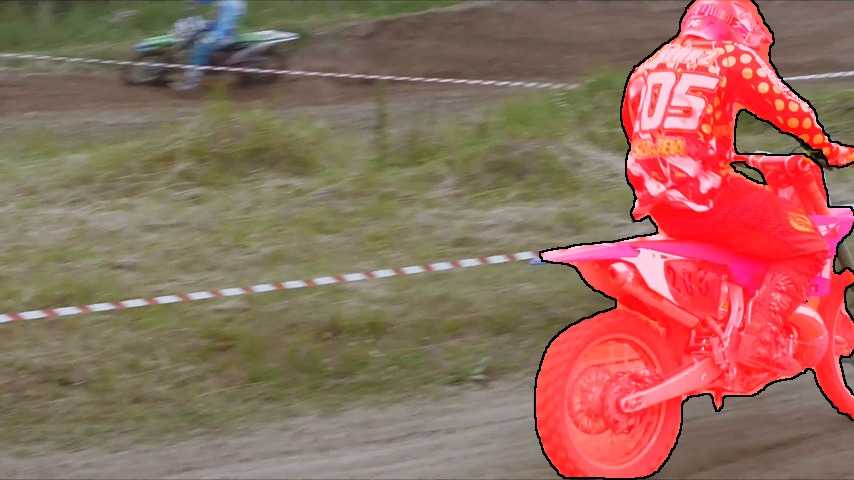}\tabularnewline
 & {\footnotesize{}\hspace{-1em}}\includegraphics[width=0.15\textwidth]{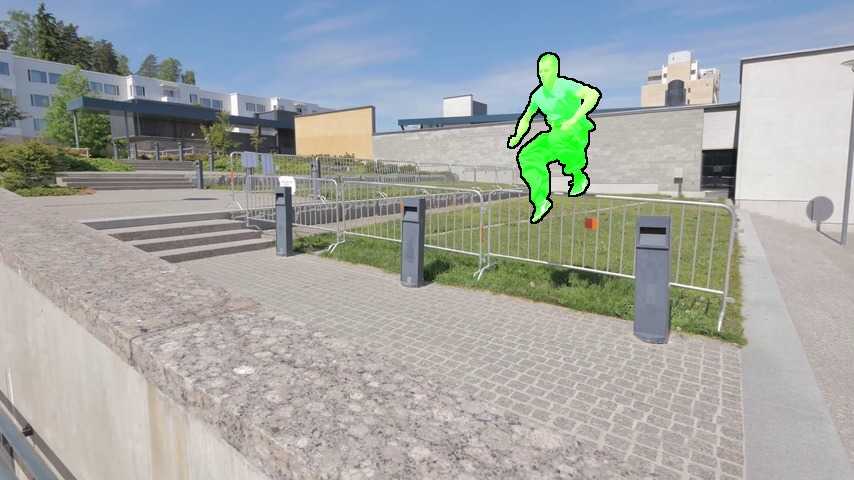} & {\footnotesize{}\hspace{-1em}}\includegraphics[width=0.15\textwidth]{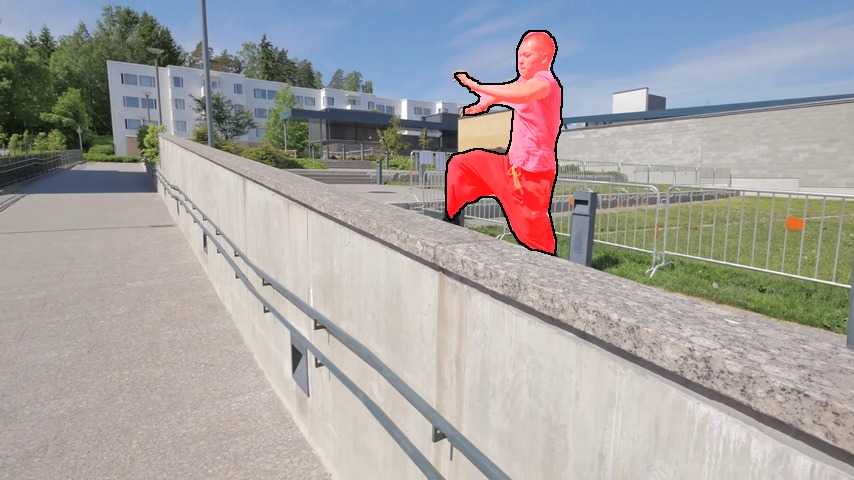} & {\footnotesize{}\hspace{-1em}}\includegraphics[width=0.15\textwidth]{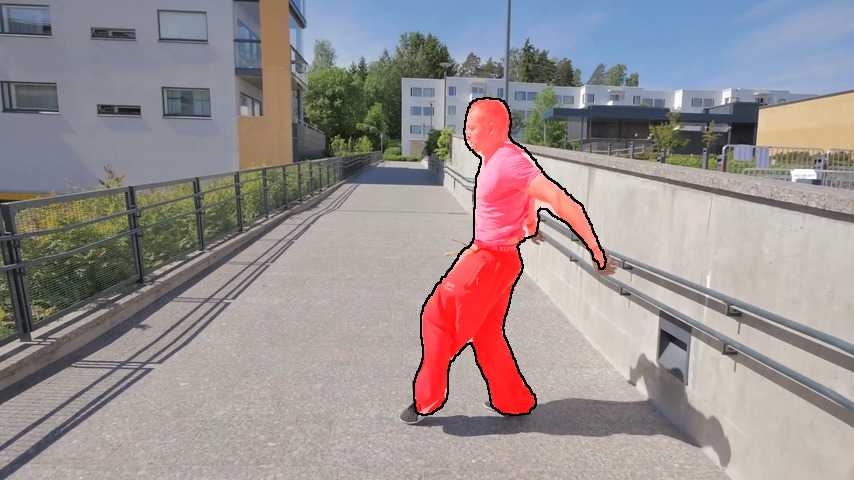} & {\footnotesize{}\hspace{-1em}}\includegraphics[width=0.15\textwidth]{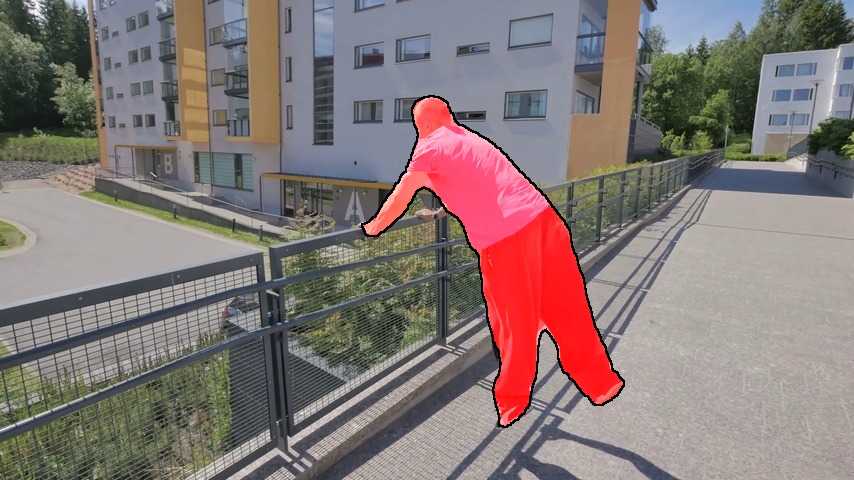} & {\footnotesize{}\hspace{-1em}}\includegraphics[width=0.15\textwidth]{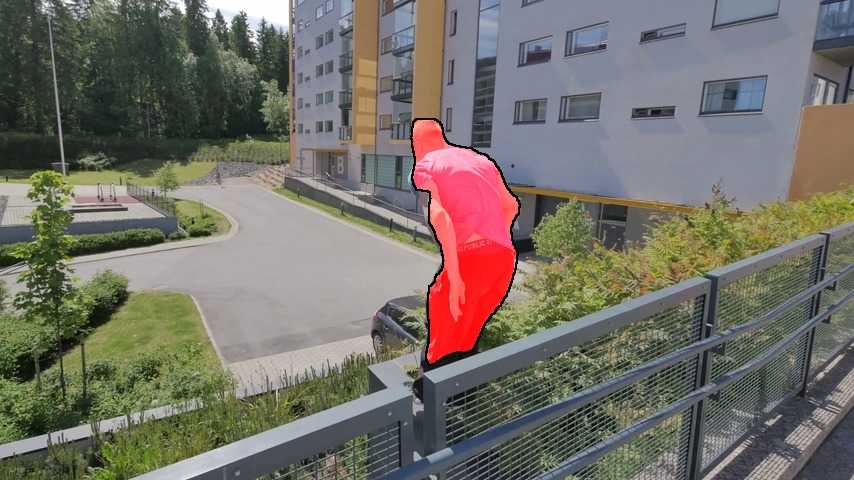} & {\footnotesize{}\hspace{-1em}}\includegraphics[width=0.15\textwidth]{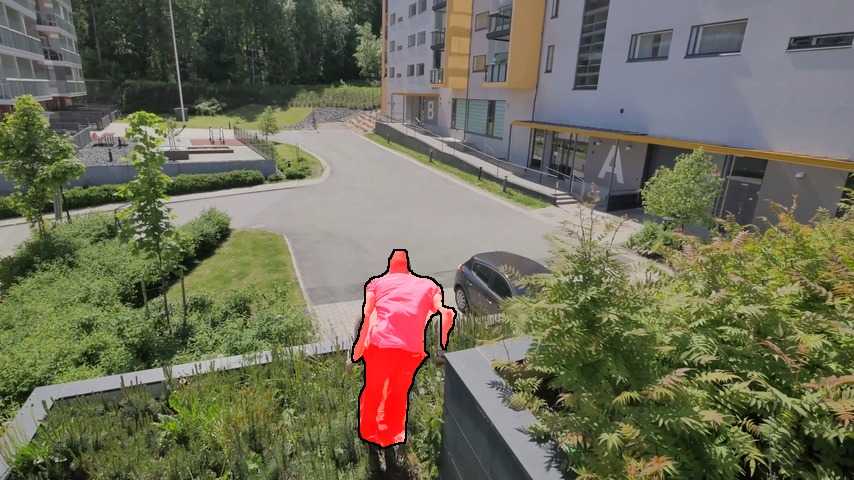}\tabularnewline
 & {\footnotesize{}\hspace{-1em}}\includegraphics[width=0.15\textwidth]{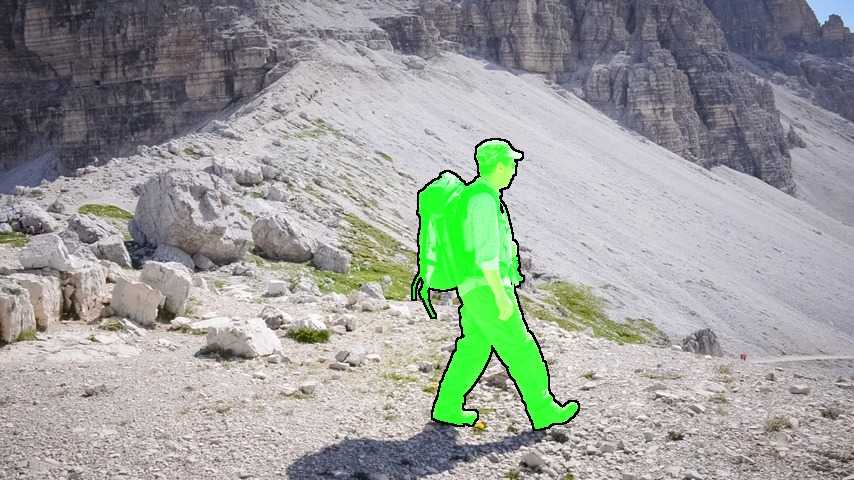} & {\footnotesize{}\hspace{-1em}}\includegraphics[width=0.15\textwidth]{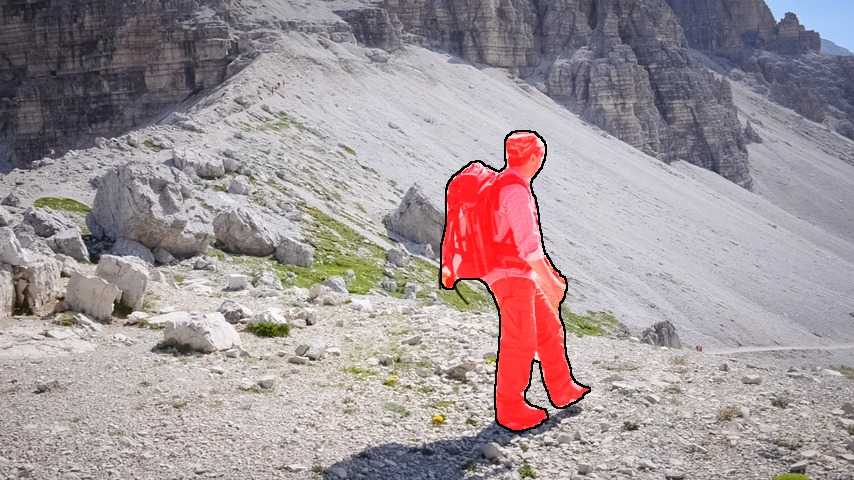} & {\footnotesize{}\hspace{-1em}}\includegraphics[width=0.15\textwidth]{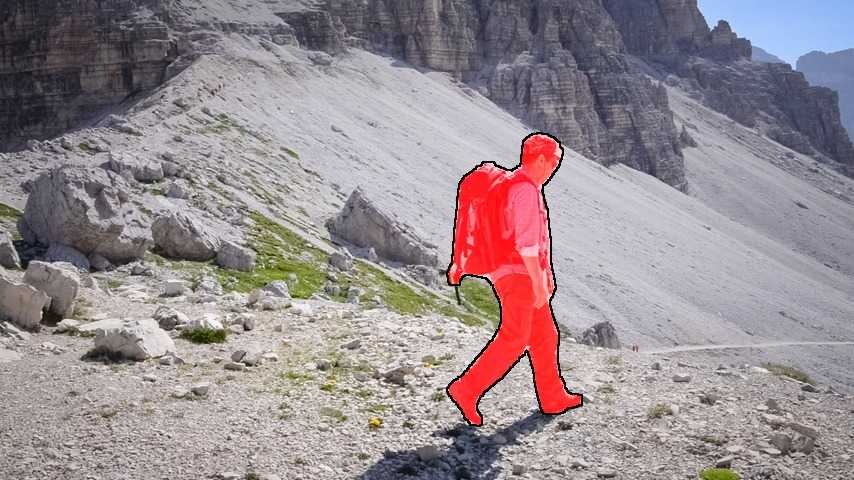} & {\footnotesize{}\hspace{-1em}}\includegraphics[width=0.15\textwidth]{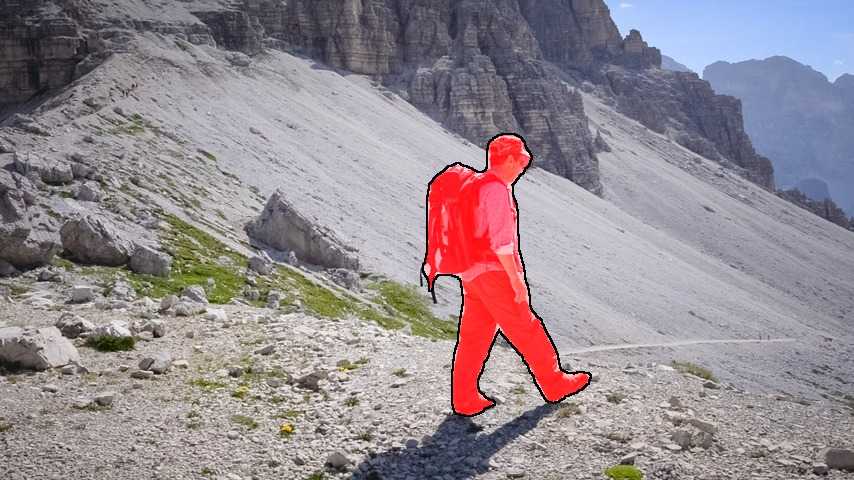} & {\footnotesize{}\hspace{-1em}}\includegraphics[width=0.15\textwidth]{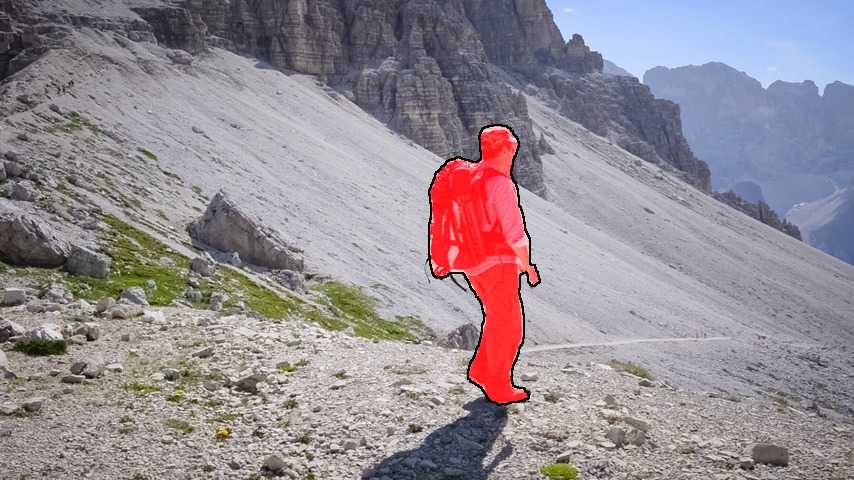} & {\footnotesize{}\hspace{-1em}}\includegraphics[width=0.15\textwidth]{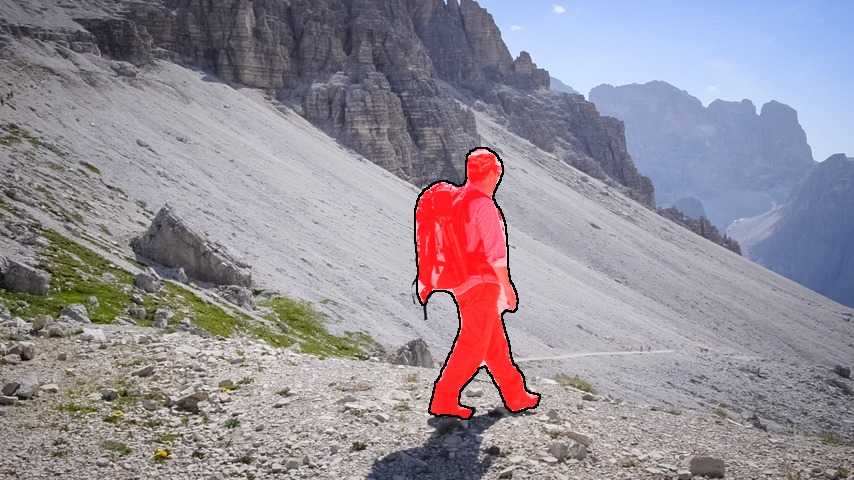}\tabularnewline
 & {\footnotesize{}\hspace{-1em}}\includegraphics[width=0.15\textwidth]{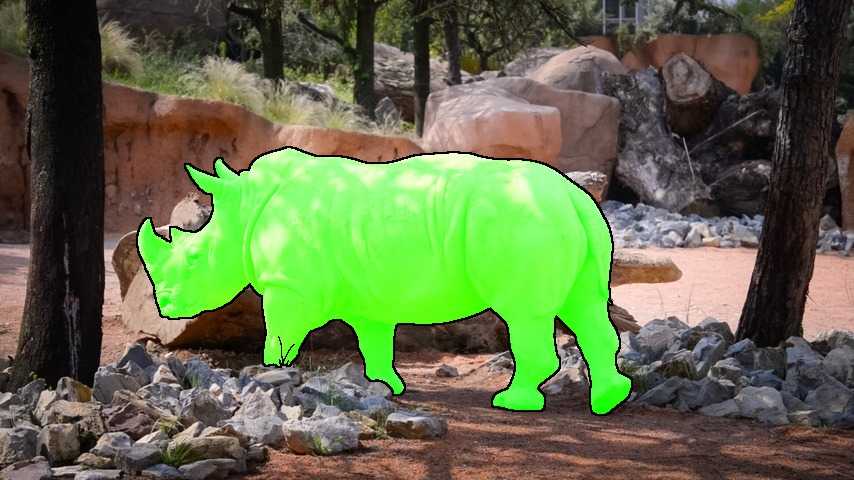} & {\footnotesize{}\hspace{-1em}}\includegraphics[width=0.15\textwidth]{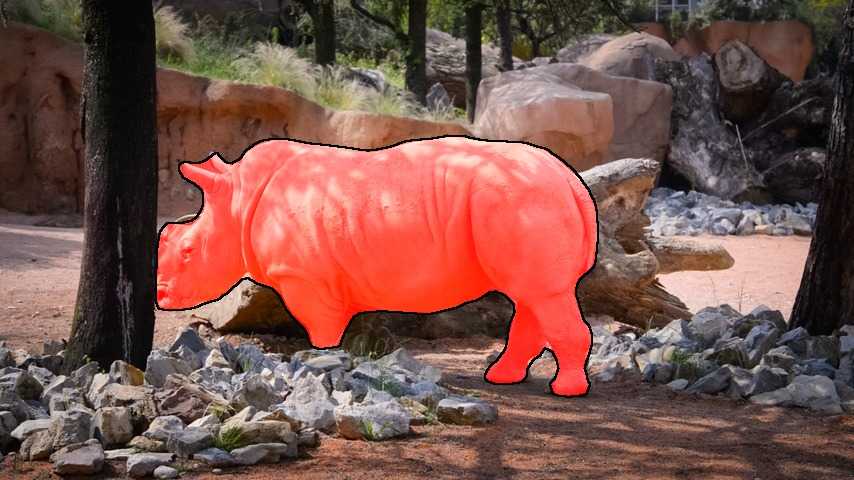} & {\footnotesize{}\hspace{-1em}}\includegraphics[width=0.15\textwidth]{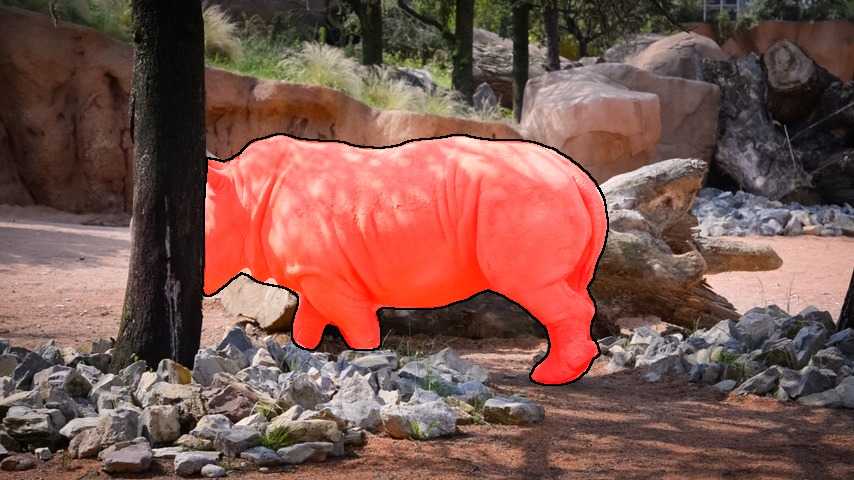} & {\footnotesize{}\hspace{-1em}}\includegraphics[width=0.15\textwidth]{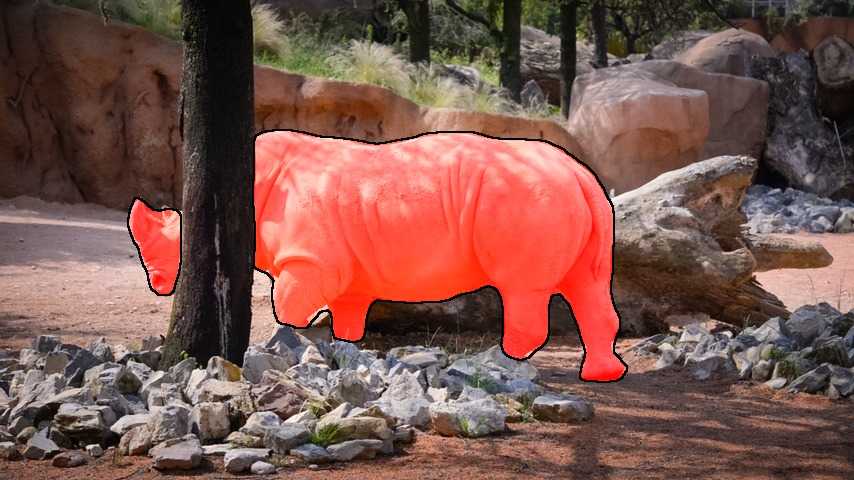} & {\footnotesize{}\hspace{-1em}}\includegraphics[width=0.15\textwidth]{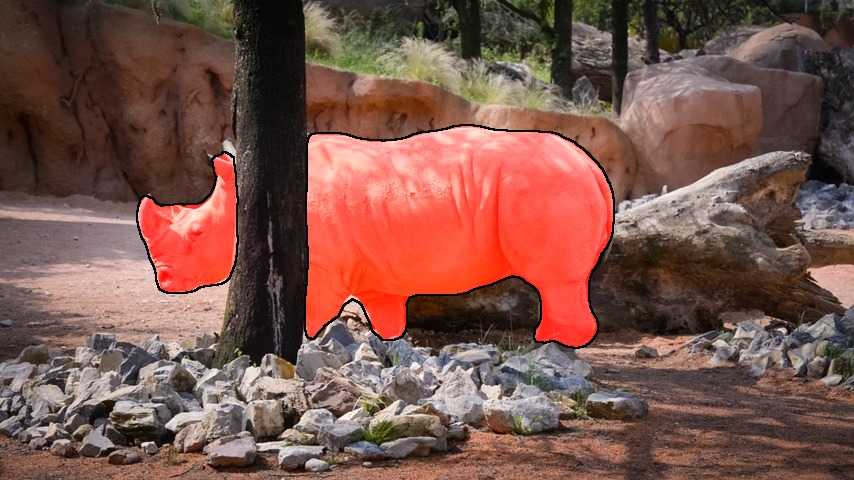} & {\footnotesize{}\hspace{-1em}}\includegraphics[width=0.15\textwidth]{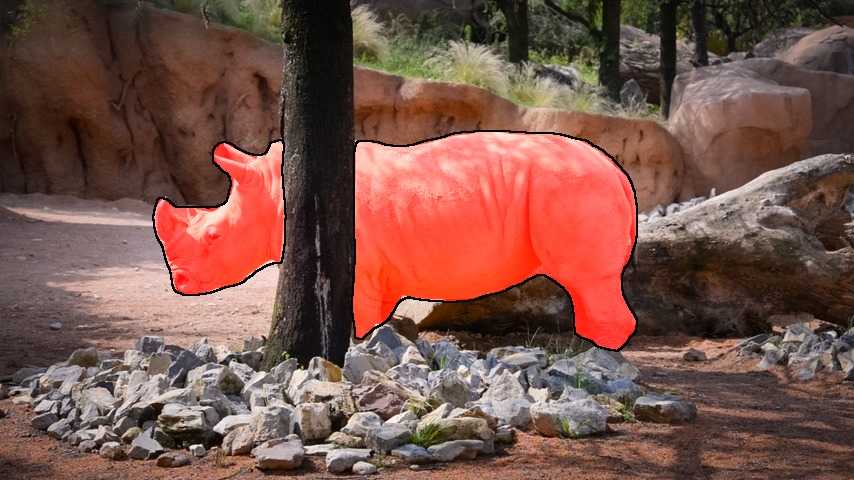}\tabularnewline
 &  &  &  &  &  & \tabularnewline
\multirow{4}{*}{\begin{turn}{90}
\hspace{-7em}YouTubeObjects
\end{turn}} & {\footnotesize{}\hspace{-1em}}\includegraphics[width=0.15\textwidth]{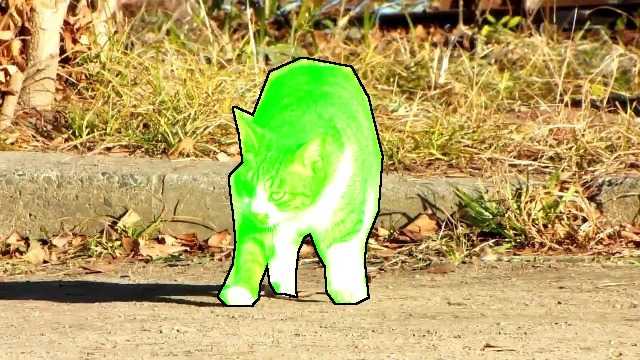} & {\footnotesize{}\hspace{-1em}}\includegraphics[width=0.15\textwidth]{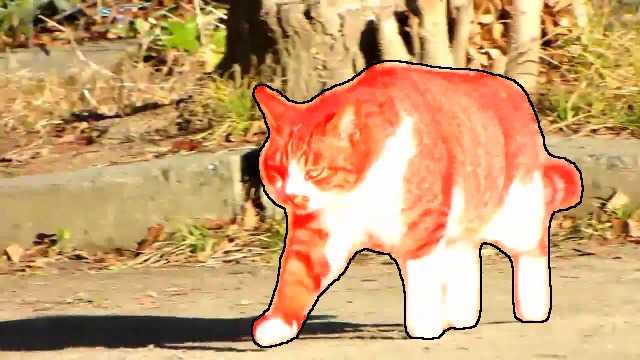} & {\footnotesize{}\hspace{-1em}}\includegraphics[width=0.15\textwidth]{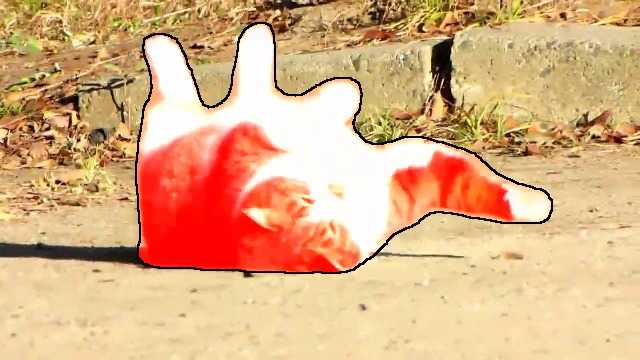} & {\footnotesize{}\hspace{-1em}}\includegraphics[width=0.15\textwidth]{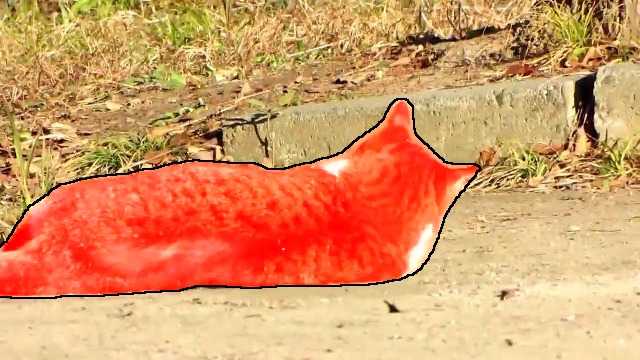} & {\footnotesize{}\hspace{-1em}}\includegraphics[width=0.15\textwidth]{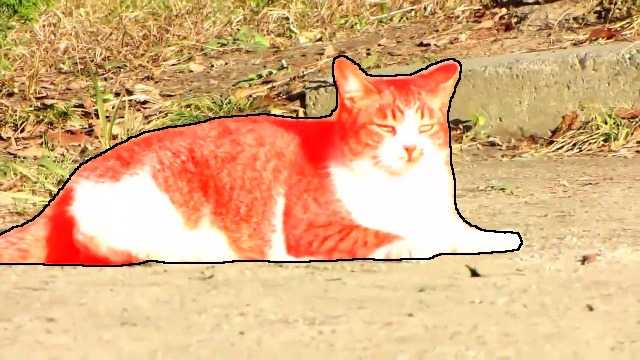} & {\footnotesize{}\hspace{-1em}}\includegraphics[width=0.15\textwidth]{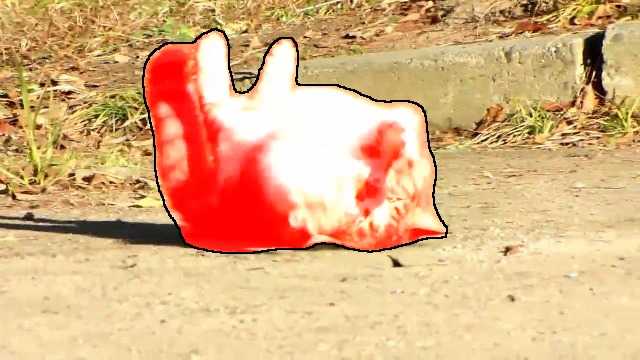}\tabularnewline
 & {\footnotesize{}\hspace{-1em}}\includegraphics[width=0.15\textwidth]{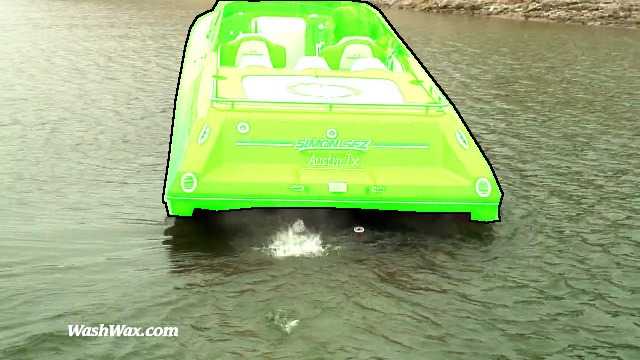} & {\footnotesize{}\hspace{-1em}}\includegraphics[width=0.15\textwidth]{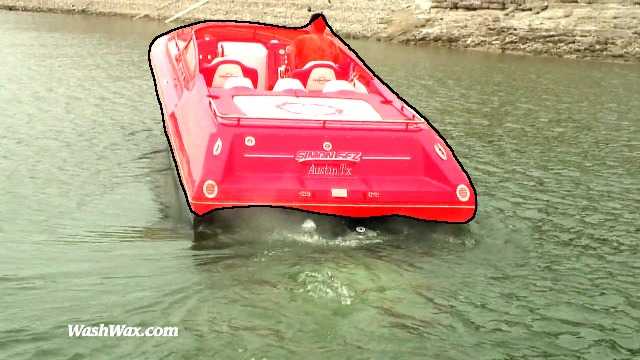} & {\footnotesize{}\hspace{-1em}}\includegraphics[width=0.15\textwidth]{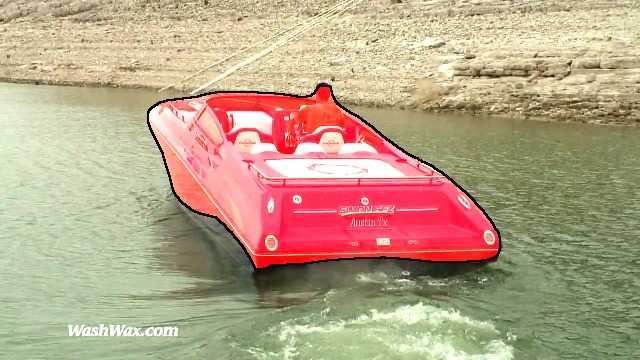} & {\footnotesize{}\hspace{-1em}}\includegraphics[width=0.15\textwidth]{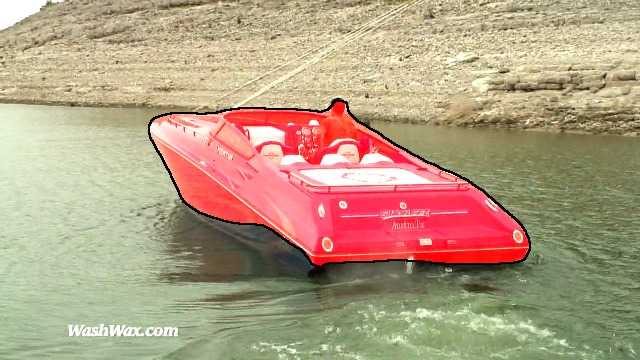} & {\footnotesize{}\hspace{-1em}}\includegraphics[width=0.15\textwidth]{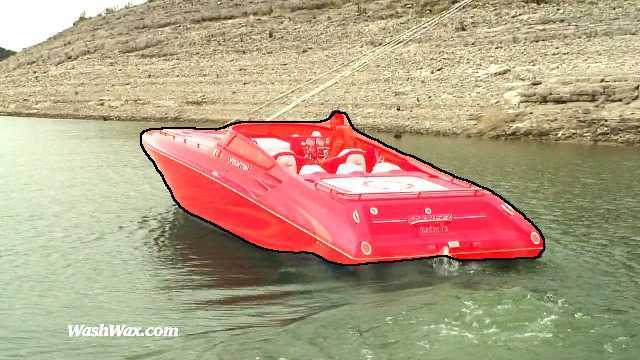} & {\footnotesize{}\hspace{-1em}}\includegraphics[width=0.15\textwidth]{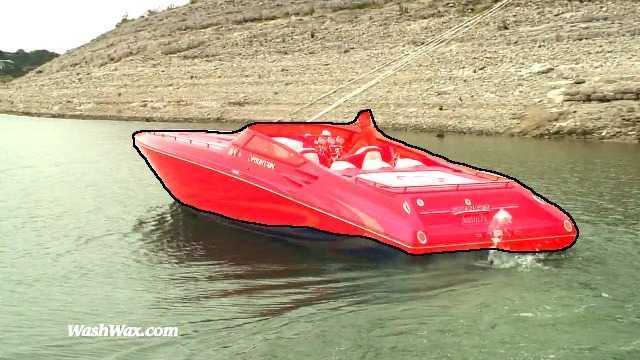}\tabularnewline
 & {\footnotesize{}\hspace{-1em}}\includegraphics[width=0.15\textwidth]{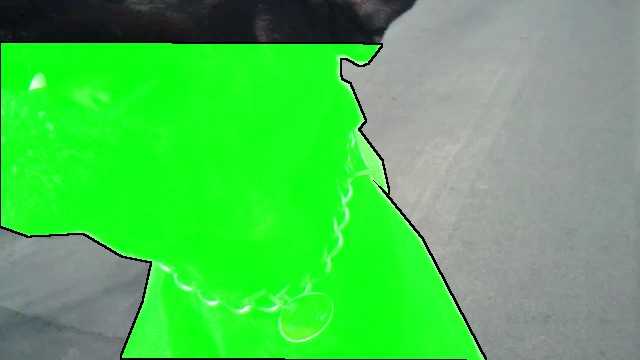} & {\footnotesize{}\hspace{-1em}}\includegraphics[width=0.15\textwidth]{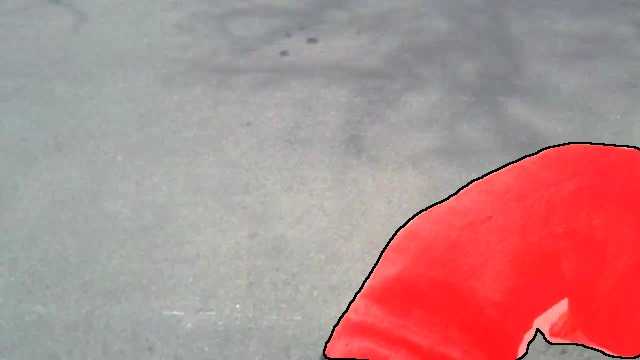} & {\footnotesize{}\hspace{-1em}}\includegraphics[width=0.15\textwidth]{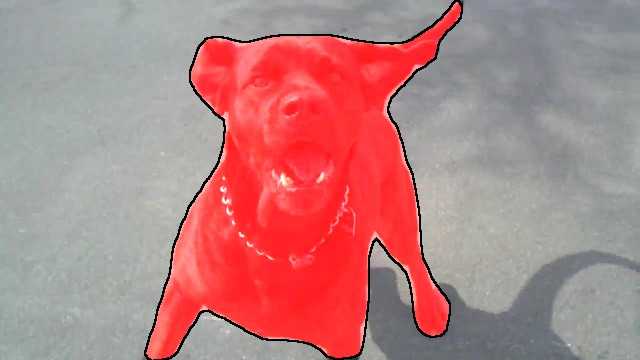} & {\footnotesize{}\hspace{-1em}}\includegraphics[width=0.15\textwidth]{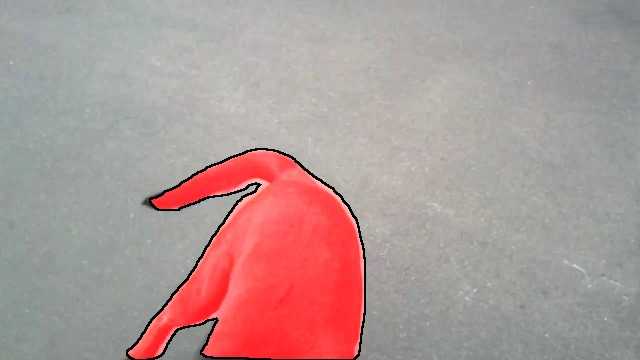} & {\footnotesize{}\hspace{-1em}}\includegraphics[width=0.15\textwidth]{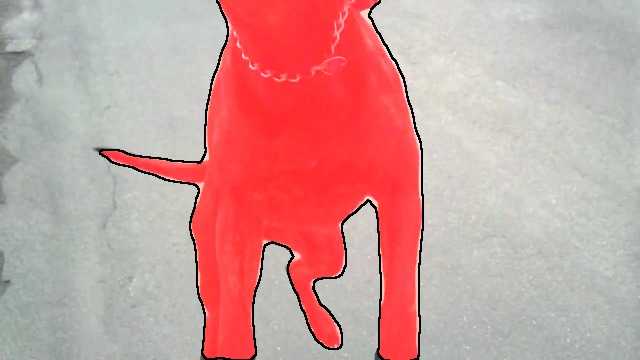} & {\footnotesize{}\hspace{-1em}}\includegraphics[width=0.15\textwidth]{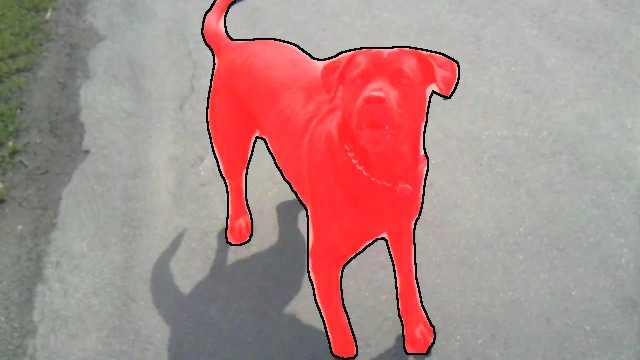}\tabularnewline
 & {\footnotesize{}\hspace{-1em}}\includegraphics[width=0.15\textwidth]{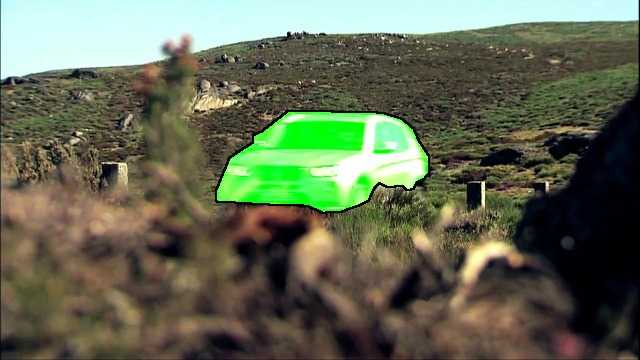} & {\footnotesize{}\hspace{-1em}}\includegraphics[width=0.15\textwidth]{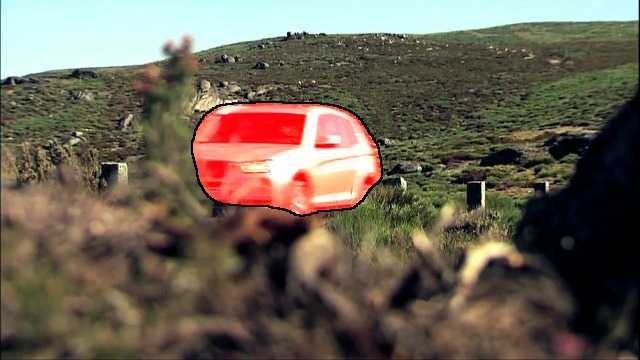} & {\footnotesize{}\hspace{-1em}}\includegraphics[width=0.15\textwidth]{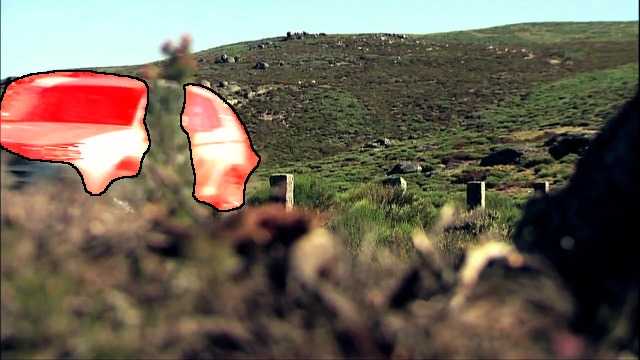} & {\footnotesize{}\hspace{-1em}}\includegraphics[width=0.15\textwidth]{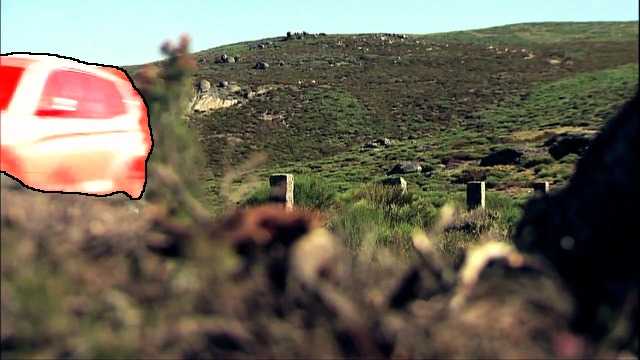} & {\footnotesize{}\hspace{-1em}}\includegraphics[width=0.15\textwidth]{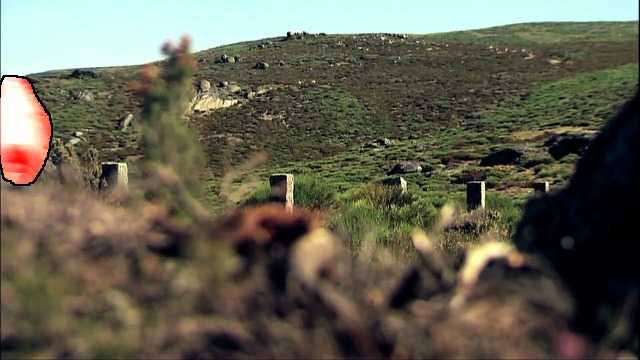} & {\footnotesize{}\hspace{-1em}}\includegraphics[width=0.15\textwidth]{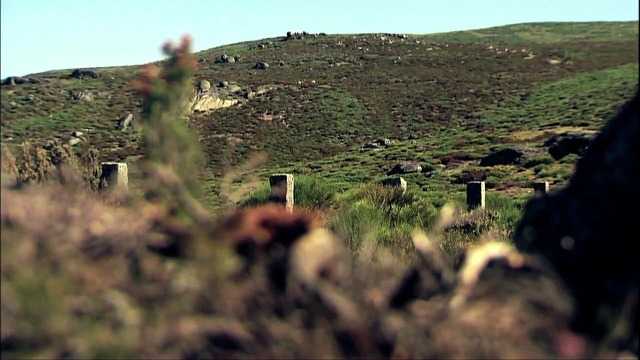}\tabularnewline
 &  &  &  &  &  & \tabularnewline
\multirow{4}{*}{\begin{turn}{90}
\hspace{-5em}$\text{SegTrack}_{\text{v2}}$\hspace{-3em}
\end{turn}} & {\footnotesize{}\hspace{-1em}}\includegraphics[width=0.15\textwidth]{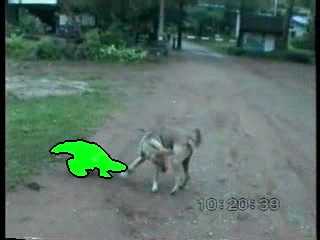} & {\footnotesize{}\hspace{-1em}}\includegraphics[width=0.15\textwidth]{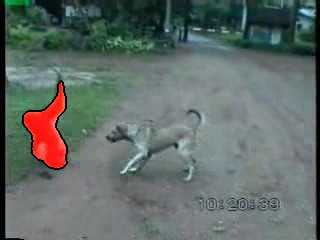} & {\footnotesize{}\hspace{-1em}}\includegraphics[width=0.15\textwidth]{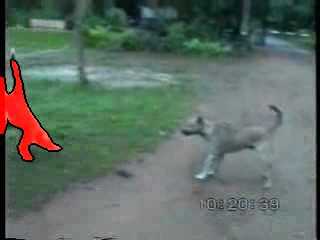} & {\footnotesize{}\hspace{-1em}}\includegraphics[width=0.15\textwidth]{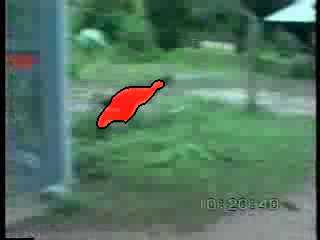} & {\footnotesize{}\hspace{-1em}}\includegraphics[width=0.15\textwidth]{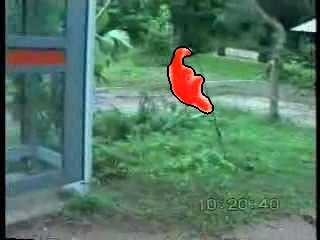} & {\footnotesize{}\hspace{-1em}}\includegraphics[width=0.15\textwidth]{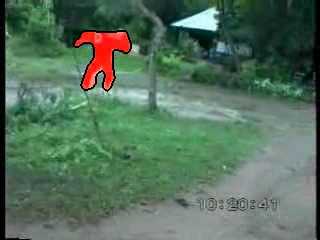}\tabularnewline
 & {\footnotesize{}\hspace{-1em}}\includegraphics[width=0.15\textwidth]{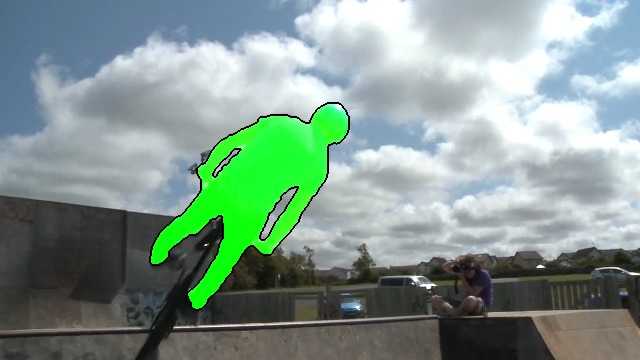} & {\footnotesize{}\hspace{-1em}}\includegraphics[width=0.15\textwidth]{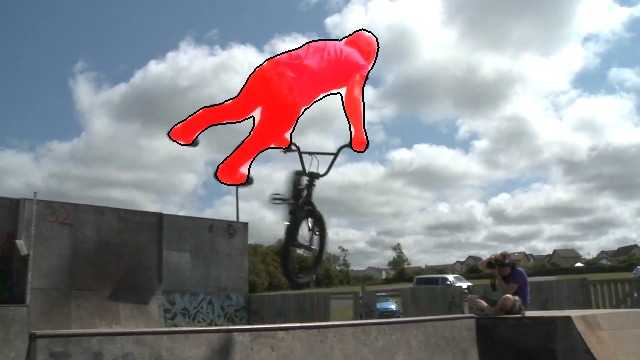} & {\footnotesize{}\hspace{-1em}}\includegraphics[width=0.15\textwidth]{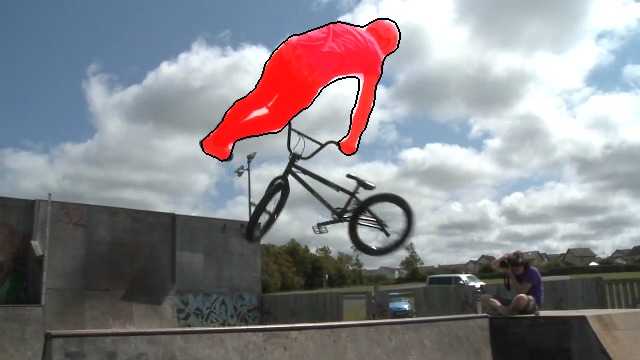} & {\footnotesize{}\hspace{-1em}}\includegraphics[width=0.15\textwidth]{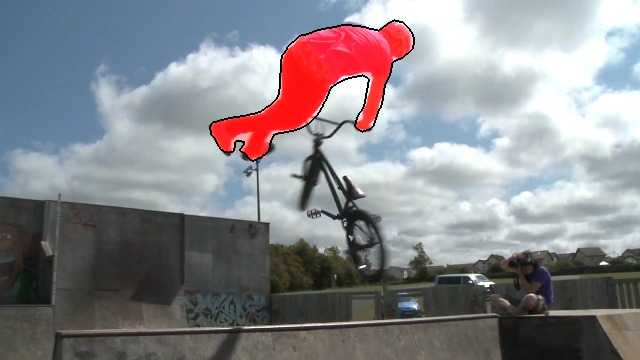} & {\footnotesize{}\hspace{-1em}}\includegraphics[width=0.15\textwidth]{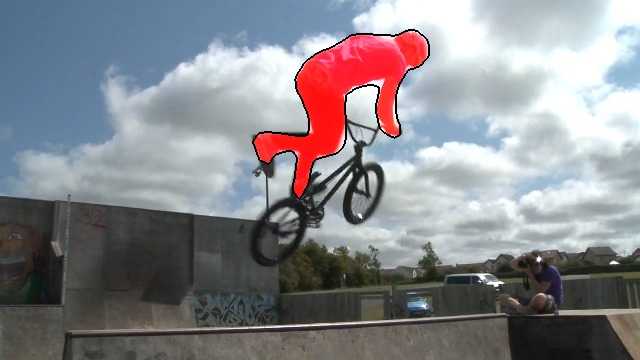} & {\footnotesize{}\hspace{-1em}}\includegraphics[width=0.15\textwidth]{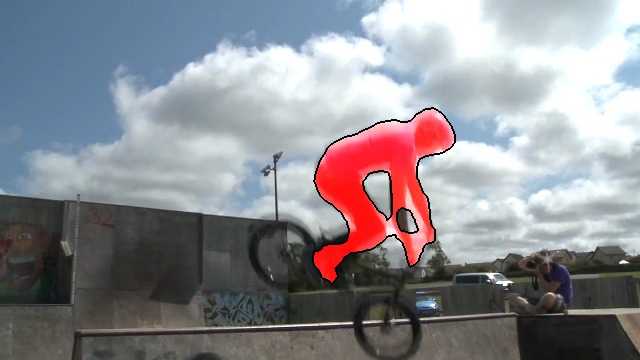}\tabularnewline
 & {\footnotesize{}\hspace{-1em}}\includegraphics[width=0.15\textwidth]{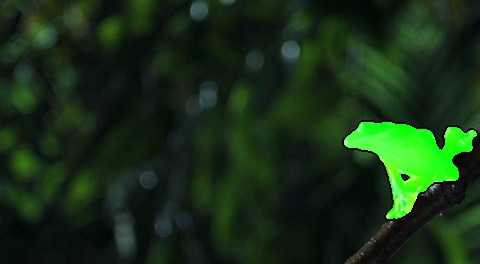} & {\footnotesize{}\hspace{-1em}}\includegraphics[width=0.15\textwidth]{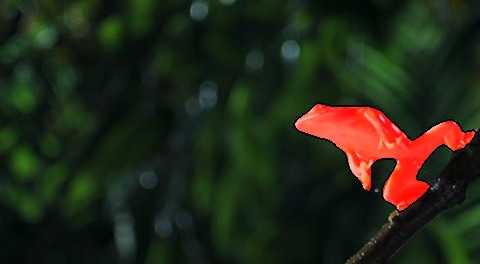} & {\footnotesize{}\hspace{-1em}}\includegraphics[width=0.15\textwidth]{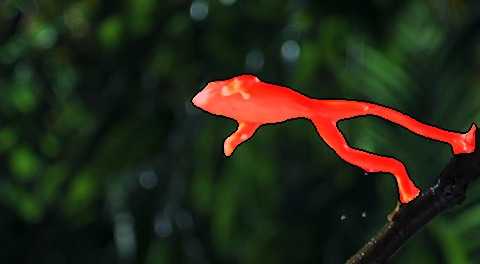} & {\footnotesize{}\hspace{-1em}}\includegraphics[width=0.15\textwidth]{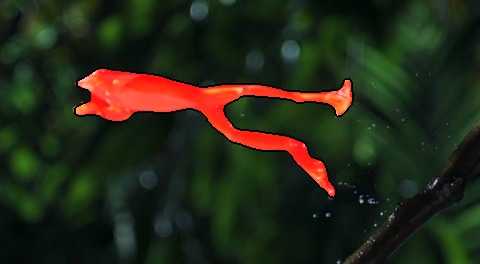} & {\footnotesize{}\hspace{-1em}}\includegraphics[width=0.15\textwidth]{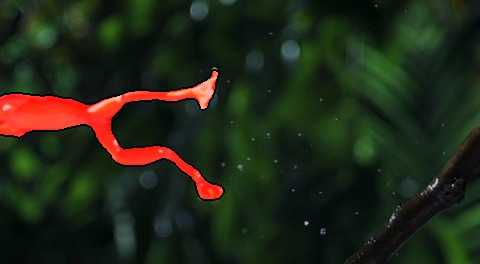} & {\footnotesize{}\hspace{-1em}}\includegraphics[width=0.15\textwidth]{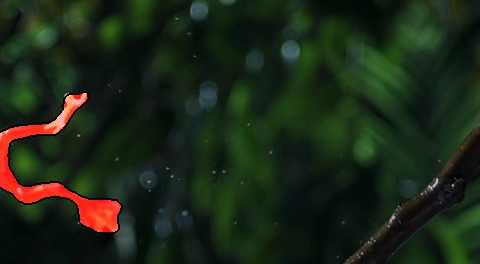}\tabularnewline
 & {\footnotesize{}\hspace{-1em}}\includegraphics[width=0.15\textwidth]{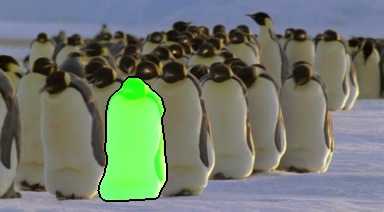} & {\footnotesize{}\hspace{-1em}}\includegraphics[width=0.15\textwidth]{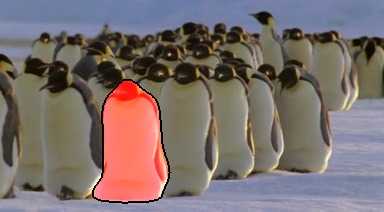} & {\footnotesize{}\hspace{-1em}}\includegraphics[width=0.15\textwidth]{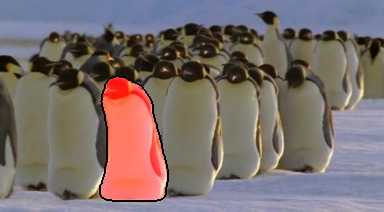} & {\footnotesize{}\hspace{-1em}}\includegraphics[width=0.15\textwidth]{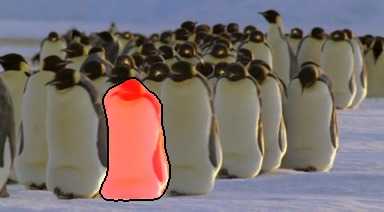} & {\footnotesize{}\hspace{-1em}}\includegraphics[width=0.15\textwidth]{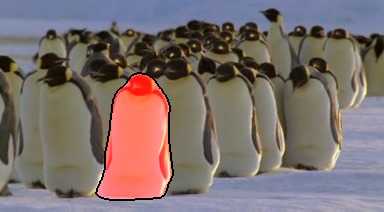} & {\footnotesize{}\hspace{-1em}}\includegraphics[width=0.15\textwidth]{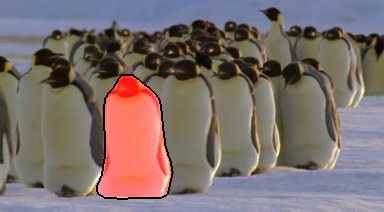}\tabularnewline
 & {\footnotesize{}\hspace{-1em}1st frame, GT segment} & {\footnotesize{}\hspace{-1em}$20\%$} & {\footnotesize{}\hspace{-1em}$40\%$} & {\footnotesize{}\hspace{-1em}$60\%$} & {\footnotesize{}\hspace{-1em}$80\%$} & {\footnotesize{}\hspace{-1em}$100\%$}\tabularnewline
\end{tabular}\hfill{}
\par\end{centering}
\caption{\label{fig:qualitative-results}LucidTracker single object segmentation
qualitative results. Frames sampled along the video duration (e.g.
$50\%$: video middle point). Our model is robust to various challenges,
such as view changes, fast motion, shape deformations, and out-of-view
scenarios.}
\vspace{0em}
\end{figure*}

\paragraph{Training details}

For training all the models we use SGD with mini-batches of $10$
images and a fixed learning policy with initial learning rate of $10^{-3}$.
The momentum and weight decay are set to $0.9$ and $5\cdot10^{-4}$
, respectively.

Models using pre-training are initialized with weights trained for
image classification on ImageNet \cite{Simonyan2015Iclr}. We then
train per-\-dataset for 40k iterations with the RGB\-+\-Mask branch
$f_{\mathcal{I}}$ and for 20k iterations for the Flow\-+\-Mask
$f_{\mathcal{F}}$ branch. When using a single stream architecture
(Section \ref{subsubsec:convnet-arch}), we use 40k iterations.

\textcolor{black}{Models without ImageNet pre-training are initialized using the Xavier (also known as Glorot) random weight initialization strategy \cite{Glorot10understandingthe}. (The weights are initialized as random draws from a truncated normal distribution with zero mean and standard deviation calculated based on the number of input and output units in the weight tensor, see \cite{Glorot10understandingthe} for details).}  The per-dataset training
needs to be longer, using 100k iterations for the $f_{\mathcal{I}}$
branch and 40k iterations for the $f_{\mathcal{F}}$ branch.

For per-video fine-tuning 2k iterations are used for $f_{\mathcal{I}}$.
To keep computing cost lower, the $f_{\mathcal{F}}$ branch is kept
fix across videos. 

All training parameters are chosen based on $\text{DAVIS}_{\text{16}}$
results. We use identical parameters on YouTubeObjects and $\text{SegTrack}_{\text{v2}}$,
showing the generalization of our approach.

It takes \textasciitilde{}3.5h to obtain each per-video model, including
data generation, per-dataset training, per-video fine-tuning and per-dataset
grid search of CRF parameters (averaged over $\text{DAVIS}_{\text{16}}$,
amortising the per-dataset training time over all videos). At test
time our $\text{LucidTracker}$\texttt{ }runs at \textasciitilde{}5s
per frame, including the optical flow estimation with FlowNet2.0 \cite{IlgCVPR17}
(\textasciitilde{}0.5s) and CRF post-processing \cite{Kraehenbuehl2011Nips}
(\textasciitilde{}2s). 

\subsection{\label{subsec:Key-results}Key results}

Table \ref{tab:comparative-result} presents our main result and compares
it to previous work. Our full system, $\mathtt{LucidTracker}$, provides
the best video segmentation quality across three datasets while being trained
on each dataset using only one frame per video ($50$ frames for $\text{DAVIS}_{\text{16}}$,
$126$ for YouTubeObjects, $24$ for $\text{SegTrack}_{\text{v2}}$),
which is \textcolor{black}{$20\times\negmedspace\sim\negmedspace1000\times$} less than
the top competing methods. Ours is the first method to reach $>75\ \text{mIoU}$
on all three datasets.

\paragraph{Oracles and baselines}

Grabcut oracle computes grabcut \cite{Rother2004SiggraphGrabcut}
using the ground truth bounding boxes (box oracle). This oracle indicates
that on the considered datasets separating foreground from background
is not easy, even if a perfect box-level tracker was available.\\
We provide three additional baselines. ``Saliency'' corresponds
to using the generic (training-free) saliency method EQCut \cite{Aytekin2015IcipEQCut}
over the RGB image $\mathcal{I}_{t}$. ``Flow saliency'' does the
same, but over the optical flow magnitude $\left\Vert \mathcal{F}_{t}\right\Vert $.
Results indicate that the objects being tracked are not particularly
salient in the image. On $\text{DAVIS}_{\text{16}}$ motion saliency
is a strong signal but not on the other two datasets. Saliency methods
ignore the first frame annotation provided for the task.
We also consider the ``Mask warping'' baseline which uses optical
flow to propagate the mask estimate from $t$ to $t+1$ via simple
warping $M_{t}=w(M_{t-1},\mathcal{\,F}_{t})$. The bad results of
this baseline indicate that the high quality flow \cite{IlgCVPR17}
that we use is by itself insufficient to solve the video object segmentation task,
and that indeed our proposed convnet does the heavy lifting.

The large fluctuation of the relative baseline results across the
three datasets empirically confirms that each of them presents unique
challenges.
\begin{table}
\setlength{\tabcolsep}{0.1em} 
\renewcommand{\arraystretch}{1.2}
\begin{centering}
\vspace{0em}
\hspace*{-1.2em}%
\begin{tabular}{cccc|ccc}
\multirow{2}{*}{} & \multirow{2}{*}{{\footnotesize{}Method}} & {\footnotesize{}\# training} & {\footnotesize{}Flow} & \multicolumn{3}{c}{{\footnotesize{}Dataset, mIoU}}\tabularnewline
 &  & {\footnotesize{}images} & {\footnotesize{}$\mathcal{F}$} & {\footnotesize{}$\text{DAVIS}_{\text{16}}$} & {\footnotesize{}YoutbObjs} & {\footnotesize{}$\mbox{SegTrck}_{\mbox{v2}}$}\tabularnewline
\hline 
\hline 
\multirow{2}{*}{\begin{turn}{90}
\end{turn}} & {\small{}Box oracle \cite{Khoreva2017CvprMaskTrack}} & {\small{}0} & \textbf{\textcolor{black}{\scriptsize{}\XSolidBrush{}}} & {\small{}45.1} & {\small{}55.3} & {\small{}56.1}\tabularnewline
 & {\small{}Grabcut oracle \cite{Khoreva2017CvprMaskTrack}} & {\small{}0} & \textbf{\textcolor{black}{\scriptsize{}\XSolidBrush{}}} & {\small{}67.3} & {\small{}67.6} & {\small{}74.2}\tabularnewline
\hline 
\multirow{8}{*}{\begin{turn}{90}
{\footnotesize{}}%
\begin{tabular}{c}
{\footnotesize{}Ignores 1st frame}\tabularnewline
{\footnotesize{}annotation}\tabularnewline
\end{tabular}{\footnotesize{} }
\end{turn}} & {\small{}Saliency} & {\small{}0} & \textbf{\textcolor{black}{\scriptsize{}\XSolidBrush{}}} & {\small{}32.7} & {\small{}40.7} & {\small{}22.2}\tabularnewline
 & {\small{}NLC \cite{Faktor2014Bmvc}} & {\small{}0} & \textbf{\textcolor{black}{\scriptsize{}\Checkmark{}}} & {\small{}64.1} & {\small{}-} & {\small{}-}\tabularnewline
 & {\small{}TRS \cite{Xiao2016Cvpr}} & {\small{}0} & \textbf{\textcolor{black}{\scriptsize{}\Checkmark{}}} & {\small{}-} & {\small{}-} & {\small{}69.1}\tabularnewline
 & {\small{}MP-Net \cite{Tokmakov2016Arxiv}} & {\small{}\textasciitilde{}22.5k} & \textbf{\textcolor{black}{\scriptsize{}\Checkmark{}}} & {\small{}69.7} & {\small{}-} & {\small{}-}\tabularnewline
 & {\small{}Flow saliency} & {\small{}0} & \textbf{\textcolor{black}{\scriptsize{}\Checkmark{}}} & {\small{}70.7} & {\small{}36.3} & {\small{}35.9}\tabularnewline
 & {\small{}FusionSeg \cite{Jain2017ArxivFusionSeg}} & {\small{}\textasciitilde{}95k} & \textbf{\textcolor{black}{\scriptsize{}\Checkmark{}}} & {\small{}71.5} & {\small{}67.9} & {\small{}-}\tabularnewline
 & \textcolor{black} {\small{}LVO \cite{TokmakovAS17}} & {\small{}\textasciitilde{}35k} & \textbf{\textcolor{black}{\scriptsize{}\Checkmark{}}} & \textit{\small{}75.9} & {\small{}-} & {\small{}57.3}\tabularnewline
 & \textcolor{black}{\small{}PDB \cite{Song_2018_ECCV}} & {\small{}\textasciitilde{}18k} & \textbf{\textcolor{black}{\scriptsize{}\XSolidBrush{}}} & \textit{\small{}77.2} & {\small{}-} & {\small{}-}\tabularnewline
\hline 
\multirow{13}{*}{\begin{turn}{90}
{\footnotesize{}}%
\begin{tabular}{c}
{\footnotesize{}Uses 1st frame}\tabularnewline
{\footnotesize{}annotation}\tabularnewline
\end{tabular}{\footnotesize{} }
\end{turn}} & {\small{}Mask warping} & {\small{}0} & \textbf{\textcolor{black}{\scriptsize{}\Checkmark{}}} & {\small{}32.1} & {\small{}43.2} & {\small{}42.0}\tabularnewline
 & {\small{}FCP \cite{Perazzi2015Iccv}} & {\small{}0} & \textcolor{black}{\scriptsize{}\Checkmark{}} & {\small{}63.1} & {\small{}-} & {\small{}-}\tabularnewline
 & {\small{}BVS \cite{Maerki2016Cvpr}} & {\small{}0} & \textcolor{black}{\scriptsize{}\XSolidBrush{}} & {\small{}66.5} & {\small{}59.7} & {\small{}58.4}\tabularnewline
 & {\small{}N15 \cite{Nagaraja2015Iccv}} & {\small{}0} & \textcolor{black}{\scriptsize{}\Checkmark{}} & {\small{}-} & {\small{}-} & {\small{}69.6}\tabularnewline
 & {\small{}ObjFlow \cite{Tsai2016Cvpr}} & {\small{}0} & \textcolor{black}{\scriptsize{}\Checkmark{}} & {\small{}71.1} & {\small{}70.1} & {\small{}67.5}\tabularnewline
 & {\small{}STV \cite{Wang2017ArxivSTV}} & {\small{}0} & \textcolor{black}{\scriptsize{}\Checkmark{}} & {\small{}73.6} & {\small{}-} & {\small{}-}\tabularnewline
 & {\small{}VPN \cite{Jampani2016Arxiv}} & {\small{}\textasciitilde{}2.3k} & \textcolor{black}{\scriptsize{}\XSolidBrush{}} & \textit{\textcolor{black}{\small{}75.0}} & {\small{}-} & {\small{}-}\tabularnewline
 & {\small{}OSVOS \cite{Caelles2017Cvpr}} & {\small{}\textasciitilde{}2.3k} & \textcolor{black}{\scriptsize{}\XSolidBrush{}} & \textit{\textcolor{black}{\small{}79.8}} & {\small{}72.5} & {\small{}65.4}\tabularnewline
 & {\small{}MaskTrack \cite{Khoreva2017CvprMaskTrack}} & {\small{}\textasciitilde{}11k} & \textcolor{black}{\scriptsize{}\Checkmark{}} & {\small{}80.3} & {\small{}72.6} & {\small{}70.3\arrayrulecolor{lightgray}}\tabularnewline
 & \textcolor{black} {\small{}PReMVOS \cite{Luiten18ACCV}} & {\small{}\textasciitilde{}145k} & \textcolor{black}{\scriptsize{}\Checkmark{}} & \textit{\textcolor{black}{\small{}84.9}} & {\small{}-} & {\small{}-}\tabularnewline 
& \textcolor{black} {\small{}OnAVOS \cite{Voigtlaender2017OnlineAO}} & {\small{}\textasciitilde{}120k} & \textcolor{black}{\scriptsize{}\XSolidBrush{}} & \textit{\textcolor{black}{\small{}86.1}} & {\small{}-} & {\small{}-}\tabularnewline 
& \textcolor{black} {\small{}VideoGCRF \cite{Chandra2018DeepSR}} & {\small{}\textasciitilde{}120k} & \textcolor{black}{\scriptsize{}\XSolidBrush{}} & \textit{\textcolor{black}{\small{}86.5}} & {\small{}-} & {\small{}-}\tabularnewline 

\cline{2-7} 
 & {\small{}\arrayrulecolor{black}}$\text{LucidTracker}$ & \textbf{\small{}24\textasciitilde{}126} & \textcolor{black}{\scriptsize{}\Checkmark{}} & \textcolor{black}{\textbf{\small{}{86.6}}} & \textbf{\small{}77.3} & \textbf{\small{}78.0}\tabularnewline
\end{tabular}
\par\end{centering}
\caption{\label{tab:comparative-result}Comparison of video object segmentation results
across three datasets. Numbers in italic are reported on subsets
of $\text{DAVIS}_{\text{16}}$. Our $\text{LucidTracker}$ consistently
improves over previous results, see Section \ref{subsec:Key-results}.}
\vspace{0em}
\end{table}

\paragraph{Comparison}

Compared to flow propagation methods such as BVS, N15, ObjFlow, and
STV, we obtain better results because we build per-video a stronger
appearance model of the tracked object (embodied in the fine-tuned
model). Compared to convnet learning methods such as VPN, OSVOS, MaskTrack, \textcolor{black}{OnAVOS},
we require significantly less training data, yet obtain better results. 

Figure \ref{fig:qualitative-results} provides qualitative results
of $\mathtt{LucidTracker}$ across three different datasets. Our system
is robust to various challenges present in videos. It handles well
camera view changes, fast motion, object shape deformation, out-of-view
scenarios, multiple similar looking objects and even low quality video\textit{\emph{.
We provide a detailed error analysis in section \ref{subsec:Error-analysis}.}}

\paragraph{Conclusion}

We show that top results can be obtained while using less training
data. This shows that our lucid dreams leverage the available training
data better. We report top results for this task while using only
$24\negmedspace\sim\negmedspace126$ training frames. 

\subsection{\label{subsec:Ablation-study}Ablation studies}

In this section we explore in more details how the different ingredients
contribute to our results.\vspace{0em}

\subsubsection{\label{subsubsec:ablation-training}Effect of training modalities}

Table \ref{tab:ablation-training} compares the effect of different
ingredients in the $\mathtt{Lucid}\mathtt{Tracker}^{-}$ training.
Results are obtained using RGB and flow, with warping, no CRF, and no temporal coherency; {\small{}$M_{t}\negmedspace=\negmedspace f\left(\mathcal{I}_{t},w(M_{t-1},\mathcal{F}_{t})\right)$}.

\paragraph{Training from a single frame}

In the bottom row (\textquotedbl{}only per-video tuning\textquotedbl{}),
the model is trained per-video without ImageNet pre-training nor per-dataset
training, i.e. using a \emph{single annotated training frame}. Our
network is based on VGG16 \cite{Chen2016ArxivDeeplabv2} and contains
$\sim\negmedspace20M$ parameters, all effectively learnt from a single
annotated image that is augmented to become $2.5k$ training samples
(see Section \ref{sec:Lucid-data-dreaming}). Even with such minimal
amount of training data, we still obtain a surprisingly good performance
(compare $80.5$ on $\text{DAVIS}_{\text{16}}$ to others in Table
\ref{tab:comparative-result}). This shows how effective is, by itself,
the proposed training strategy based on lucid dreaming of the data.

\begin{table}
\setlength{\tabcolsep}{0.1em}
\renewcommand{\arraystretch}{1.3}
\begin{centering}
\vspace{0em}
\begin{tabular}{c|ccc|ccc}
\multirow{2}{*}{{\scriptsize{}Variant}} & \multirow{2}{*}{{\scriptsize{}}%
\begin{tabular}{c}
{\scriptsize{}ImgNet}\tabularnewline
{\scriptsize{}pre-train.}\tabularnewline
\end{tabular}} & \multirow{2}{*}{{\scriptsize{}}%
\begin{tabular}{c}
{\scriptsize{}per-dataset}\tabularnewline
{\scriptsize{}training}\tabularnewline
\end{tabular}} & \multirow{2}{*}{{\scriptsize{}}%
\begin{tabular}{c}
{\scriptsize{}per-video}\tabularnewline
{\scriptsize{}fine-tun.}\tabularnewline
\end{tabular}} & \multicolumn{3}{c}{{\scriptsize{}Dataset, mIoU}}\tabularnewline
 &  &  &  & {\scriptsize{}$\text{DAVIS}_{\text{16}}$} & {\scriptsize{}YoutbObjs} & {\scriptsize{}$\mbox{SegTrck}_{\mbox{v2}}$}\tabularnewline
\hline 
\hline 
{\footnotesize{}LucidTracker}\textcolor{black}{\small{}$^{-}$} & \textcolor{black}{\scriptsize{}\Checkmark{}} & \textcolor{black}{\scriptsize{}\Checkmark{}} & \textcolor{black}{\scriptsize{}\Checkmark{}} & \textbf{\small{}83.7} & \textbf{\small{}76.2} & \textbf{\small{}76.8}\tabularnewline
{\footnotesize{}(no ImgNet)} & \textcolor{black}{\scriptsize{}\XSolidBrush{}} & \textcolor{black}{\scriptsize{}\Checkmark{}} & \textcolor{black}{\scriptsize{}\Checkmark{}} & {\small{}82.0} & {\small{}74.3} & {\small{}71.2}\tabularnewline
\hline 
\multirow{2}{*}{{\footnotesize{}}%
\begin{tabular}{c}
{\footnotesize{}No per-video}\tabularnewline
{\footnotesize{}tuning}\tabularnewline
\end{tabular}} & \textcolor{black}{\scriptsize{}\Checkmark{}} & \textcolor{black}{\scriptsize{}\Checkmark{}} & \textcolor{black}{\scriptsize{}\XSolidBrush{}} & {\small{}82.7} & {\small{}72.3} & {\small{}71.9}\tabularnewline
 & \textcolor{black}{\scriptsize{}\XSolidBrush{}} & \textcolor{black}{\scriptsize{}\Checkmark{}} & \textcolor{black}{\scriptsize{}\XSolidBrush{}} & {\small{}78.4} & {\small{}69.7} & {\small{}68.2\arrayrulecolor{lightgray}}\tabularnewline
\hline 
\multirow{2}{*}{{\small{}\arrayrulecolor{black}}{\footnotesize{}}%
\begin{tabular}{c}
{\footnotesize{}Only per-}\tabularnewline
{\footnotesize{}-video tuning}\tabularnewline
\end{tabular}} & \textcolor{black}{\scriptsize{}\Checkmark{}} & \textcolor{black}{\scriptsize{}\XSolidBrush{}} & \textcolor{black}{\scriptsize{}\Checkmark{}} & {\small{}79.4} & {\small{}-} & {\small{}70.4}\tabularnewline
 & \textcolor{black}{\scriptsize{}\XSolidBrush{}} & \textcolor{black}{\scriptsize{}\XSolidBrush{}} & \textcolor{black}{\scriptsize{}\Checkmark{}} & \emph{\small{}80.5} & {\small{}-} & \textit{\small{}66.8}\tabularnewline
\end{tabular}
\par\end{centering}

\caption{\label{tab:ablation-training}Ablation study of training modalities.
ImageNet pre-training and per-video tuning provide additional improvement
over per-dataset training. Even with one frame annotation for only
per-video tuning we obtain good performance. See Section \ref{subsubsec:ablation-training}. }
\vspace{0em}
\end{table}

\paragraph{Pre-training \& fine-tuning}

We see that ImageNet pre-training does provide $2\negmedspace\sim\negmedspace5$
percent point improvement (depending on the dataset of interest; e.g.
$82.0\negmedspace\rightarrow\negmedspace83.7\ \text{mIoU}$ on $\text{DAVIS}_{\text{16}}$).
Per-video fine-tuning (after doing per-dataset training) provides
an additional $1\negmedspace\sim\negmedspace2$ percent point gain
(e.g. $82.7\negmedspace\rightarrow\negmedspace83.7\ \text{mIoU}$
on $\text{DAVIS}_{\text{16}}$). Both ingredients clear\-ly contribute
to the segmentation results.

Note that training a model using only per-video tuning takes about
one full GPU day per video sequence; making these results insightful
but not decidedly practical.

Preliminary experiments evaluating on $\text{DAVIS}_{\text{16}}$
the impact of the different ingredients of our lucid dreaming data
generation showed, depending on the exact setup, $3\negmedspace\sim\negmedspace10$
percent mIoU points fluctuations between a basic version (e.g. without
non-rigid deformations nor scene re-composi\-tion) and the full synthesis
process described in Section \ref{sec:Lucid-data-dreaming}. Having
a sophisticated data generation process directly impacts the segmentation
quality.

\paragraph{Conclusion}

Surprisingly, we discovered that per-video training from a single
annotated frame provides already much of the information needed for
the video object segmentation task. Additionally using ImageNet pre-training, and per-dataset
training, provide complementary gains. 

\subsubsection{\label{subsubsec:ablation-flow}Effect of optical flow}

Table \ref{tab:ablation-flow} shows the effect of optical flow on
$\mathtt{LucidTracker}$ results. Comparing our full system to the
\textquotedbl{}No OF\textquotedbl{} row, we see that the effect of
optical flow varies across datasets, from minor improvement in YouTubeObjects,
to major difference in $\text{SegTrack}_{\text{v2}}$. In this last
dataset, using mask warping is particularly useful too. We additionally
explored tuning the optical flow stream per-video, which resulted
in a minor improvement ($83.7\negmedspace\rightarrow\negmedspace83.9\ \text{mIoU}$
on $\text{DAVIS}_{\text{16}}$).

Our \textquotedbl{}No OF\textquotedbl{} results can be compared to
OSVOS {\cite{Caelles2017Cvpr}} which does not use optical
flow. However OSVOS uses a per-frame mask post-processing based on
a boundary detector (trained on further external data), which provides
$\sim\negmedspace2$ percent point gain. Accounting for this, our
\textquotedbl{}No OF\textquotedbl{} (and no CRF, no temporal coherency) result matches theirs
on $\text{DAVIS}_{\text{16}}$ and YouTubeObjects despite using significantly
less training data (see Table \ref{tab:comparative-result}, e.g.
$79.8-2\approx78.0$ on $\text{DAVIS}_{\text{16}}$).

Table \ref{tab:ablation-OFmethod} shows the effect of using different
optical flow estimation methods. For $\mathtt{LucidTracker}$ results,
FlowNet2.0 \cite{IlgCVPR17} was employed. We also explored using
EpicFlow \cite{EpicFlowCVPR15}, as in \cite{Khoreva2017CvprMaskTrack}.
Table \ref{tab:ablation-OFmethod} indicates that employing a robust
optical flow estimation across datasets is crucial to the performance
(FlowNet2.0 provides $\sim1.5-15$ points gain on each dataset). We
found EpicFlow to be brittle when going across different datasets,
providing improvement for $\text{DAVIS}_{\text{16}}$ and $\text{SegTrack}_{\text{v2}}$
($\sim2-5$ points gain), but underperforming for YouTubeObjects ($74.7\negmedspace\rightarrow\negmedspace71.3\ \text{mIoU}$).

\begin{table}
\setlength{\tabcolsep}{0.45em} 
\renewcommand{\arraystretch}{1.3}
\begin{centering}
\vspace{0em}
\begin{tabular}{c|ccc|ccc}
\multirow{2}{*}{{\small{}Variant}} & \multirow{2}{*}{{\small{}$\mathcal{I}$}} & \multirow{2}{*}{{\small{}$\mathcal{F}$}} & {\small{}warp.} & \multicolumn{3}{c}{{\small{}Dataset, mIoU}}\tabularnewline
 &  &  & {\small{}$w$} & {\footnotesize{}$\text{DAVIS}_{\text{16}}$} & {\footnotesize{}YoutbObjs} & {\footnotesize{}$\mbox{SegTrck}_{\mbox{v2}}$}\tabularnewline
\hline 
\hline 
\textcolor{black}{\small{}$\text{LucidTracker}$} & \textcolor{black}{\scriptsize{}\Checkmark{}} & \textcolor{black}{\scriptsize{}\Checkmark{}} & \textcolor{black}{\scriptsize{}\Checkmark{}} &  \textcolor{black}{\textbf{\small{}{86.6}}} & \textbf{\small{}77.3} & \textbf{\small{}78.0}{\small{}\arrayrulecolor{lightgray}}\tabularnewline
\hline 
{\small{}\arrayrulecolor{black}}\textcolor{black}{\small{}LucidTracker$^{-}$} & \textcolor{black}{\scriptsize{}\Checkmark{}} & \textcolor{black}{\scriptsize{}\Checkmark{}} & \textcolor{black}{\scriptsize{}\Checkmark{}} &  \textcolor{black}{\small{}83.7} & \textcolor{black}{\small{}76.2} & \textcolor{black}{\small{}76.8}\tabularnewline
\hline 
{\small{}No warping} & \textcolor{black}{\scriptsize{}\Checkmark{}} & \textcolor{black}{\scriptsize{}\Checkmark{}} & \textcolor{black}{\scriptsize{}\XSolidBrush{}} &  {\small{}82.0} & {\small{}74.6} & {\small{}70.5}\tabularnewline
{\small{}No OF} & \textcolor{black}{\scriptsize{}\Checkmark{}} & \textcolor{black}{\scriptsize{}\XSolidBrush{}} & \textcolor{black}{\scriptsize{}\XSolidBrush{}} &  {\small{}78.0} & {\small{}74.7} & {\small{}61.8}\tabularnewline
{\small{}OF only} & \textcolor{black}{\scriptsize{}\XSolidBrush{}} & \textcolor{black}{\scriptsize{}\Checkmark{}} & \textcolor{black}{\scriptsize{}\Checkmark{}} & {\small{}74.5} & {\small{}43.1} & {\small{}55.8}\tabularnewline
\end{tabular}
\par\end{centering}

\caption{\label{tab:ablation-flow}Ablation study of flow ingredients. Flow
complements image only results, with large fluctuations across datasets.
See Section \ref{subsubsec:ablation-flow}.}
\vspace{0em}
\end{table}

\begin{table}
\setlength{\tabcolsep}{0.5em}
\renewcommand{\arraystretch}{1.3}
\begin{centering}
\begin{tabular}{c|c|ccc}
\multirow{2}{*}{{\scriptsize{}Variant}} & {\footnotesize{}Optical} & \multicolumn{3}{c}{{\footnotesize{}Dataset, mIoU}}\tabularnewline
 & {\footnotesize{}flow} & {\footnotesize{}$\text{DAVIS}_{\text{16}}$} & {\footnotesize{}YoutbObjs} & {\footnotesize{}$\mbox{SegTrck}_{\mbox{v2}}$}\tabularnewline
\hline 
\hline 
\multirow{3}{*}{{\footnotesize{}LucidTracker}\textcolor{black}{\small{}$^{-}$}} & {\footnotesize{}FlowNet2.0} & \textbf{\small{}83.7} & \textbf{\small{}76.2} & \textbf{\small{}76.8}\tabularnewline
 & {\footnotesize{}EpicFlow} & {\small{}80.2} & {\small{}71.3} & {\small{}67.0}\tabularnewline
 & {\footnotesize{}No flow} & {\small{}78.0} & {\small{}74.7} & {\small{}61.8}\tabularnewline
\hline 
\multirow{3}{*}{{\footnotesize{}}%
\begin{tabular}{c}
{\footnotesize{}No ImageNet}\tabularnewline
{\footnotesize{}pre-training}\tabularnewline
\end{tabular}} & {\footnotesize{}FlowNet2.0} & {\small{}82.0} & {\small{}74.3} & {\small{}71.2}\tabularnewline
 & {\footnotesize{}EpicFlow} & {\small{}80.0} & {\small{}72.3} & {\small{}68.8}\tabularnewline
 & {\footnotesize{}No flow} & {\small{}76.7} & {\small{}71.4} & {\small{}63.0}\tabularnewline
\end{tabular}
\par\end{centering}
\caption{\label{tab:ablation-OFmethod}Effect of optical flow estimation.}
\end{table}

\paragraph{Conclusion}

The results show that flow provides a complementary signal to RGB
image only and having a robust optical flow estimation across datasets
is crucial. Despite its simplicity our fusion strategy ($f_{\mathcal{I}}+f_{\mathcal{F}}$
) provides gains on all datasets, and leads to competitive results.

\subsubsection{\label{sec:Effect-of-CRF}Effect of CRF tuning}

\begin{figure*}
\begin{centering}
\setlength{\tabcolsep}{0em} 
\renewcommand{\arraystretch}{0}
\par\end{centering}
\begin{centering}
\hfill{}%
\begin{tabular}{ccccccc}
\multirow{2}{*}{{\footnotesize{}\hspace{-1em}}\includegraphics[width=0.15\textwidth]{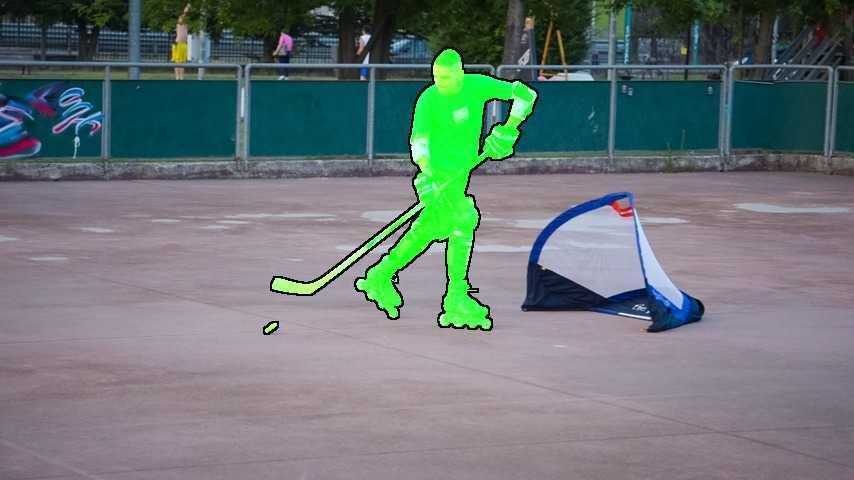}} & \begin{turn}{90}
{\footnotesize{}\hspace{1em}no CRF\hspace{-2em}}
\end{turn} & {\footnotesize{}\hspace{-1em}}\includegraphics[width=0.15\textwidth]{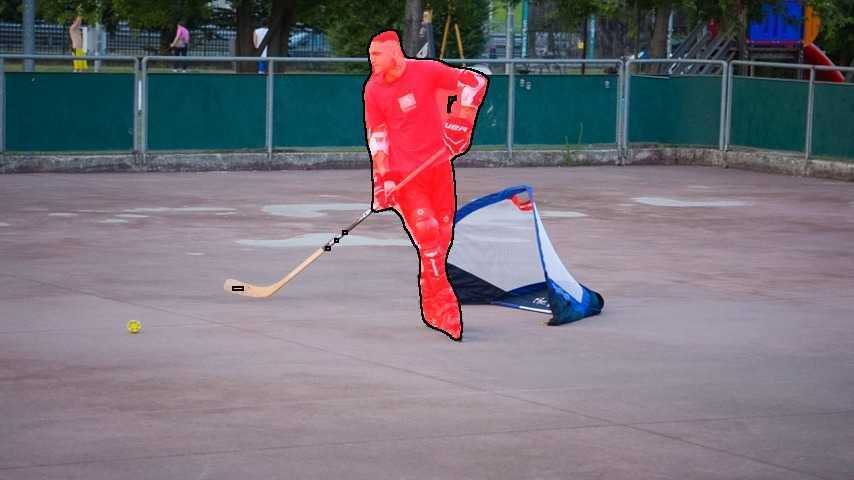} & {\footnotesize{}\hspace{-1em}}\includegraphics[width=0.15\textwidth]{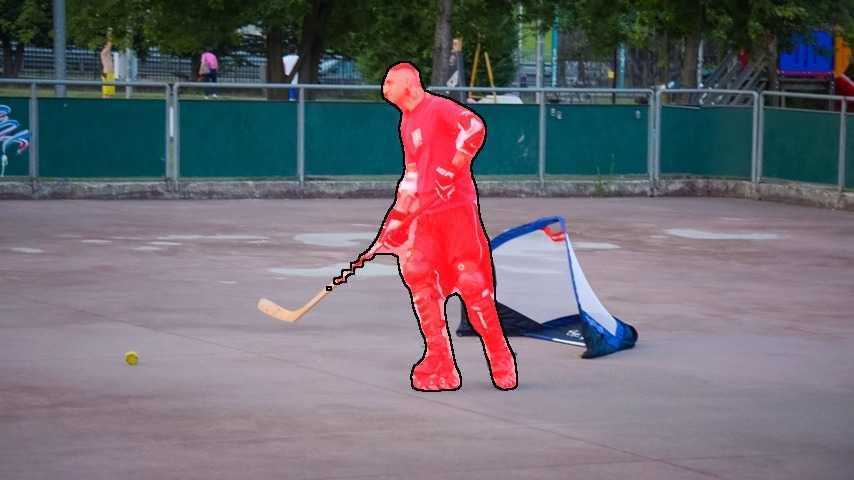} & {\footnotesize{}\hspace{-1em}}\includegraphics[width=0.15\textwidth]{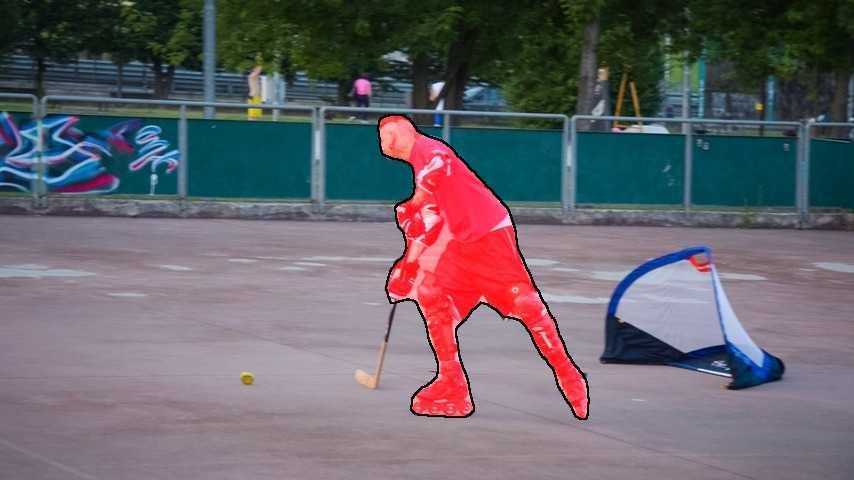} & {\footnotesize{}\hspace{-1em}}\includegraphics[width=0.15\textwidth]{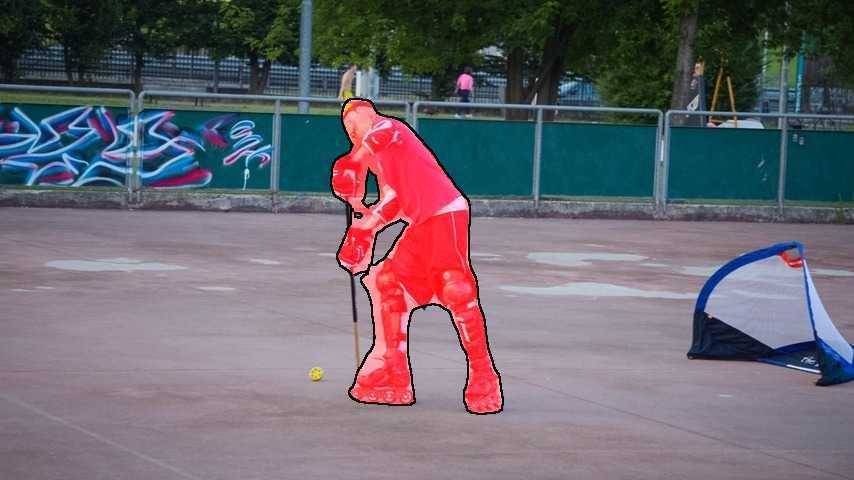} & {\footnotesize{}\hspace{-1em}}\includegraphics[width=0.15\textwidth]{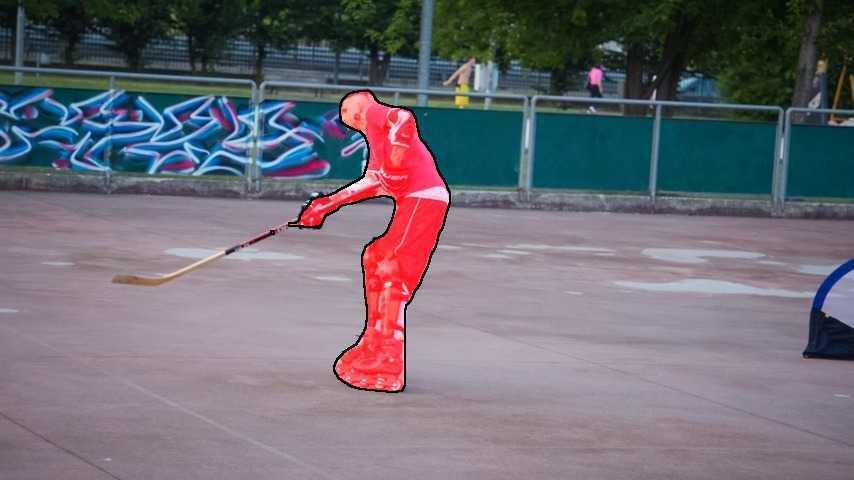}\tabularnewline
 & \begin{turn}{90}
{\footnotesize{}\hspace{0.5em}with CRF\hspace{-2em}}
\end{turn} & {\footnotesize{}\hspace{-1em}}\includegraphics[width=0.15\textwidth]{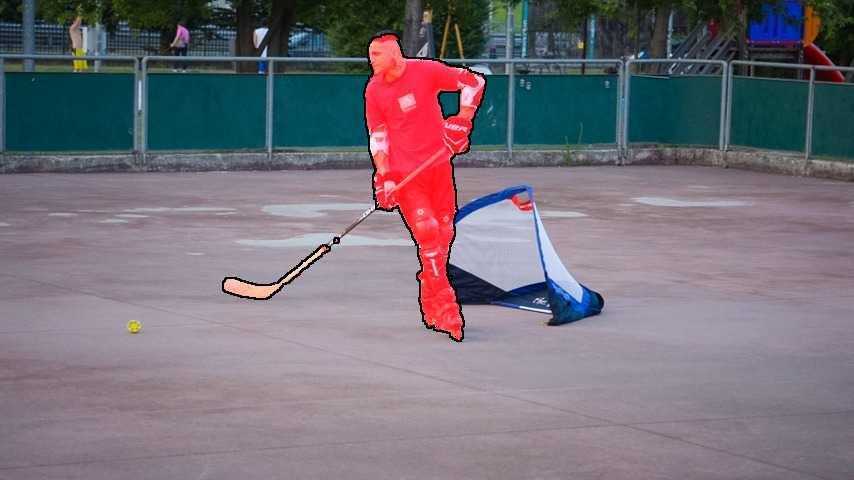} & {\footnotesize{}\hspace{-1em}}\includegraphics[width=0.15\textwidth]{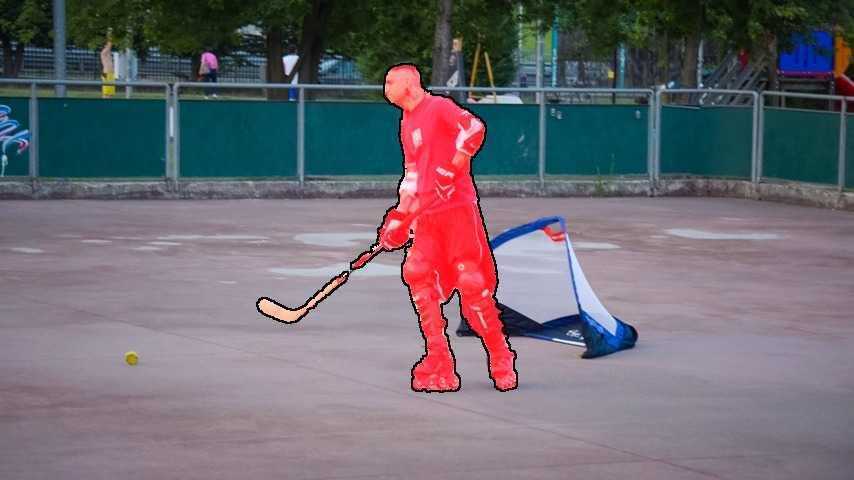} & {\footnotesize{}\hspace{-1em}}\includegraphics[width=0.15\textwidth]{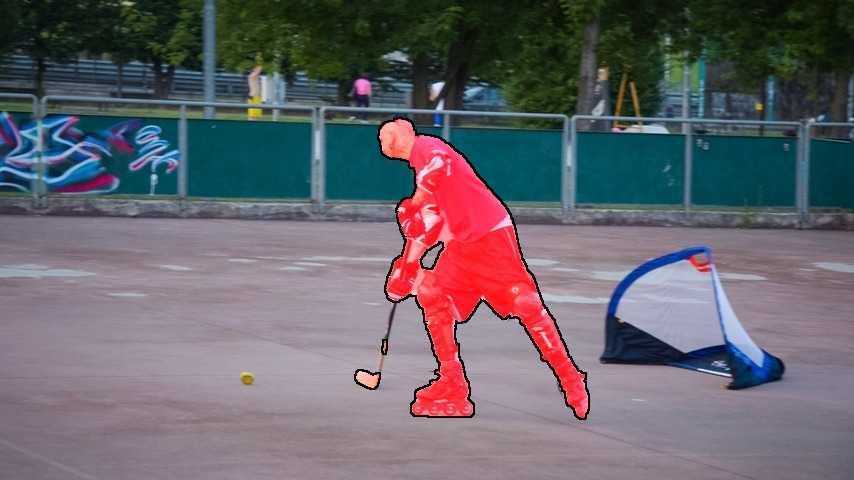} & {\footnotesize{}\hspace{-1em}}\includegraphics[width=0.15\textwidth]{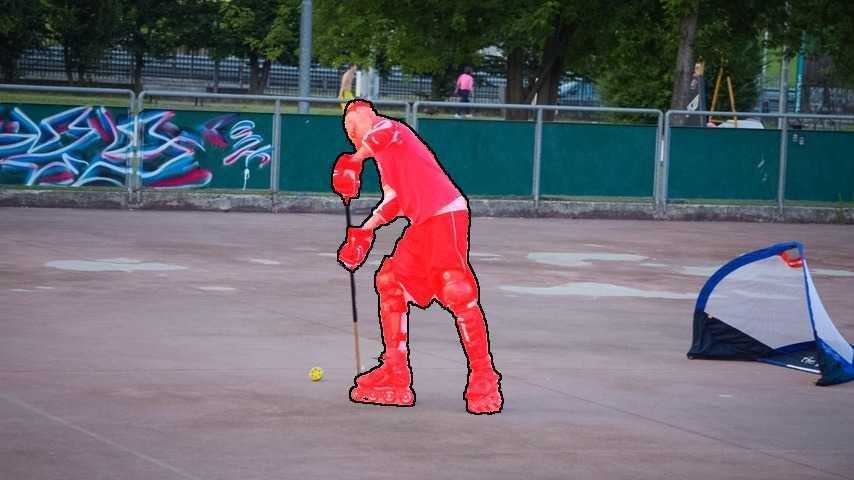} & {\footnotesize{}\hspace{-1em}}\includegraphics[width=0.15\textwidth]{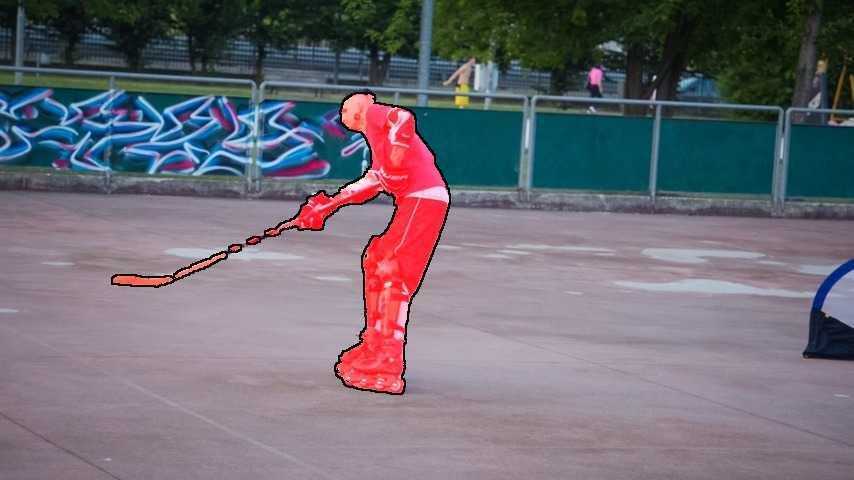}\tabularnewline
\multicolumn{7}{c}{\smallskip{}
}\tabularnewline
\multirow{2}{*}{{\footnotesize{}\hspace{-1em}}\includegraphics[width=0.15\textwidth]{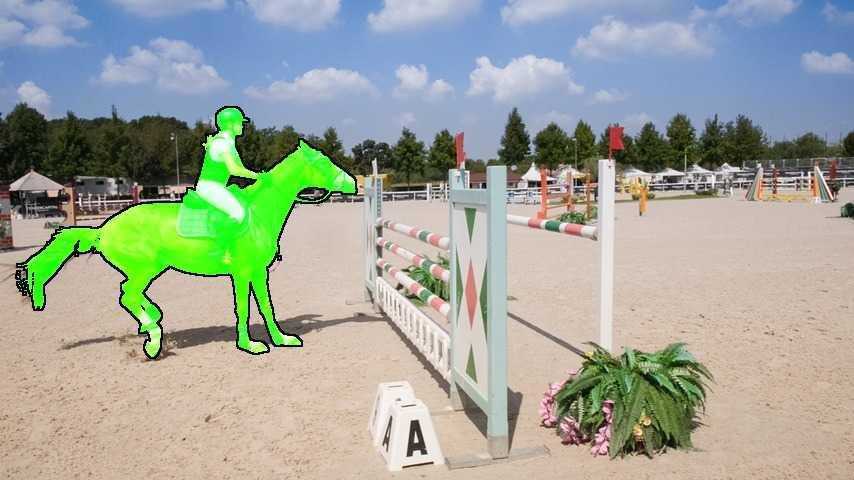}} & \begin{turn}{90}
{\footnotesize{}\hspace{1em}no CRF\hspace{-2em}}
\end{turn} & {\footnotesize{}\hspace{-1em}}\includegraphics[width=0.15\textwidth]{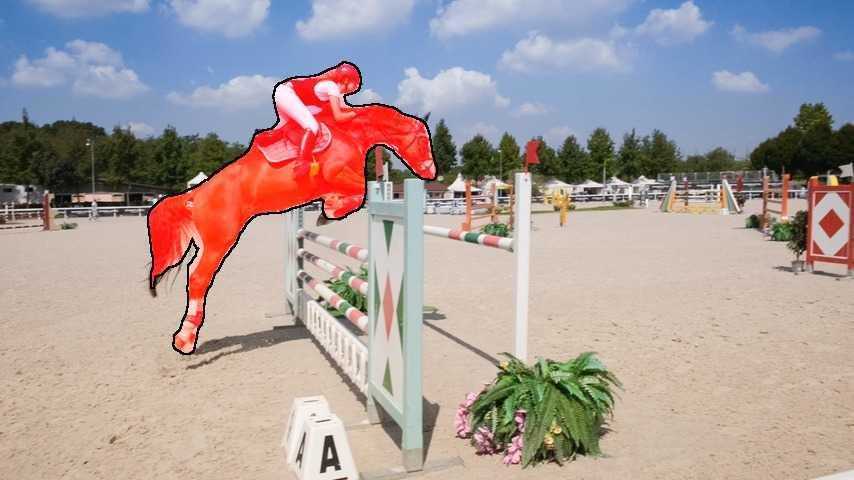} & {\footnotesize{}\hspace{-1em}}\includegraphics[width=0.15\textwidth]{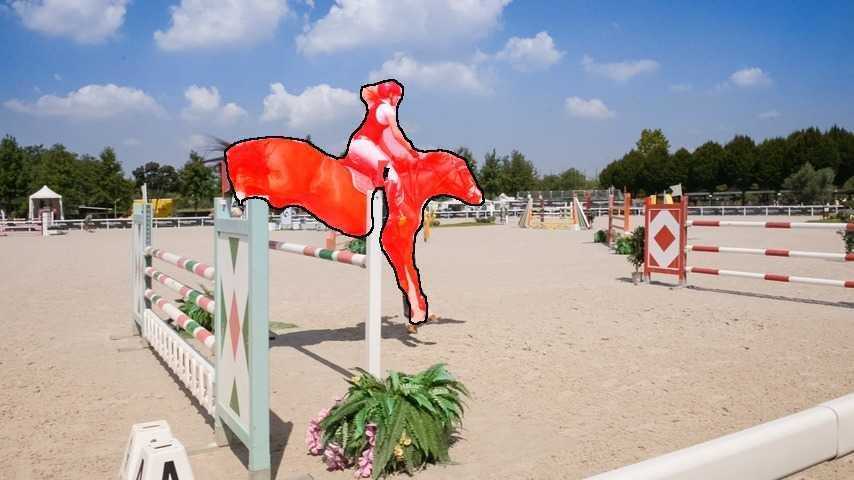} & {\footnotesize{}\hspace{-1em}}\includegraphics[width=0.15\textwidth]{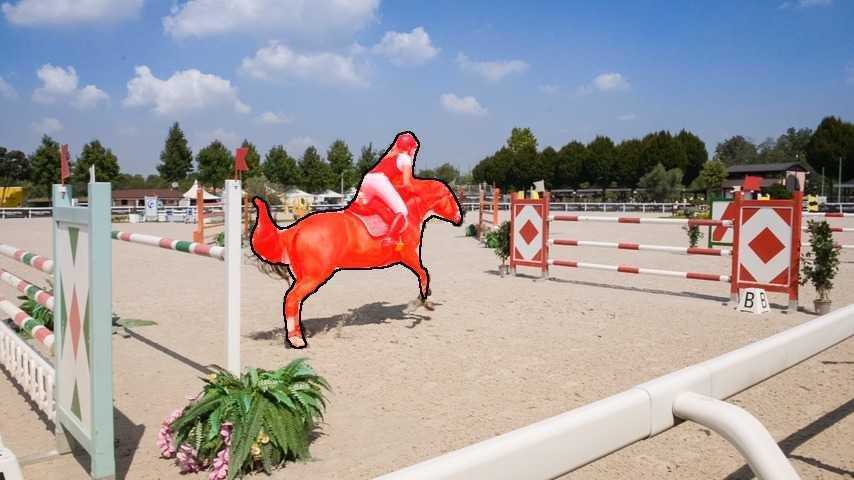} & {\footnotesize{}\hspace{-1em}}\includegraphics[width=0.15\textwidth]{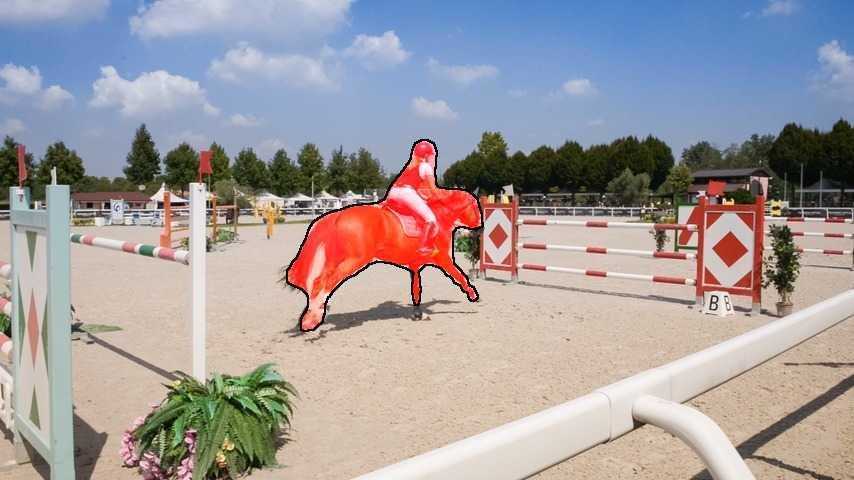} & {\footnotesize{}\hspace{-1em}}\includegraphics[width=0.15\textwidth]{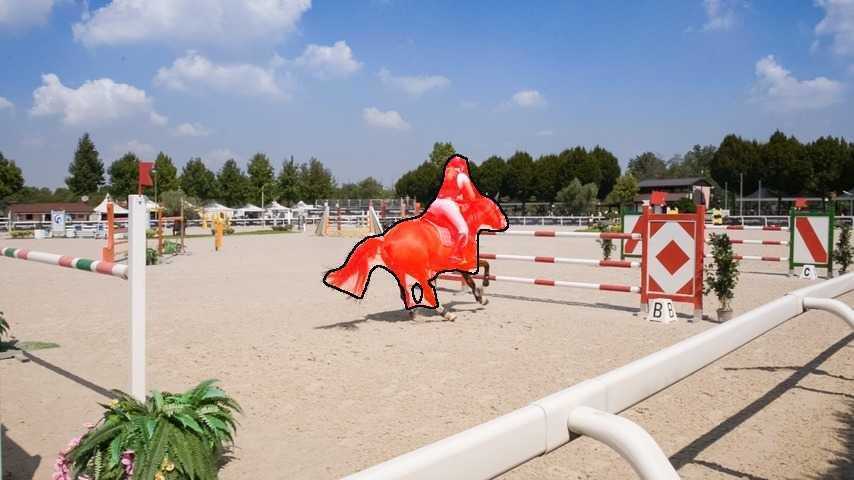}\tabularnewline
 & \begin{turn}{90}
{\footnotesize{}\hspace{0.5em}with CRF\hspace{-2em}}
\end{turn} & {\footnotesize{}\hspace{-1em}}\includegraphics[width=0.15\textwidth]{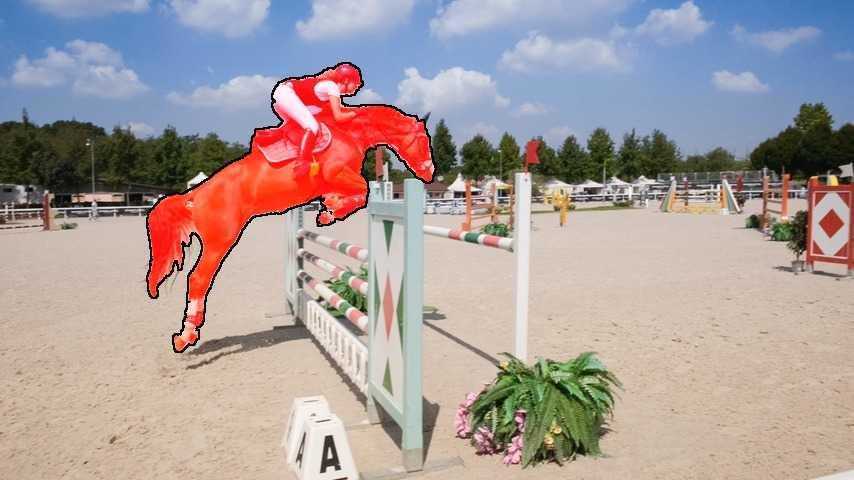} & {\footnotesize{}\hspace{-1em}}\includegraphics[width=0.15\textwidth]{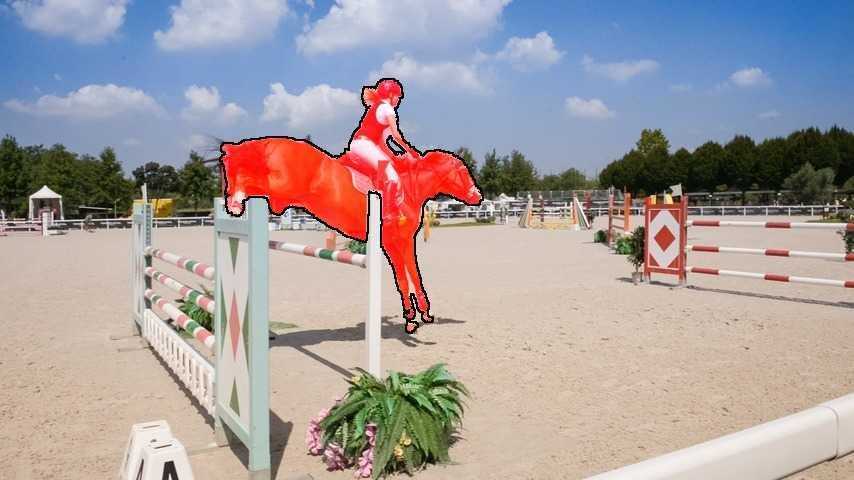} & {\footnotesize{}\hspace{-1em}}\includegraphics[width=0.15\textwidth]{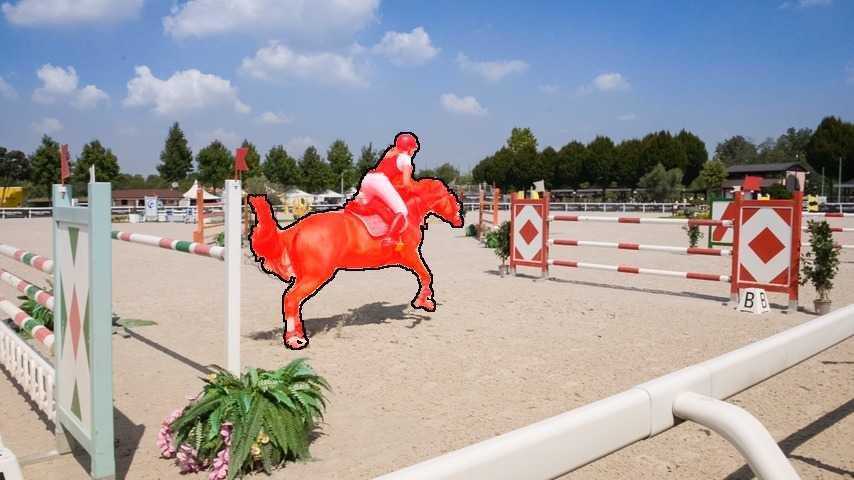} & {\footnotesize{}\hspace{-1em}}\includegraphics[width=0.15\textwidth]{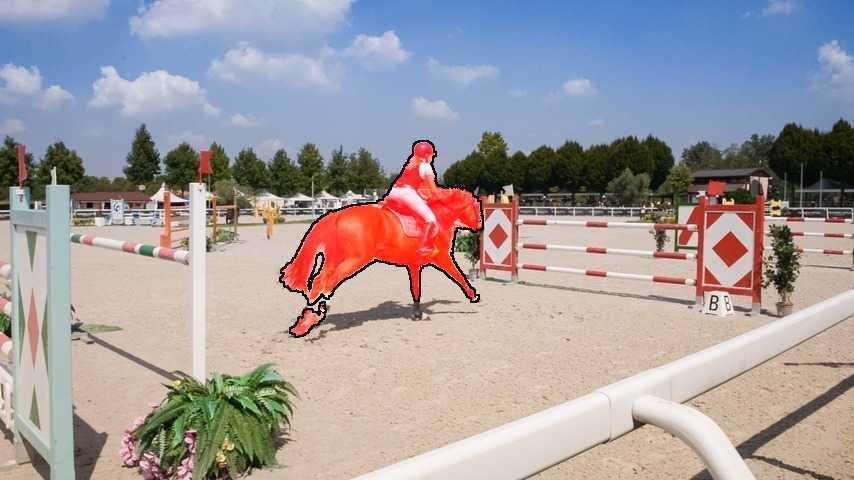} & {\footnotesize{}\hspace{-1em}}\includegraphics[width=0.15\textwidth]{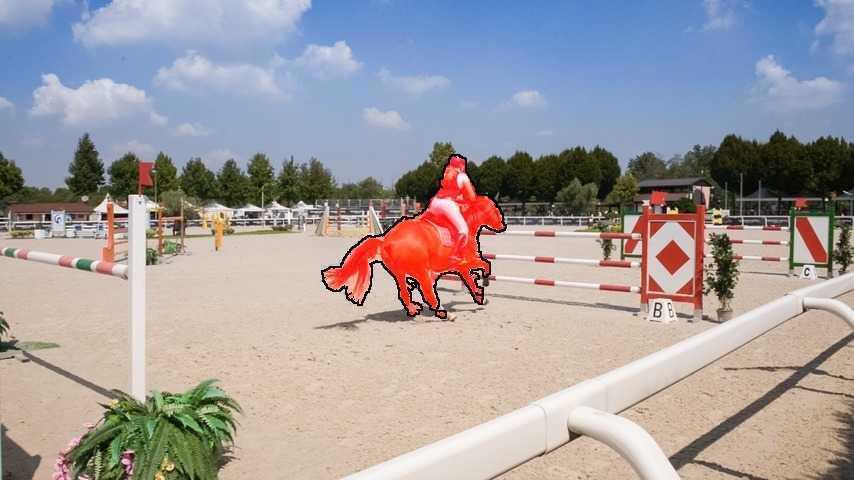}\tabularnewline
{\footnotesize{}\hspace{-1em}1st frame, GT segment} &  & {\footnotesize{}\hspace{-1em}$20\%$} & {\footnotesize{}\hspace{-1em}$40\%$} & {\footnotesize{}\hspace{-1em}$60\%$} & {\footnotesize{}\hspace{-1em}$80\%$} & {\footnotesize{}\hspace{-1em}$100\%$}\tabularnewline
\end{tabular}\hfill{}
\par\end{centering}
\smallskip{}

\caption{\label{fig:crf_effect}Effect of CRF tuning. The shown $\text{DAVIS}_{\text{16}}$
videos have the highest margin between with and without CRF post-processing
(based on mIoU over the video). }
\end{figure*}

As a final stage of our pipeline, we refine the generated mask using
DenseCRF \cite{Kraehenbuehl2011Nips} per frame. This captures small
image details that the network might have missed. It is known by practitioners
that DenseCRF is quite sensitive to its parameters and can easily
worsen results. We use our lucid dreams to enable automatic per-dataset
CRF-tuning.

Following \cite{Chen2016ArxivDeeplabv2} we employ grid search scheme
for tuning CRF parameters. Once the per-dataset model is
trained, we apply it over a subset of its training set (5 random images
from the lucid dreams per video sequence), apply DenseCRF with the
given parameters over this output, and then compare to the lucid dream
ground truth.

The impact of the tuned parameter of DenseCRF post-processing is shown
in Table \ref{tab:Effect-of-CRF} and Figure \ref{fig:crf_effect}.
Table \ref{tab:Effect-of-CRF} indicates that without per-dataset
tuning DenseCRF is under-performing. Our automated tuning procedure
allows to obtain consistent gains without the need for case-by-case
manual tuning.

\begin{table}
\setlength{\tabcolsep}{0.3em} 
\renewcommand{\arraystretch}{1.3}
\centering{}\hspace*{\fill}%
\begin{tabular}{c|c|ccc}
\multirow{2}{*}{{\footnotesize{}Method}} & {\footnotesize{}CRF} & \multicolumn{3}{c}{{\footnotesize{}Dataset, mIoU}}\tabularnewline
 & {\footnotesize{}parameters} & {\footnotesize{}$\text{DAVIS}_{\text{16}}$} & {\footnotesize{}YoutbObjs} & {\footnotesize{}$\mbox{SegTrck}_{\mbox{v2}}$}\tabularnewline
\hline 
\hline 
{\footnotesize{}$\text{LucidTracker}^{-}$} & {\footnotesize{}-} & {\small{}83.7} & {\small{}76.2} & {\small{}76.8}\tabularnewline
{\footnotesize{}$\text{LucidTracker}^{\ \ }$} & {\footnotesize{}default} & {\small{}84.2} & {\small{}75.5} & {\small{}72.2}\tabularnewline
{\footnotesize{}$\text{LucidTracker}^{\ \ }$} & {\footnotesize{}tuned per-dataset} & \textbf{\small{}84.8} & \textbf{\small{}76.2} & \textbf{\small{}77.6}\tabularnewline
\end{tabular}\hspace*{\fill}\smallskip{}
\caption{\label{tab:Effect-of-CRF}Effect of CRF tuning ($\mathtt{LucidTracker}$ without temporal coherency). 
Without the automated per-dataset tuning DenseCRF will under-perform.}
\end{table}

\paragraph{Conclusion}

Using default DenseCRF parameters would degrade performance. Our lucid
dreams enable automatic per-dataset CRF-tuning which allows to further improve
the results.

\subsection{\label{subsec:Additional-experiments}Additional experiments}

Other than adding or removing ingredients, as in Section \ref{subsec:Ablation-study},
we also want to understand how the training data itself affects the
obtained results. 

\subsubsection{\label{subsubsec:How-much-data}Generalization across videos}

Table \ref{tab:number-of-videos} explores the effect of segmentation
quality as a function of the number of training samples. To see more
directly the training data effects we use a base model with RGB image
$\mathcal{I}_{t}$ only (no flow $\mathcal{F}$, no CRF, no temporal coherency), and per-dataset
training (no ImageNet pre-training, no per-video fine-tuning). We
evaluate on two disjoint subsets of $15$ $\text{DAVIS}_{\text{16}}$
videos each, where the first frames for per-dataset training are taken
from only one subset. The reported numbers are thus comparable within
Table \ref{tab:number-of-videos}, but not across to the other tables
in the paper. Table \ref{tab:number-of-videos} reports results with
varying number of training videos and with/without including the first
frames of each test video for per-dataset training. When excluding
the test set first frames, the image frames used for training are
separate from the test videos; and we are thus operating across (related)
domains. When including the test set first frames, we operate in the
usual LucidTracker mode, where the first frame from each test video
is used to build the per-dataset training set.

Comparing the top and bottom parts of the table, we see that when
the annotated images from the test set video sequences are not included, segmentation
quality drops drastically (e.g. $68.7\negmedspace\rightarrow\negmedspace36.4\ \text{mIoU}$).
Conversely, on subset of videos for which the first frame annotation
is used for training, the quality is much higher and improves as the
training samples become more and more specific (in-domain) to the
target video ($65.4\negmedspace\rightarrow\negmedspace78.3\ \text{mIoU}$).
Adding extra videos for training does not improve the performance.
It is better ($68.7\negmedspace\rightarrow\negmedspace78.3\ \text{mIoU}$)
to have $15$ models each trained and evaluated on a single video
(row top-1-1) than having one model trained over $15$ test videos
(row top-15-1{\scriptsize{})}.

Training with an additional frame from each video (we added the last
frame of each train video) significantly boosts the resulting within-video
quality (e.g. row top-30-2 $65.4\negmedspace\rightarrow\negmedspace74.3\ \text{mIoU}$),
because the training samples cover better the test domain.

\begin{table}
\setlength{\tabcolsep}{0.5em} 
\renewcommand{\arraystretch}{1.3}
\begin{centering}
\vspace{0em}
\begin{tabular}{ccc|c}
 & \textcolor{black}{\small{}\# training} & \textcolor{black}{\small{}\# frames} & \multirow{2}{*}{\textcolor{black}{\small{}mIoU}}\tabularnewline
Training set & \textcolor{black}{\small{}videos} & \textcolor{black}{\small{}per video} & \tabularnewline
\hline 
\hline 
\multirow{5}{*}{{\small{}}%
\begin{tabular}{c}
{\small{}Includes 1st frames}\tabularnewline
{\small{}from test set}\tabularnewline
\end{tabular}{\small{} }} & \textcolor{black}{\small{}1} & \textcolor{black}{\small{}1} & \textcolor{black}{\small{}78.3}\tabularnewline
 & \textcolor{black}{\small{}2} & \textcolor{black}{\small{}1} & \textcolor{black}{\small{}75.4}\tabularnewline
 & \textcolor{black}{\small{}15} & \textcolor{black}{\small{}1} & \textcolor{black}{\small{}68.7}\tabularnewline
 & \textcolor{black}{\small{}30} & \textcolor{black}{\small{}1} & \textcolor{black}{\small{}65.4\arrayrulecolor{lightgray}}\tabularnewline
\cline{2-4} 
 & \textcolor{black}{\small{}\arrayrulecolor{black}30} & \textcolor{black}{\small{}2} & \textcolor{black}{\small{}74.3}\tabularnewline
\hline 
\hline 
\multirow{4}{*}{{\small{}}%
\begin{tabular}{c}
{\small{}Excludes 1st frames}\tabularnewline
{\small{}from test set}\tabularnewline
\end{tabular}{\small{} }} & \textcolor{black}{\small{}2} & \textcolor{black}{\small{}1} & \textcolor{black}{\small{}11.6}\tabularnewline
 & \textcolor{black}{\small{}15} & \textcolor{black}{\small{}1} & \textcolor{black}{\small{}36.4}\tabularnewline
 & \textcolor{black}{\small{}30} & \textcolor{black}{\small{}1} & \textcolor{black}{\small{}41.7\arrayrulecolor{lightgray}}\tabularnewline
\cline{2-4} 
 & \textcolor{black}{\small{}\arrayrulecolor{black}30} & \textcolor{black}{\small{}2} & \textcolor{black}{\small{}48.4}\tabularnewline
\end{tabular}
\par\end{centering}
\caption{\label{tab:number-of-videos}Varying the number of training videos.
A smaller training set closer to the target domain is better than
a larger one. See Section \ref{subsubsec:How-much-data}.}
\vspace{0em}
\end{table}

\paragraph{Conclusion}

These results show that, when using RGB information ($\mathcal{I}_{t}$),
increasing the number of training videos \emph{does not} improve the
resulting quality of our system. Even within a dataset, properly using
the training sample(s) from within each video matters more than collecting
more videos to build a larger training set.

\subsubsection{\label{subsubsec:Which-training-set}Generalization across datasets}

Section \ref{subsubsec:How-much-data} has explored the effect of
changing the volume of training data within one dataset, Table \ref{tab:general-across-datasets}
compares results when using different datasets for training. Results
are obtained using a base model with RGB and flow ($M_{t}=f\left(\mathcal{I}_{t},\,M_{t-1}\right)$,
no warping, no CRF, no temporal coherency), ImageNet pre-training, per-dataset training,
and no per-video tuning to accentuate the effect of the training dataset.

The best performance is obtained when training on the first frames
of the target set. There is a noticeable $\sim\negmedspace10$ percent
points drop when moving to the second best choice (e.g. $80.9\negmedspace\rightarrow\negmedspace67.0$
for $\text{DAVIS}_{\text{16}}$). Interestingly, when putting all
the datasets together for training (\textquotedbl{}all-in-one\textquotedbl{}
row, a dataset-agnostic model) the results degrade, reinforcing the
idea that \textquotedbl{}just adding more data\textquotedbl{} does
not automatically make the performance better.

\paragraph{Conclusion}

Best results are obtained when using training data that focuses on
the test video sequences, using similar datasets or combining multiple
datasets degrades the performance for our system.

\begin{table}
\setlength{\tabcolsep}{0.3em} 
\renewcommand{\arraystretch}{1.3}
\begin{centering}
\vspace{0em}
{\small{}}%
\begin{tabular}{c|ccc|c}
\multirow{2}{*}{{\small{}Training set}} & \multicolumn{3}{c|}{{\small{}Dataset, mIoU}} & \multirow{2}{*}{{\small{}Mean}}\tabularnewline
 & {\small{}$\text{DAVIS}_{\text{16}}$} & {\small{}YoutbObjs} & {\small{}$\mbox{SegTrck}_{\mbox{v2}}$} & \tabularnewline
\hline 
\hline 
{\small{}$\text{DAVIS}_{\text{16}}$} & \textcolor{black}{\small{}\uline{80.9}} & {\small{}50.9} & {\small{}46.9} & {\small{}59.6}\tabularnewline
{\small{}YoutbObjs} & \emph{\small{}67.0} & \textcolor{black}{\small{}\uline{71.5}} & \emph{\small{}52.0} & {\small{}63.5}\tabularnewline
{\small{}$\mbox{SegTrack}_{\mbox{v2}}$} & {\small{}56.0} & \emph{\small{}52.2} & \textcolor{black}{\small{}\uline{66.4}} & {\small{}58.2}\tabularnewline
\hline 
{\small{}Best} & \textbf{\small{}80.9} & \textbf{\small{}71.5} & \textbf{\small{}66.4} & \textbf{\small{}72.9}\tabularnewline
{\small{}Second best} & {\small{}67.0} & {\small{}52.2} & {\small{}52.0} & {\small{}57.1\arrayrulecolor{lightgray}}\tabularnewline
\hline 
{\small{}\arrayrulecolor{black}All-in-one} & {\small{}71.9} & {\small{}70.7} & {\small{}60.8} & \textit{\emph{\small{}67.8}}\tabularnewline
\end{tabular}
\par\end{centering}{\small \par}
\caption{\label{tab:general-across-datasets}Generalization across datasets.
Results with underline are the best per dataset, and in italic are
the second best per dataset (ignoring all-in-one setup). We observe
a significant quality gap between training from the target videos,
versus training from other datasets; see Section \ref{subsubsec:Which-training-set}. }
\vspace{0em}
\end{table}

\subsubsection{\label{subsubsec:convnet-arch}Experimenting with the convnet architecture}

Section \ref{subsec:Architecture} and Figure \ref{fig:Two-and-single-stream-architectures}
described two possible architectures to handle $\mathcal{I}_{t}$
and $\mathcal{F}_{t}$. Previous experiments are all based on the
two streams architecture. 

Table \ref{tab:Two-versus-one-stream-architectre} compares two streams
versus one stream, where the network to accepts 5 input channels (RGB
+ previous mask + flow magnitude) in one stream: $M_{t}=f_{\mathcal{I+F}}\left(\mathcal{I}_{t}\right.$,
$\text{\ensuremath{\mathcal{F}_{t}}}$, $\left.w(M_{t-1},\mathcal{\,F}_{t})\right)$.
Results are obtained using a base model with RGB and optical flow
(no warping, no CRF, no temporal coherency), ImageNet pre-training, per-dataset training,
and no per-video tuning. 

We observe that both one stream and two stream architecture with naive
averaging perform on par. Using a one stream network makes the training
more affordable and allows more easily to expand the architecture
with additional input channels. 
\begin{table}
\setlength{\tabcolsep}{0.3em}
\renewcommand{\arraystretch}{1.3}
\begin{centering}
\vspace{0em}
\begin{tabular}{c|ccc|c}
\multirow{2}{*}{{\small{}Architecture}} & \multirow{2}{*}{{\small{}}%
\begin{tabular}{c}
{\small{}ImgNet}\tabularnewline
{\small{}pre-train.}\tabularnewline
\end{tabular}} & \multirow{2}{*}{{\small{}}%
\begin{tabular}{c}
{\small{}per-dataset}\tabularnewline
{\small{}training}\tabularnewline
\end{tabular}} & \multirow{2}{*}{{\small{}}%
\begin{tabular}{c}
{\small{}per-video}\tabularnewline
{\small{}fine-tun.}\tabularnewline
\end{tabular}} & \multicolumn{1}{c}{$\text{DAVIS}_{\text{16}}$}\tabularnewline
 &  &  &  & {\small{} mIoU}\tabularnewline
\hline 
\hline 
{\small{}two streams} & \textcolor{black}{\small{}\Checkmark{}} & \textcolor{black}{\small{}\Checkmark{}} & \textcolor{black}{\small{}\XSolidBrush{}} & \textbf{\small{}80.9}\tabularnewline
{\small{}one stream} & \textcolor{black}{\small{}\Checkmark{}} & \textcolor{black}{\small{}\Checkmark{}} & \textcolor{black}{\small{}\XSolidBrush{}} & {\small{}80.3}\tabularnewline
\end{tabular}
\par\end{centering}

\caption{\label{tab:Two-versus-one-stream-architectre}Experimenting with the
convnet architecture. See Section \ref{subsubsec:convnet-arch}. }
\vspace{0em}
\end{table}

\paragraph{Conclusion}

The lighter one stream network performs as well as a network with
two streams. We will thus use the one stream architecture in Section
\ref{sec:Multiple-object-results}.

\begin{table*}
\setlength{\tabcolsep}{0.4em} 
\renewcommand{\arraystretch}{1.35}
\begin{centering}
\hspace*{\fill}%
\begin{tabular}{cccc|ccc|ccc|c}
\multirow{3}{*}{} & \multirow{3}{*}{{\footnotesize{}Method}} & \multirow{3}{*}{{\footnotesize{}}%
\begin{tabular}{c}
{\footnotesize{}\# training}\tabularnewline
{\footnotesize{}images}\tabularnewline
\end{tabular}} & \multirow{3}{*}{{\footnotesize{}}%
\begin{tabular}{c}
{\footnotesize{}Flow}\tabularnewline
{\footnotesize{}$\mathcal{F}$}\tabularnewline
\end{tabular}} & \multicolumn{7}{c}{{\small{}$\text{DAVIS}_{\text{16}}$}}\tabularnewline
\cline{5-11} 
 &  &  &  & \multicolumn{3}{c|}{{\footnotesize{}Region,}{\small{} \ensuremath{J} }} & \multicolumn{3}{c|}{{\footnotesize{}Boundary,}{\small{} \ensuremath{F} }} & {\footnotesize{}Temporal stability,}{\small{} \ensuremath{T} }\tabularnewline
 &  &  &  & {\footnotesize{}Mean $\uparrow$} & {\footnotesize{}Recall $\uparrow$} & {\footnotesize{}Decay $\downarrow$} & {\footnotesize{}Mean $\uparrow$} & {\footnotesize{}Recall $\uparrow$} & {\footnotesize{}Decay $\downarrow$} & {\footnotesize{}Mean $\downarrow$}\tabularnewline
\hline 
\hline 
\multirow{2}{*}{} & {\small{}Box oracle \cite{Khoreva2017CvprMaskTrack}} & {\small{}0} & \textbf{\textcolor{black}{\scriptsize{}\XSolidBrush{}}} & {\small{}45.1} & {\small{}39.7} & \textbf{\small{}-0.7} & {\small{}21.4} & {\small{}6.7} & {\small{}1.8} & \textbf{\small{}1.0}\tabularnewline
 & {\small{}Grabcut oracle \cite{Khoreva2017CvprMaskTrack}} & {\small{}0} & \textbf{\textcolor{black}{\scriptsize{}\XSolidBrush{}}} & {\small{}67.3} & {\small{}76.9} & {\small{}1.5} & {\small{}65.8} & {\small{}77.2} & {\small{}2.9} & {\small{}34.0}\tabularnewline
\hline 
\multirow{7}{*}{{\footnotesize{}}%
\begin{tabular}{c}
{\footnotesize{}Ignores 1st frame}\tabularnewline
{\footnotesize{}annotation}\tabularnewline
\end{tabular}{\footnotesize{} }} & {\small{}Saliency} & {\small{}0} & \textbf{\textcolor{black}{\scriptsize{}\XSolidBrush{}}} & {\small{}32.7} & {\small{}22.6} & {\small{}-0.2} & {\small{}26.9} & {\small{}10.3} & {\small{}0.9} & {\small{}32.8}\tabularnewline
 & {\small{}NLC \cite{Faktor2014Bmvc}} & {\small{}0} & \textbf{\textcolor{black}{\scriptsize{}\Checkmark{}}} & {\small{}64.1} & {\small{}73.1} & {\small{}8.6} & {\small{}59.3} & {\small{}65.8} & {\small{}8.6} & {\small{}35.8}\tabularnewline
 & {\small{}MP-Net \cite{Tokmakov2016Arxiv}} & {\small{}\textasciitilde{}22.5k} & \textbf{\textcolor{black}{\scriptsize{}\Checkmark{}}} & {\small{}69.7} & {\small{}82.9} & {\small{}5.6} & {\small{}66.3} & {\small{}78.3} & {\small{}6.7} & {\small{}68.6}\tabularnewline
 & {\small{}Flow saliency} & {\small{}0} & \textbf{\textcolor{black}{\scriptsize{}\Checkmark{}}} & {\small{}70.7} & {\small{}83.2} & {\small{}6.7} & {\small{}69.7} & {\small{}82.9} & {\small{}7.9} & {\small{}48.2}\tabularnewline
 & {\small{}FusionSeg \cite{Jain2017ArxivFusionSeg}} & {\small{}\textasciitilde{}95k} & {\textcolor{black}{\scriptsize{}\Checkmark{}}} & {\small{}71.5} & {\small{}-} & {\small{}-} & {\small{}-} & {\small{}-} & {\small{}-} & {\small{}-}\tabularnewline
 
& \textcolor{black} {\small{}LVO \cite{TokmakovAS17}} & {\small{}\textasciitilde{}35k} & \textbf{\textcolor{black}{\scriptsize{}\Checkmark{}}} & \textit{\small{}75.9}  & \textit{\small{}89.1} & \textit{\small{}0.0} & \textit{\small{}72.1} & \textit{\small{}83.4} & \textit{\small{}1.3} & \textit{\small{}26.5}\tabularnewline
   & \textcolor{black} {\small{}PDB \cite{Song_2018_ECCV}} & {\small{}\textasciitilde{}18k} & \textbf{\textcolor{black}{\scriptsize{}\XSolidBrush{}}} & \textit{\small{}77.2} & \textit{\small{}90.1} & \textit{\small{}0.9} & \textit{\small{}74.5} & \textit{\small{}84.4} & \textbf{\textit{\small{}-0.2}} & \textit{\small{}29.1}\tabularnewline
   
\hline 
\multirow{12}{*}{{\footnotesize{}}%
\begin{tabular}{c}
{\footnotesize{}Uses 1st frame}\tabularnewline
{\footnotesize{}annotation}\tabularnewline
\end{tabular}{\footnotesize{} }} & {\small{}Mask warping} & {\small{}0} & \textbf{\textcolor{black}{\scriptsize{}\Checkmark{}}} & {\small{}32.1} & {\small{}25.5} & {\small{}31.7} & {\small{}36.3} & {\small{}23.0} & {\small{}32.8} & {\small{}8.4}\tabularnewline
 & {\small{}FCP \cite{Perazzi2015Iccv}} & {\small{}0} & \textcolor{black}{\scriptsize{}\Checkmark{}} & {\small{}63.1} & {\small{}77.8} & {\small{}3.1} & {\small{}54.6} & {\small{}60.4} & {\small{}3.9} & {\small{}28.5}\tabularnewline
 & {\small{}BVS \cite{Maerki2016Cvpr}} & {\small{}0} & \textcolor{black}{\scriptsize{}\XSolidBrush{}} & {\small{}66.5} & {\small{}76.4} & {\small{}26.0} & {\small{}65.6} & {\small{}77.4} & {\small{}23.6} & {\small{}31.6}\tabularnewline
 & {\small{}ObjFlow \cite{Tsai2016Cvpr}} & {\small{}0} & \textcolor{black}{\scriptsize{}\Checkmark{}} & {\small{}71.1} & {\small{}80.0} & {\small{}22.7} & {\small{}67.9} & {\small{}78.0} & {\small{}24.0} & {\small{}22.1}\tabularnewline
 & {\small{}STV \cite{Wang2017ArxivSTV}} & {\small{}0} & \textcolor{black}{\scriptsize{}\Checkmark{}} & {\small{}73.6} & {\small{}-} & {\small{}-} & {\small{}72.0} & {\small{}-} & {\small{}-} & {\small{}-}\tabularnewline
 & {\small{}VPN \cite{Jampani2016Arxiv}} & {\small{}\textasciitilde{}2.3k} & \textcolor{black}{\scriptsize{}\XSolidBrush{}} & \textit{\textcolor{black}{\small{}75.0}} & {\small{}-} & {\small{}-} & \textit{\small{}72.4} & {\small{}-} & {\small{}-} & \textit{\small{}29.5}\tabularnewline
 & {\small{}OSVOS \cite{Caelles2017Cvpr}} & {\small{}\textasciitilde{}2.3k} & \textcolor{black}{\scriptsize{}\XSolidBrush{}} & \textit{\textcolor{black}{\small{}79.8}} & \textit{\small{}93.6} & \textit{\small{}14.9} & \textit{\small{}80.6} & \textit{\small{}92.6} & \textit{\small{}15.0} & \textit{\small{}37.6}\tabularnewline
 & {\small{}MaskTrack \cite{Khoreva2017CvprMaskTrack}} & {\small{}\textasciitilde{}11k} & \textcolor{black}{\scriptsize{}\Checkmark{}} & {\small{}80.3} & {\small{}93.5} & {\small{}8.9} & {\small{}75.8} & {\small{}88.2} & {\small{}9.5} & {\small{}18.3\arrayrulecolor{lightgray}}\tabularnewline
 
& \textcolor{black} {\small{}PReMVOS \cite{Luiten18ACCV}} & {\small{}\textasciitilde{}145k} & \textcolor{black}{\scriptsize{}\Checkmark{}} & \textit{\textcolor{black}{\small{}84.9}} & \textit{\small{}96.1} & \textit{\small{}8.8} & \textbf{\textit{\small{}88.6}} & \textbf{\textit{\small{}94.7}} & \textit{\small{}9.8} & \textit{\small{}19.7}\tabularnewline
& \textcolor{black} {\small{}OnAVOS \cite{Voigtlaender2017OnlineAO}} & {\small{}\textasciitilde{}120k} & \textcolor{black}{\scriptsize{}\XSolidBrush{}} & \textit{\textcolor{black}{\small{}86.1}} & \textit{\small{}96.1} & \textit{\small{}5.2} & \textit{\small{}84.9} & \textit{\small{}89.7} & \textit{\small{}5.8} & \textit{\small{}19.0}\tabularnewline
& \textcolor{black} {\small{}VideoGCRF \cite{Chandra2018DeepSR}} & {\small{}\textasciitilde{}120k} & \textcolor{black}{\scriptsize{}\XSolidBrush{}} & \textit{\textcolor{black}{\small{}86.5}} & \textit{\small{}-} & \textit{\small{}-} & \textit{\small{}-} & \textit{\small{}-} & \textit{\small{}-} & \textit{\small{}-}\tabularnewline
 
\cline{2-11} 
 & {\small{}\arrayrulecolor{black}}$\text{LucidTracker}$ & \textbf{\small{}50} & \textcolor{black}{\scriptsize{}\Checkmark{}} & \textcolor{black}{\textbf{\small{86.6}}} & \textcolor{black}{\textbf{\small{}{97.3}}} & \textcolor{black}{{\small{}{5.3}}} & \textcolor{black}{{\small{}{84.8}}} & \textcolor{black}{{\small{}{93.1}}} & \textcolor{black}{{\small{}{7.5}}} & \textcolor{black}{\small{}{15.9}} \tabularnewline
 \end{tabular}\hfill{}\medskip{}
\par\end{centering}

\caption{\label{tab:comparative-result-davis}Comparison of video object segmentation
results on $\text{DAVIS}_{\text{16}}$ benchmark. Numbers in italic
are computed based on subsets of $\text{DAVIS}_{\text{16}}$. Our
$\text{LucidTracker}$ improves over previous results.}
\end{table*}

\subsection{\label{subsec:Error-analysis}Error analysis}

Table \ref{tab:comparative-result-davis} presents an expanded evaluation
on $\text{DAVIS}_{\text{16}}$ using evaluation metrics proposed in
\cite{Perazzi2016Cvpr}. Three measures are used: region similarity
in terms of intersection over union (J), contour accuracy (F, higher
is better), and temporal instability of the masks (T, lower is better).
We outperform the competitive methods of \cite{Khoreva2017CvprMaskTrack,Caelles2017Cvpr}
on all three measures.

Table \ref{tab:Attribute-evaluation} reports the per-attribute based
evaluation as defined in $\text{DAVIS}_{\text{16}}$. $\mathtt{Lucid}$\-$\mathtt{Tracker}$
is best on all 15 video attribute categories. This shows that our 
$\mathtt{Lucid}$\-$\mathtt{Tracker}$ can handle the various video challenges
present in $\text{DAVIS}_{\text{16}}$.

\begin{figure*}[t]
\begin{centering}
\includegraphics[width=1\textwidth]{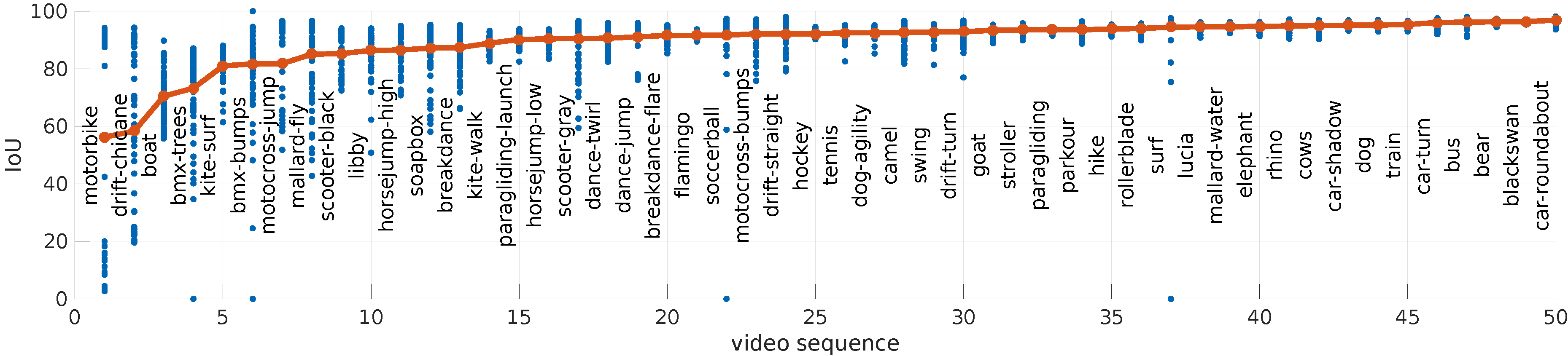}
\par\end{centering}
\caption{\label{fig:Per-sequence-results-davis16}\textcolor{black}{Per-sequence results on $\text{DAVIS}_{\text{16}}$.}}
\end{figure*}

We present the per-sequence and per-frame results of $\mathtt{Lucid}$\-$\mathtt{Tracker}$
over $\text{DAVIS}_{\text{16}}$ in Figure \ref{fig:Per-sequence-results-davis16}.
On the whole we observe that the proposed approach is quite robust,
most video sequences reach an average performance above $80$ mIoU. 

\begin{figure*}
\begin{centering}
\setlength{\tabcolsep}{0em} 
\renewcommand{\arraystretch}{0}
\par\end{centering}
\begin{centering}
\hfill{}%
\begin{tabular}{ccccccc}
\multirow{2}{*}{\begin{turn}{90}
 {\footnotesize{}$\text{DAVIS}_{\text{16}}$}
\end{turn}} & {\footnotesize{}\hspace{-1em}}\includegraphics[width=0.15\textwidth]{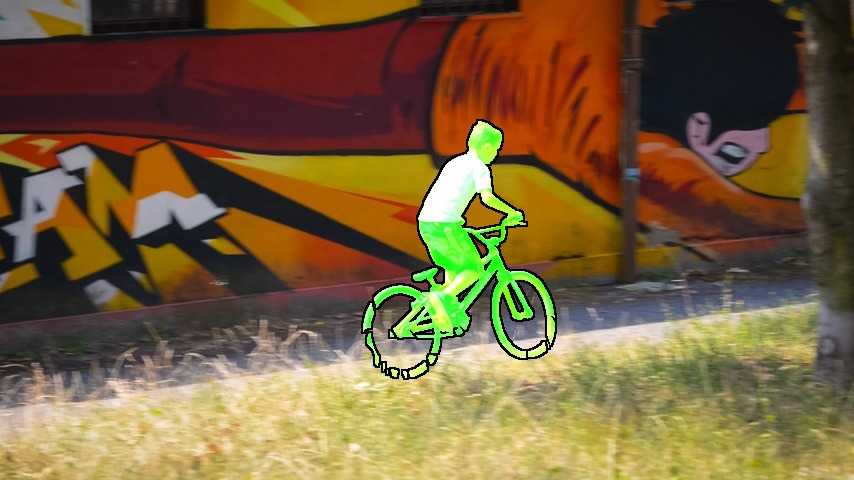} & {\footnotesize{}\hspace{-1em}}\includegraphics[width=0.15\textwidth]{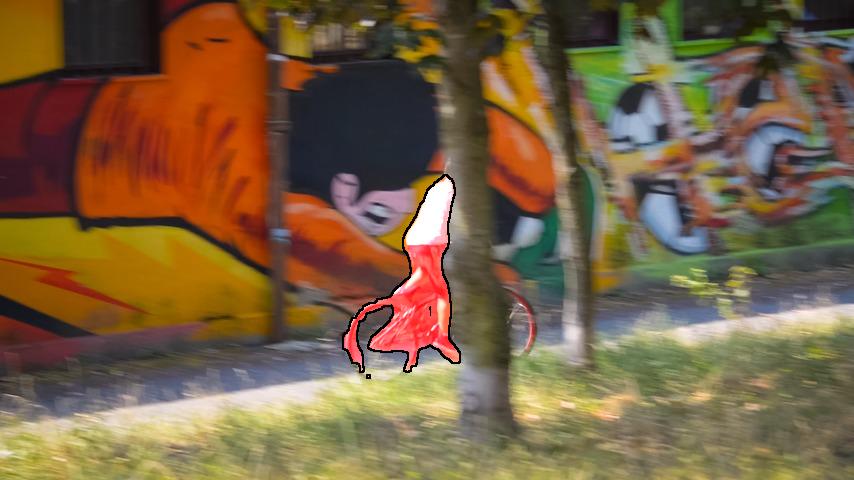} & {\footnotesize{}\hspace{-1em}}\includegraphics[width=0.15\textwidth]{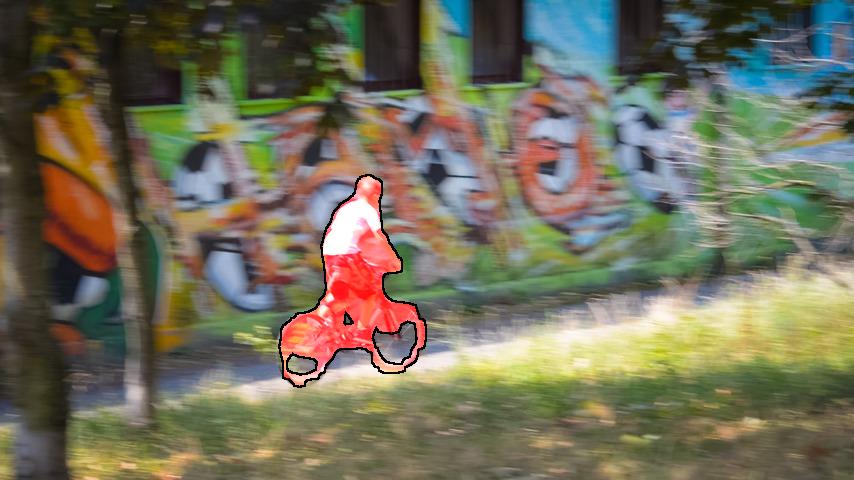} & {\footnotesize{}\hspace{-1em}}\includegraphics[width=0.15\textwidth]{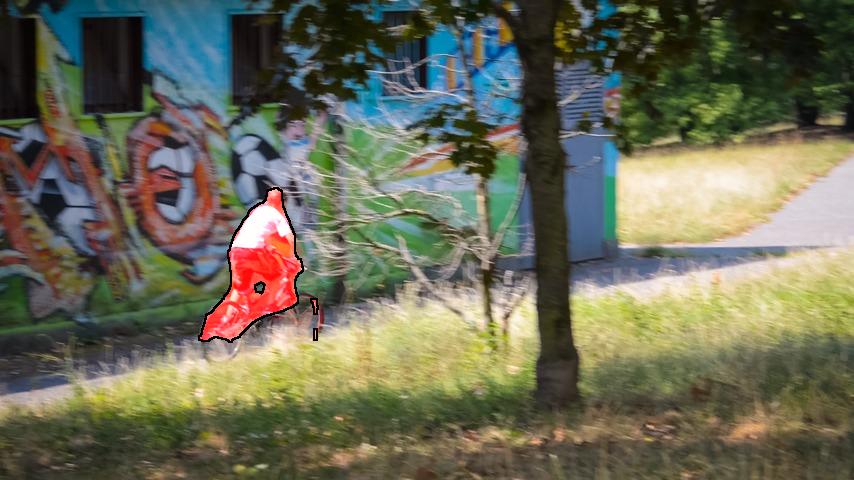} & {\footnotesize{}\hspace{-1em}}\includegraphics[width=0.15\textwidth]{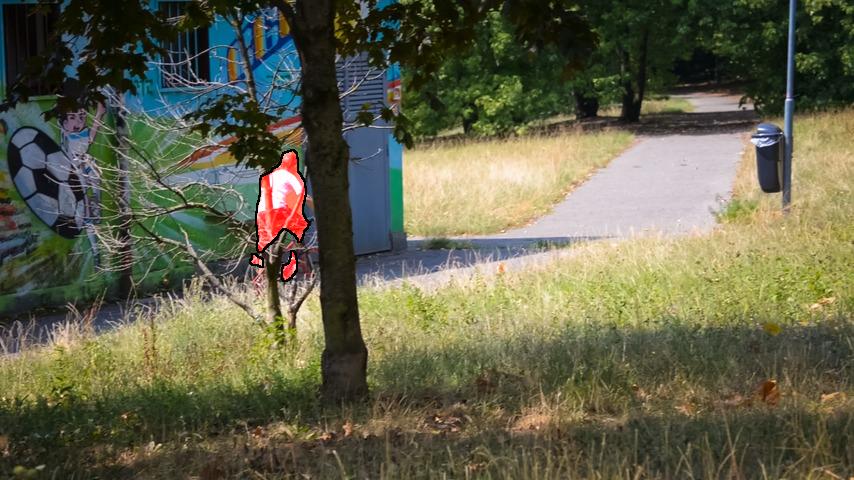} & {\footnotesize{}\hspace{-1em}}\includegraphics[width=0.15\textwidth]{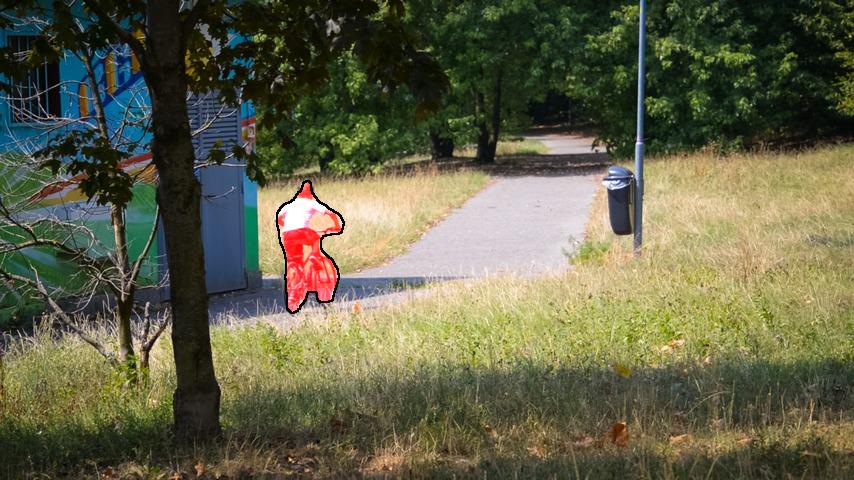}\tabularnewline
 & {\footnotesize{}\hspace{-1em}}\includegraphics[width=0.15\textwidth]{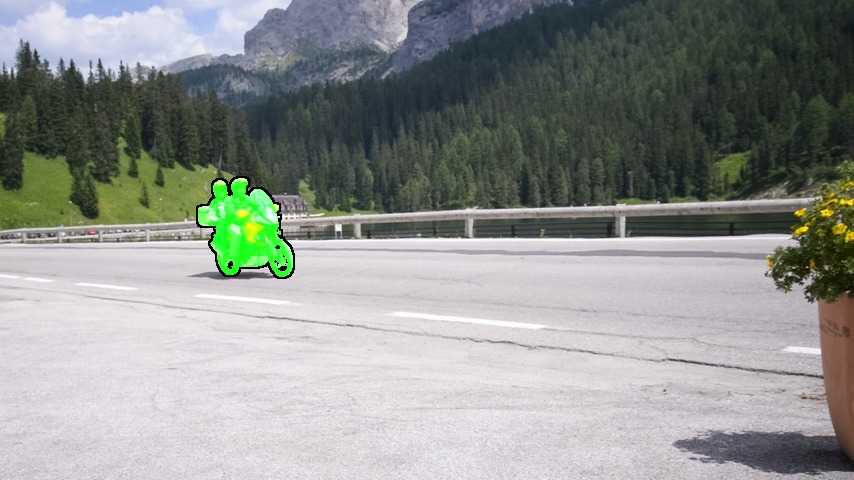} & {\footnotesize{}\hspace{-1em}}\includegraphics[width=0.15\textwidth]{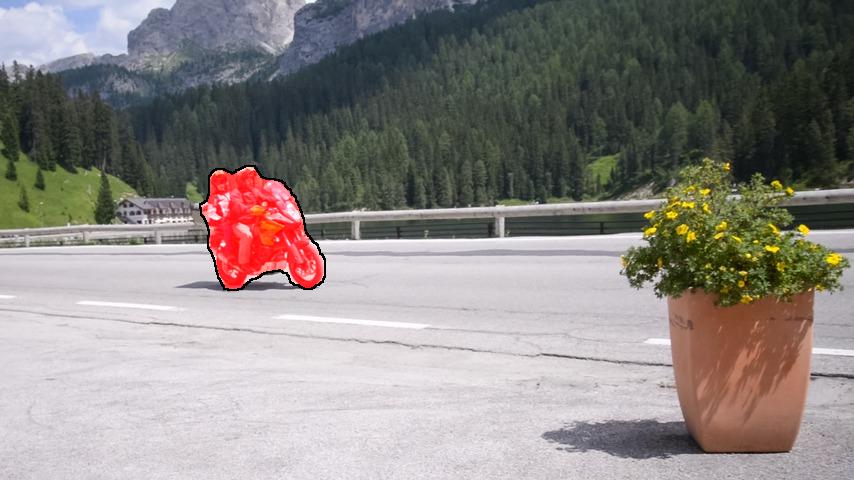} & {\footnotesize{}\hspace{-1em}}\includegraphics[width=0.15\textwidth]{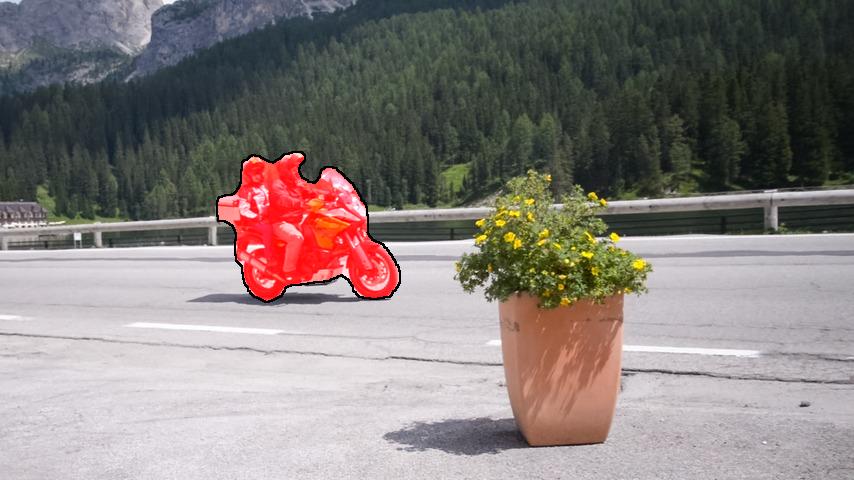} & {\footnotesize{}\hspace{-1em}}\includegraphics[width=0.15\textwidth]{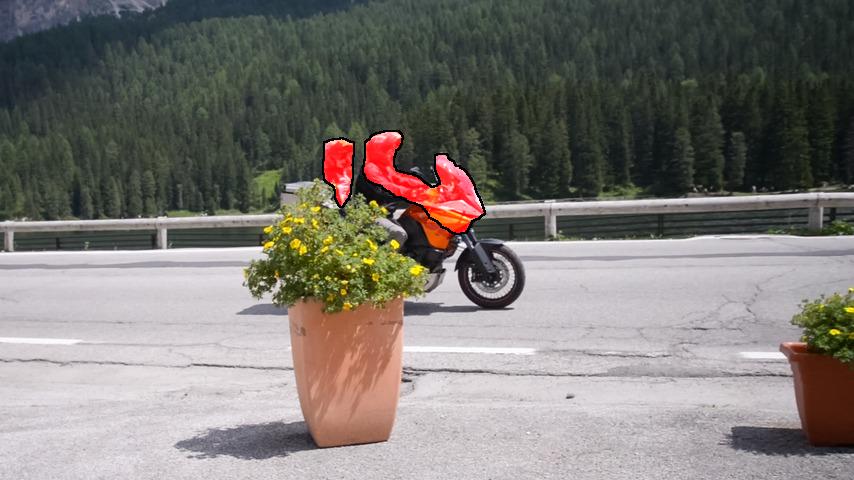} & {\footnotesize{}\hspace{-1em}}\includegraphics[width=0.15\textwidth]{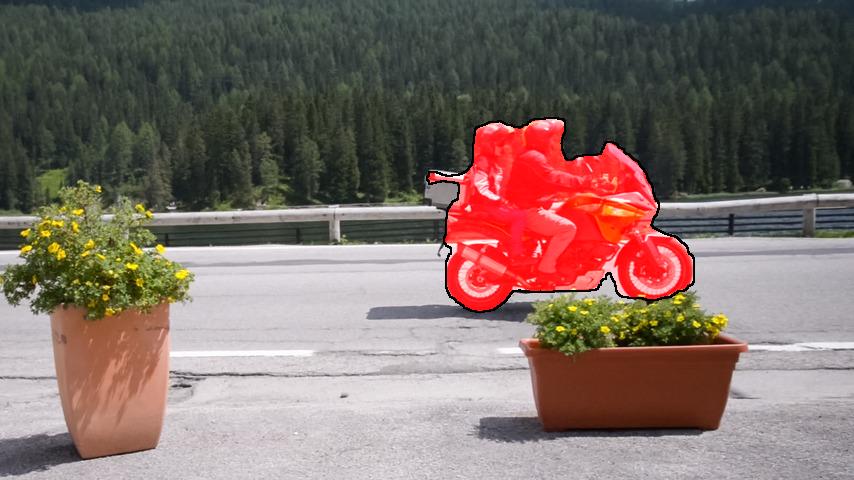} & {\footnotesize{}\hspace{-1em}}\includegraphics[width=0.15\textwidth]{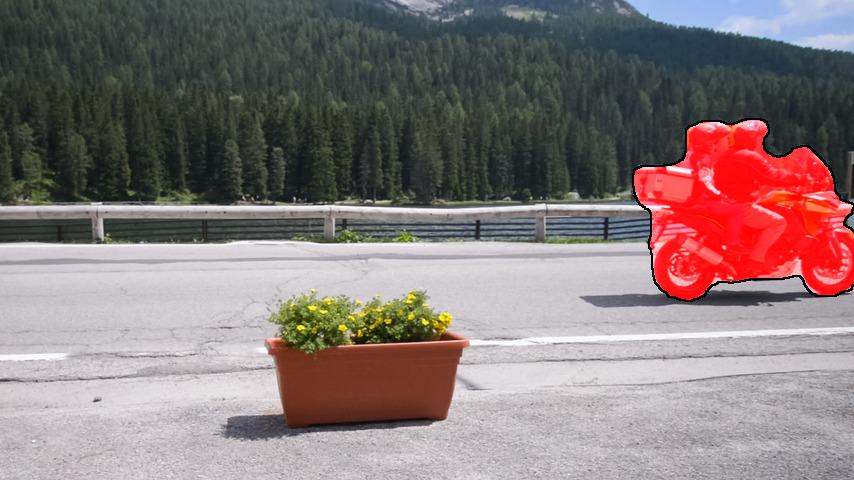}\tabularnewline
\multirow{2}{*}{\begin{turn}{90}
{\footnotesize{}YouTubeObjects\hspace{-2em}}
\end{turn}} & {\footnotesize{}\hspace{-1em}}\includegraphics[width=0.15\textwidth]{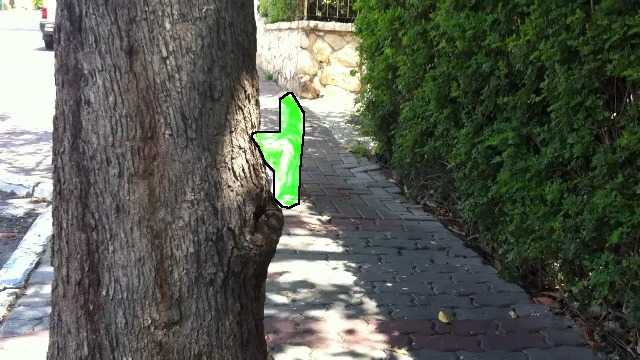} & {\footnotesize{}\hspace{-1em}}\includegraphics[width=0.15\textwidth]{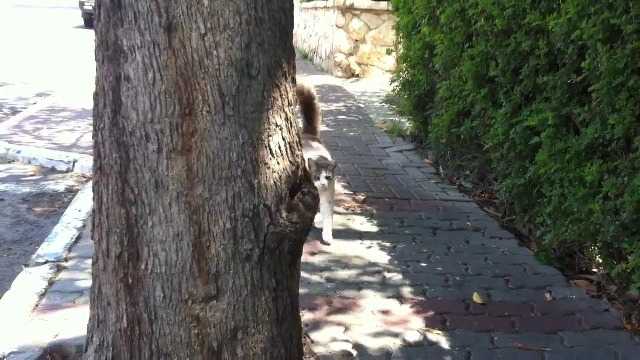} & {\footnotesize{}\hspace{-1em}}\includegraphics[width=0.15\textwidth]{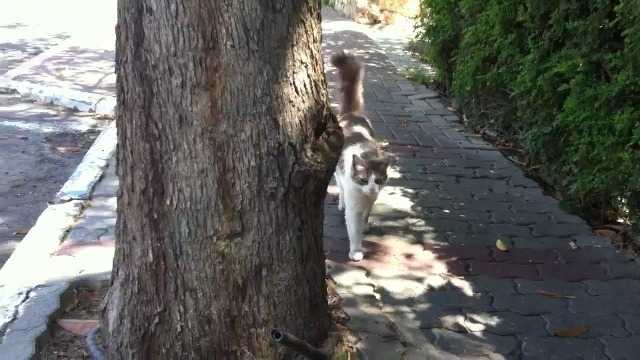} & {\footnotesize{}\hspace{-1em}}\includegraphics[width=0.15\textwidth]{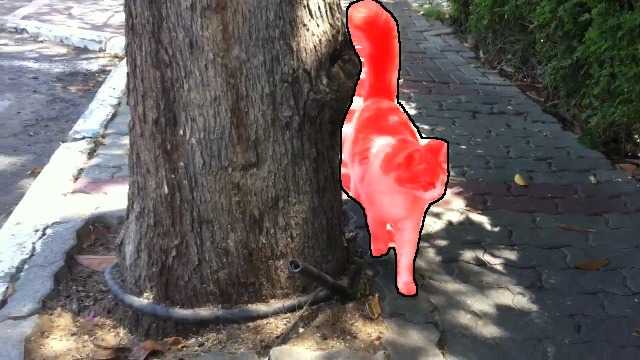} & {\footnotesize{}\hspace{-1em}}\includegraphics[width=0.15\textwidth]{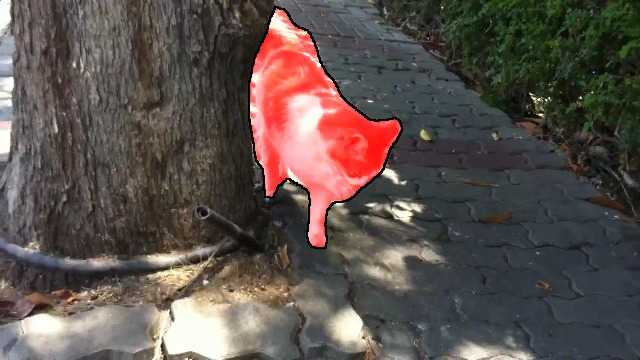} & {\footnotesize{}\hspace{-1em}}\includegraphics[width=0.15\textwidth]{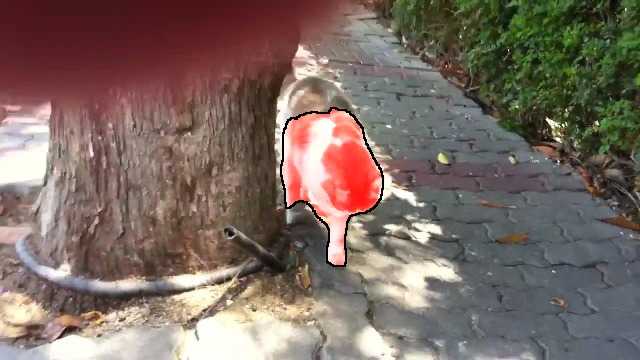}\tabularnewline
 & {\footnotesize{}\hspace{-1em}}\includegraphics[width=0.15\textwidth]{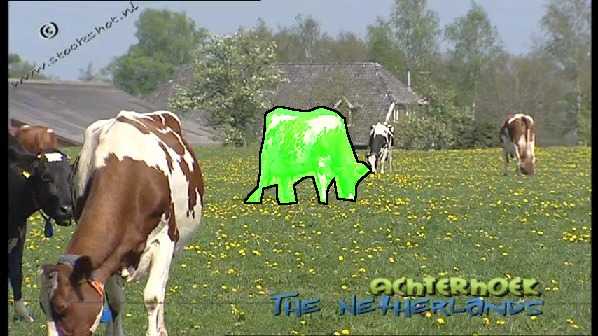} & {\footnotesize{}\hspace{-1em}}\includegraphics[width=0.15\textwidth]{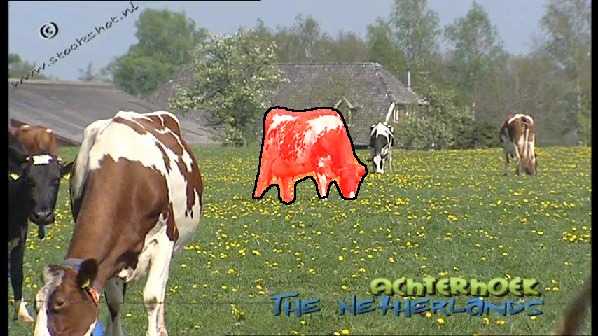} & {\footnotesize{}\hspace{-1em}}\includegraphics[width=0.15\textwidth]{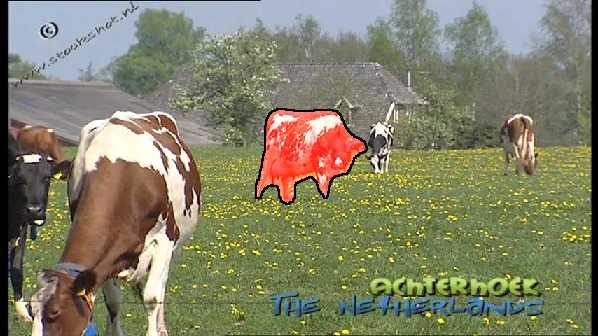} & {\footnotesize{}\hspace{-1em}}\includegraphics[width=0.15\textwidth]{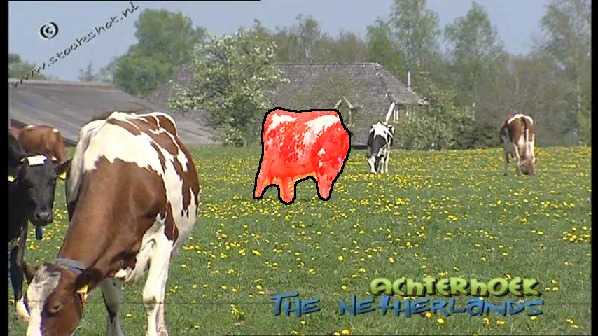} & {\footnotesize{}\hspace{-1em}}\includegraphics[width=0.15\textwidth]{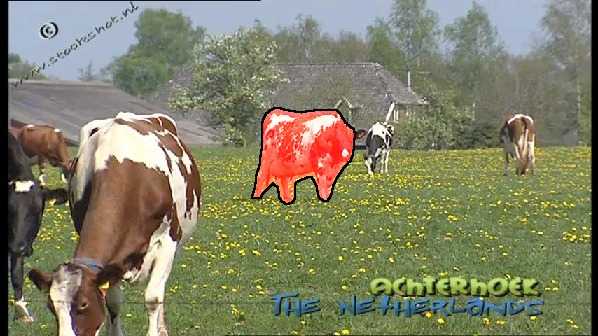} & {\footnotesize{}\hspace{-1em}}\includegraphics[width=0.15\textwidth]{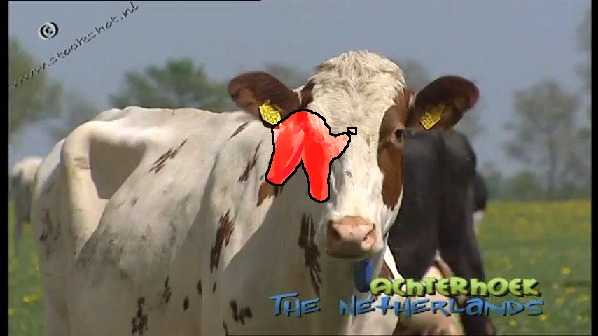}\tabularnewline
\multirow{2}{*}{\begin{turn}{90}
{\footnotesize{}$\text{SegTrack}_{\text{v2}}$\hspace{-0em}}
\end{turn}} & {\footnotesize{}\hspace{-1em}}\includegraphics[width=0.15\textwidth]{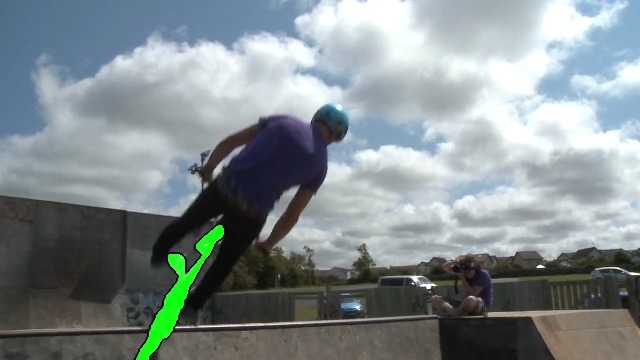} & {\footnotesize{}\hspace{-1em}}\includegraphics[width=0.15\textwidth]{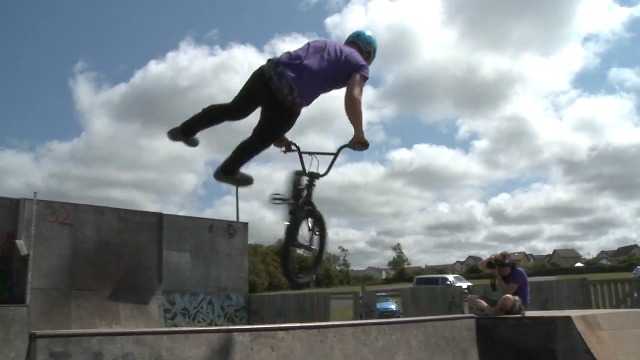} & {\footnotesize{}\hspace{-1em}}\includegraphics[width=0.15\textwidth]{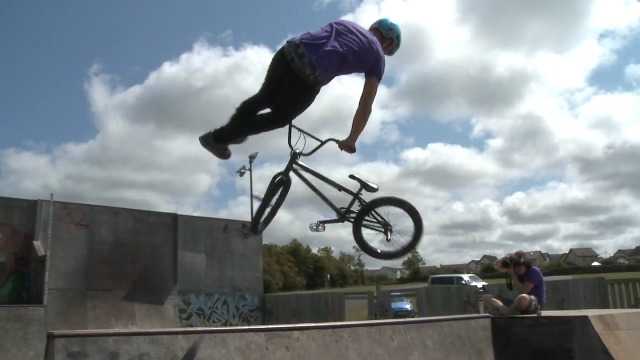} & {\footnotesize{}\hspace{-1em}}\includegraphics[width=0.15\textwidth]{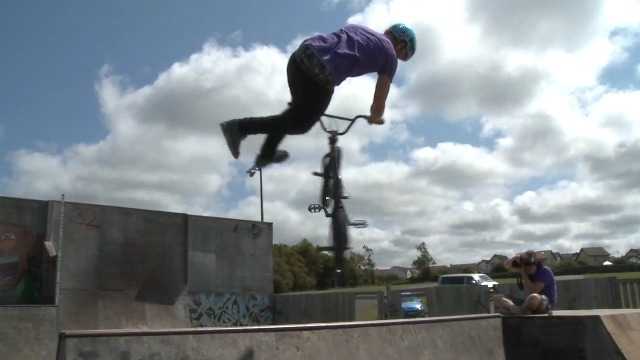} & {\footnotesize{}\hspace{-1em}}\includegraphics[width=0.15\textwidth]{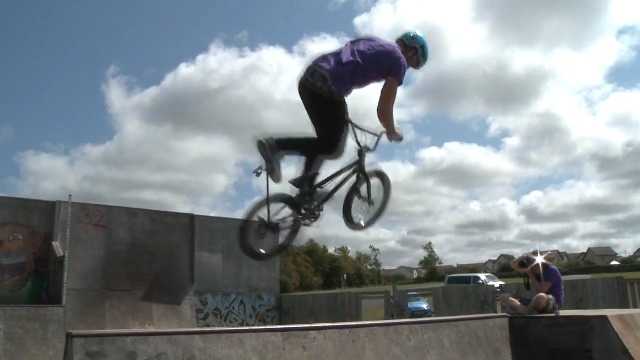} & {\footnotesize{}\hspace{-1em}}\includegraphics[width=0.15\textwidth]{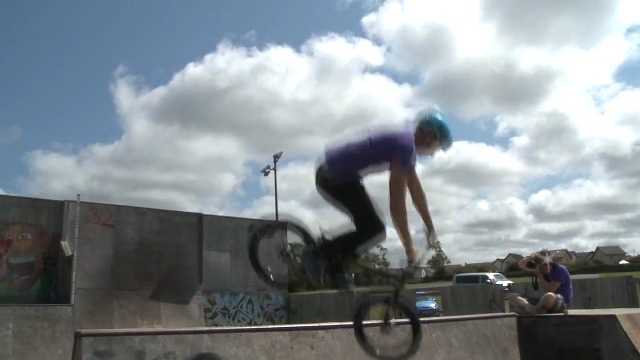}\tabularnewline
 & {\footnotesize{}\hspace{-1em}}\includegraphics[width=0.15\textwidth]{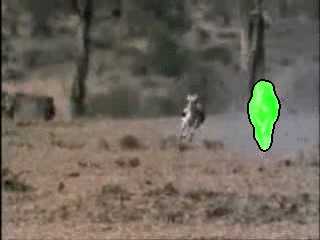} & {\footnotesize{}\hspace{-1em}}\includegraphics[width=0.15\textwidth]{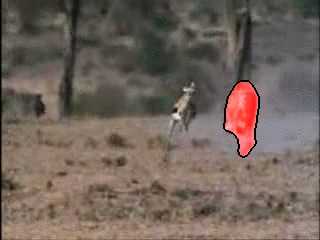} & {\footnotesize{}\hspace{-1em}}\includegraphics[width=0.15\textwidth]{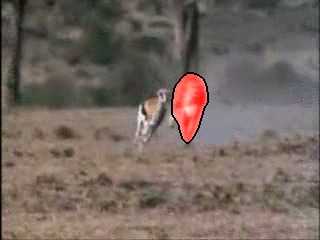} & {\footnotesize{}\hspace{-1em}}\includegraphics[width=0.15\textwidth]{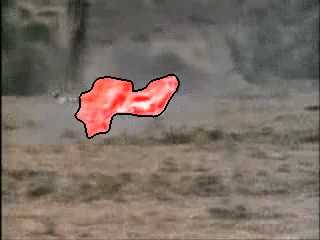} & {\footnotesize{}\hspace{-1em}}\includegraphics[width=0.15\textwidth]{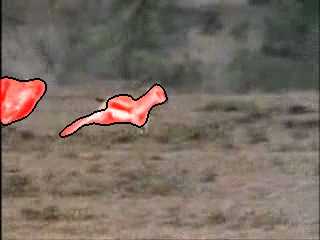} & {\footnotesize{}\hspace{-1em}}\includegraphics[width=0.15\textwidth]{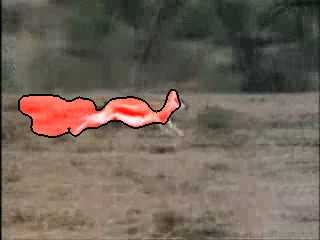}\tabularnewline
 & {\footnotesize{}\hspace{-1em}1st frame, GT segment} & {\footnotesize{}\hspace{-1em}$20\%$} & {\footnotesize{}\hspace{-1em}$40\%$} & {\footnotesize{}\hspace{-1em}$60\%$} & {\footnotesize{}\hspace{-1em}$80\%$} & {\footnotesize{}\hspace{-1em}$100\%$}\tabularnewline
\end{tabular}\hfill{}
\par\end{centering}
\medskip{}

\caption{\label{fig:fail_cases}Failure cases. Frames sampled along the video
duration (e.g. $50\%$: video middle point). For each dataset we show
2 out of 5 worst results (based on mIoU over the video). }
\end{figure*}

However, by looking at per-frame results for each video (blue dots
in Figure \ref{fig:Per-sequence-results-davis16}) one can see several
frames where our approach has failed (IoU less than $50$) to correctly
track the object. Investigating closely those cases we notice conditions
where $\mathtt{Lucid}$\-$\mathtt{Tracker}$ is more likely to fail.
The same behaviour was observed across all three datasets. A few representatives
of failure cases are visualized in Figure \ref{fig:fail_cases}.

Since we are using only the mask annotation of the first frame for training
the tracker, a clear failure case is caused by dramatic view point
changes of the object from its first frame appearance, as in row 5
of Figure \ref{fig:fail_cases}. 
\textcolor{black}{Performing online adaptation every certain time step while exploiting the previous frame segments for data synthesis and marking unsure regions as ignore for training, similarly to \cite{Voigtlaender2017OnlineAO}, might resolve the potential problems caused by relying only on the first frame mask.}
The proposed approach also under-performs
when recovering from occlusions: it might takes several frames for the full
object mask to re-appear (rows 1-3 in Figure \ref{fig:fail_cases}).
This is mainly due to the convnet having learnt to follow-up the previous
frame mask. Augmenting the lucid dreams with plausible occlusions
might help mitigate this case. Another failure case occurs when two
similar looking objects cross each other, as in row 6 in Figure \ref{fig:fail_cases}.
Here both cues: the previous frame guidance and learnt via per-video
tuning appearance, are no longer discriminative to correctly continue
propagating the mask.

We also observe that the $\mathtt{Lucid}$\-$\mathtt{Tracker}$ struggles
to track the fine structures or details of the object, e.g. wheels
of the bicycle or motorcycle in rows 1-2 in Figure \ref{fig:fail_cases}.
This is the issue of the underlying choice of the convnet architecture,
due to the several pooling layers the spatial resolution is lost and
hence the fine details of the object are missing. This issue can be
mitigated by switching to more recent semantic labelling architectures
(e.g. \cite{pohlen2017FRRN,ChenPSA17}).


\begin{table}
\setlength{\tabcolsep}{0.07em} 
\renewcommand{\arraystretch}{1}

\hspace*{\fill}%
\begin{tabular}{c|c|c|c|c|c}
\multirow{2}{*}{{\small{}Attribute}} & \multicolumn{5}{c}{{\small{}Method}}\tabularnewline
 & {\scriptsize{}BVS\,\cite{Maerki2016Cvpr}} & {\scriptsize{}ObjFlow\,\cite{Tsai2016Cvpr}} & {\scriptsize{}OSVOS\,\cite{Caelles2017Cvpr}} & {\scriptsize{}MaskTrack\,\cite{Khoreva2017CvprMaskTrack}} & {\scriptsize{}$\text{LucidTracker}$}\tabularnewline
\hline 
{\scriptsize{}Appearance change} & {\small{}0.46} & {\small{}0.54} & {\textit{\small{}0.81}} & {\small{}0.76} & \textcolor{black}{\textbf{\small{}0.84}}\tabularnewline
{\scriptsize{}Background clutter} & {\small{}0.63} & {\small{}0.68} & \textit{\small{}0.83} & {\small{}0.79} & \textcolor{black}{\textbf{\small{}0.86}}\tabularnewline
{\scriptsize{}Camera-shake} & {\small{}0.62} & {\small{}0.72} & \textit{\small{}0.78} & {\small{}0.78} & \textcolor{black}{\textbf{\small{}0.88}}\tabularnewline
{\scriptsize{}Deformation} & {\small{}0.7} & {\small{}0.77} & \textit{\small{}0.79} & {\small{}0.78} & \textcolor{black}{\textbf{\small{}0.87}}\tabularnewline
{\scriptsize{}Dynamic background} & {\small{}0.6} & {\small{}0.67} & \textit{\small{}0.74} & {\small{}0.76} & \textcolor{black}{\textbf{\small{}0.82}}\tabularnewline
{\scriptsize{}Edge ambiguity} & {\small{}0.58} & {\small{}0.65} & \textit{\small{}0.77} & {\small{}0.74} & \textcolor{black}{\textbf{\small{}0.82}}\tabularnewline
{\scriptsize{}Fast-motion} & {\small{}0.53} & {\small{}0.55} & \textit{\small{}0.76} & {\small{}0.75} & \textcolor{black}{\textbf{\small{}0.85}}\tabularnewline
{\scriptsize{}Heterogeneous object} & {\small{}0.63} & {\small{}0.66} & \textit{\small{}0.75} & {\small{}0.79} & \textcolor{black}{\textbf{\small{}0.85}}\tabularnewline
{\scriptsize{}Interacting objects} & {\small{}0.63} & {\small{}0.68} & \textit{\small{}0.75} & {\small{}0.77} & \textcolor{black}{\textbf{\small{}0.85}}\tabularnewline
{\scriptsize{}Low resolution} & {\small{}0.59} & {\small{}0.58} & {\textit{\small{}0.77}} & {\small{}0.77} & \textcolor{black}{\textbf{\small{}0.84}}\tabularnewline
{\scriptsize{}Motion blur} & {\small{}0.58} & {\small{}0.6} & \textit{\small{}0.74} & {\small{}0.74} & \textcolor{black}{\textbf{\small{}0.83}}\tabularnewline
{\scriptsize{}Occlusion} & {\small{}0.68} & {\small{}0.66} & \textit{\small{}0.77} & {\small{}0.77} & \textcolor{black}{\textbf{\small{}0.84}}\tabularnewline
{\scriptsize{}Out-of-view} & {\small{}0.43} & {\small{}0.53} & \textit{\small{}0.72} & {\small{}0.71} & \textcolor{black}{\textbf{\small{}0.84}}\tabularnewline
{\scriptsize{}Scale variation} & {\small{}0.49} & {\small{}0.56} & \textit{\small{}0.74} & {\small{}0.73} & \textcolor{black}{\textbf{\small{}0.81}}\tabularnewline
{\scriptsize{}Shape complexity} & {\small{}0.67} & {\small{}0.69} & \textit{\small{}0.71} & {\small{}0.75} & \textcolor{black}{\textbf{\small{}0.82}}\tabularnewline
\end{tabular}\hspace*{\fill}

\medskip{}
\caption{\label{tab:Attribute-evaluation}$\text{DAVIS}_{\text{16}}$ per-attribute
evaluation. $\text{LucidTracker}$ improves across all video object segmentation
challenges.}
\end{table}

\paragraph{Conclusion}

$\mathtt{LucidTracker}$ shows robust performance ac\-ross different
videos. However, a few failure cases were observed due to the underlying
convnet architecture, its training, or limited visibility of the object
in the first frame.

\section{\label{sec:Multiple-object-results}Multiple object segmentation results}

We present here an empirical evaluation of LucidTracker for multiple
object segmentation task: given a first frame labelled with the masks
of several object instances, one aims to find the corresponding masks
of objects in future frames. 
\begin{figure*}
\begin{centering}
\setlength{\tabcolsep}{0em} 
\renewcommand{\arraystretch}{0}
\par\end{centering}
\begin{centering}
\vspace{0em}
\hfill{}%
\begin{tabular}{cccccc}
{\footnotesize{}\hspace{-1em}}\includegraphics[width=0.15\textwidth]{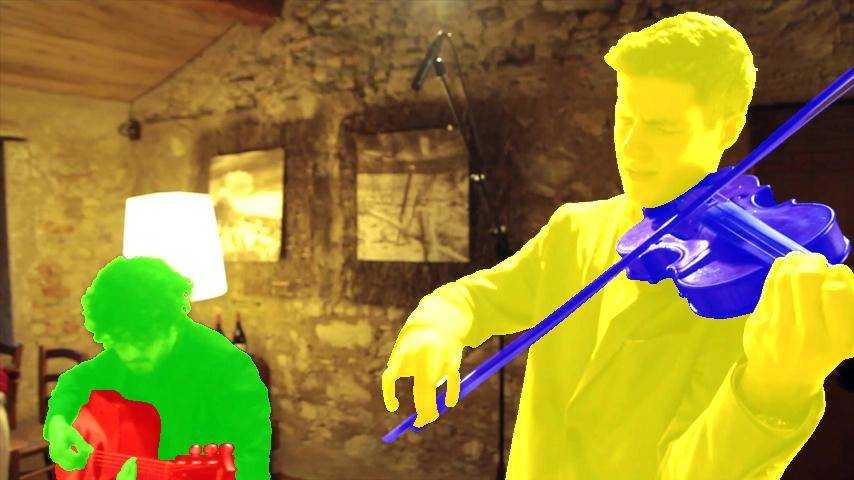} & {\footnotesize{}\hspace{-1em}}\includegraphics[width=0.15\textwidth]{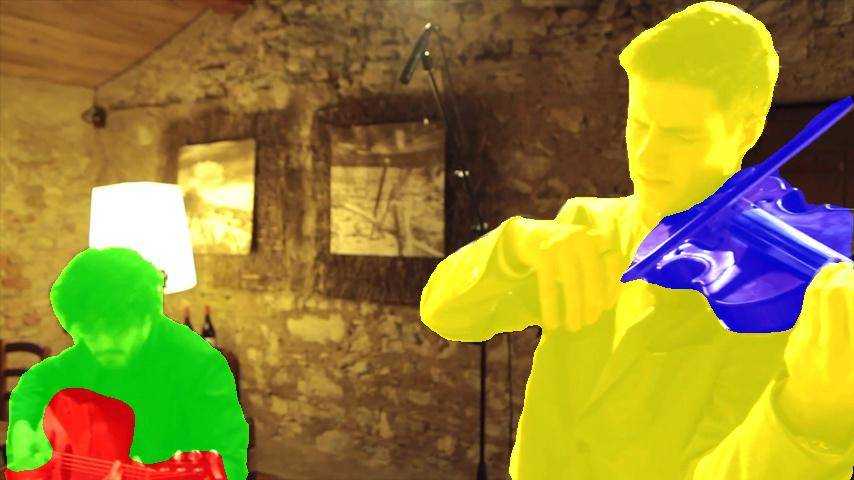} & {\footnotesize{}\hspace{-1em}}\includegraphics[width=0.15\textwidth]{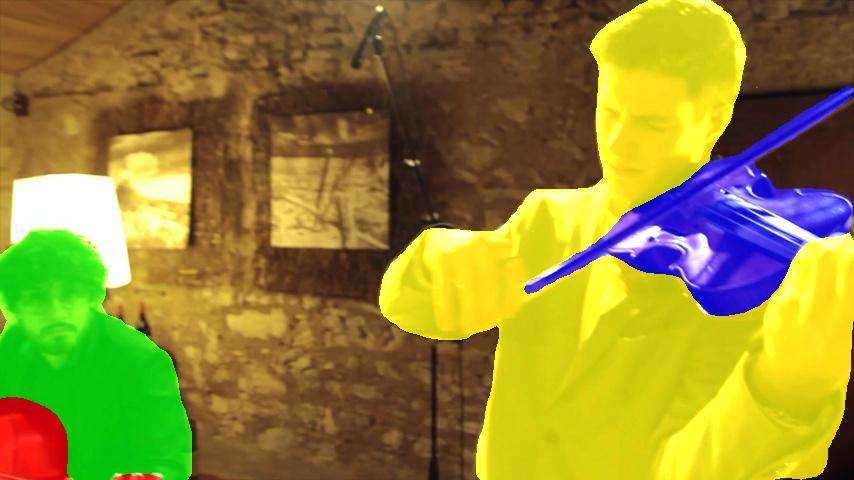} & {\footnotesize{}\hspace{-1em}}\includegraphics[width=0.15\textwidth]{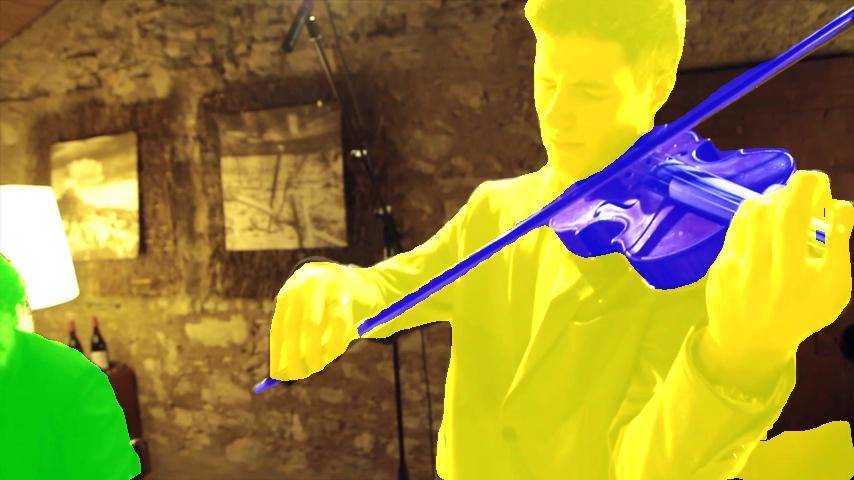} & {\footnotesize{}\hspace{-1em}}\includegraphics[width=0.15\textwidth]{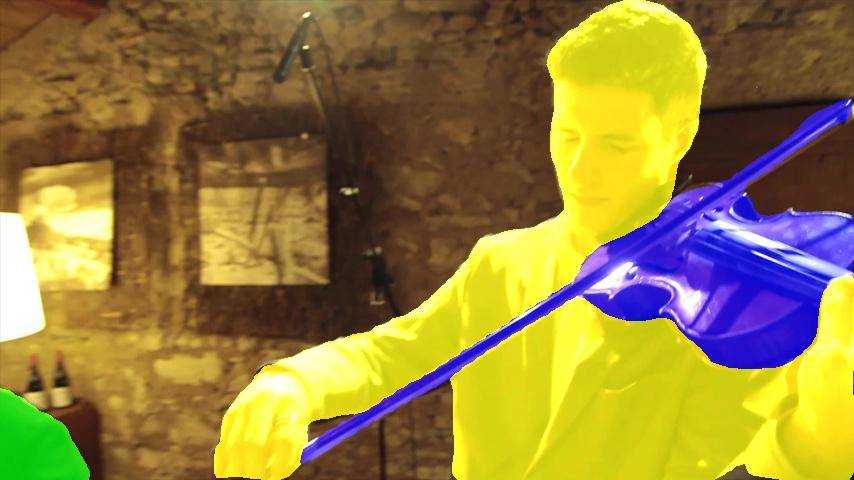} & {\footnotesize{}\hspace{-1em}}\includegraphics[width=0.15\textwidth]{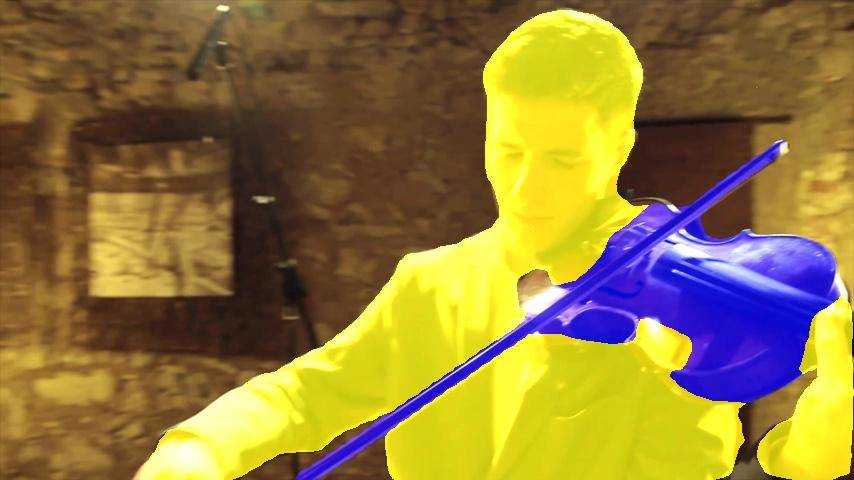}\tabularnewline
{\footnotesize{}\hspace{-1em}}\includegraphics[width=0.15\textwidth]{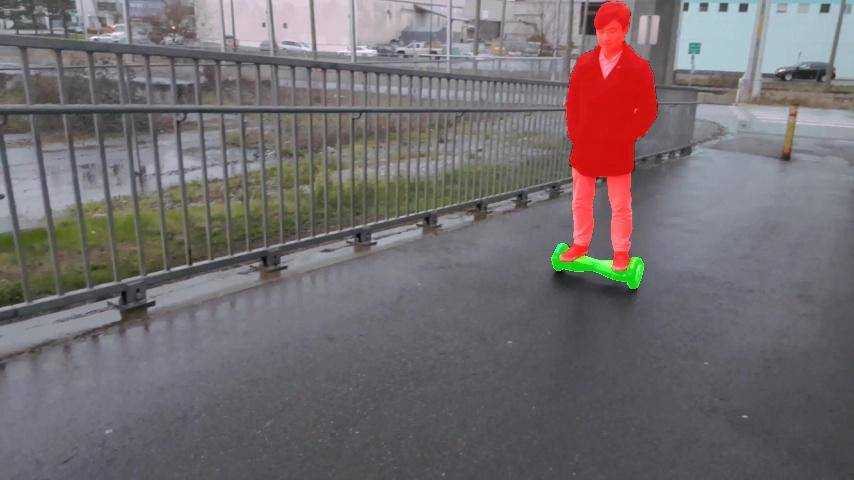} & {\footnotesize{}\hspace{-1em}}\includegraphics[width=0.15\textwidth]{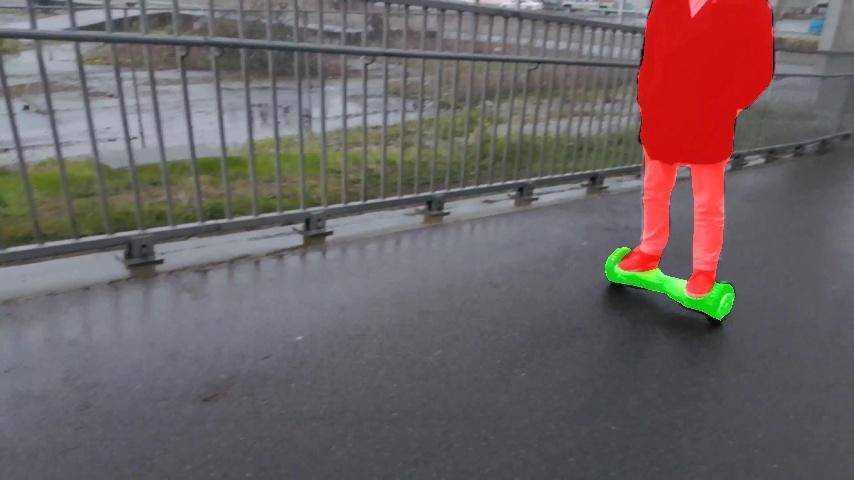} & {\footnotesize{}\hspace{-1em}}\includegraphics[width=0.15\textwidth]{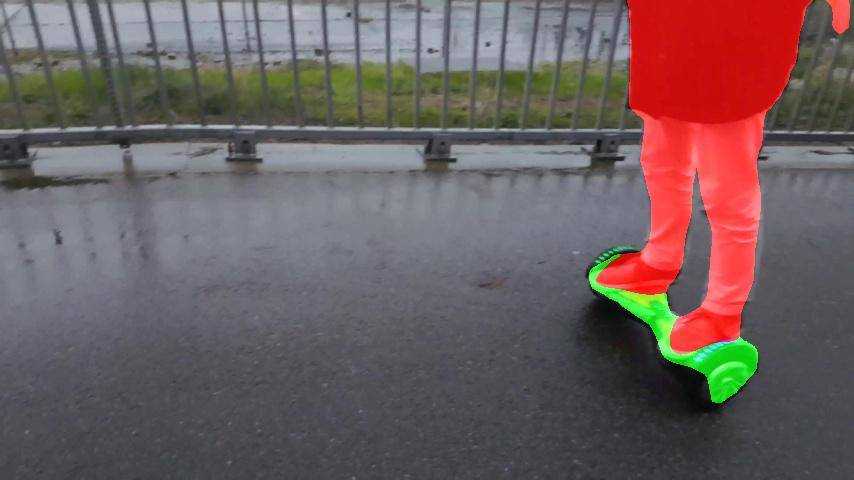} & {\footnotesize{}\hspace{-1em}}\includegraphics[width=0.15\textwidth]{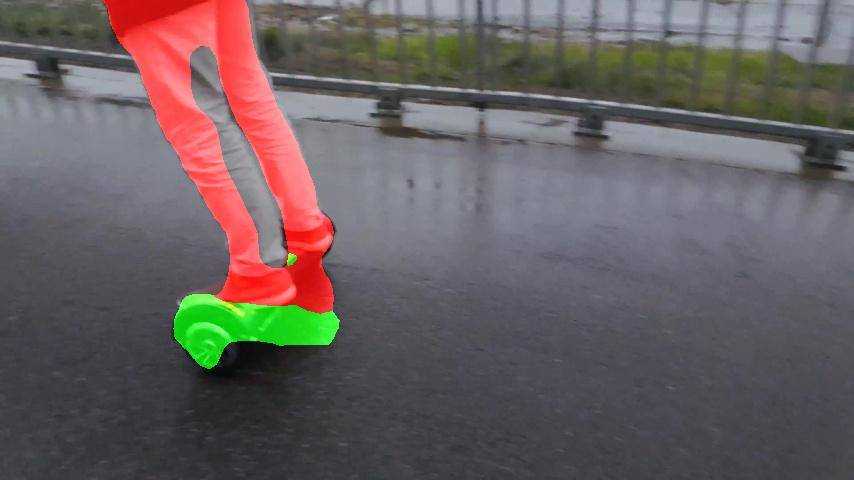} & {\footnotesize{}\hspace{-1em}}\includegraphics[width=0.15\textwidth]{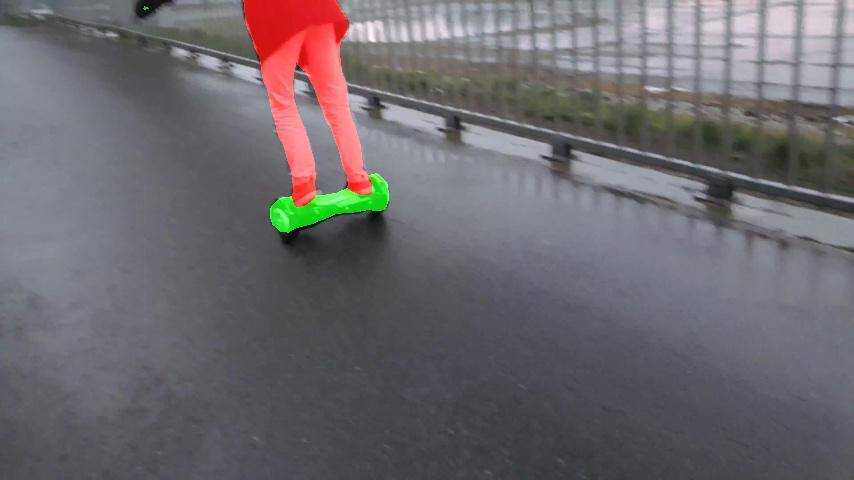} & {\footnotesize{}\hspace{-1em}}\includegraphics[width=0.15\textwidth]{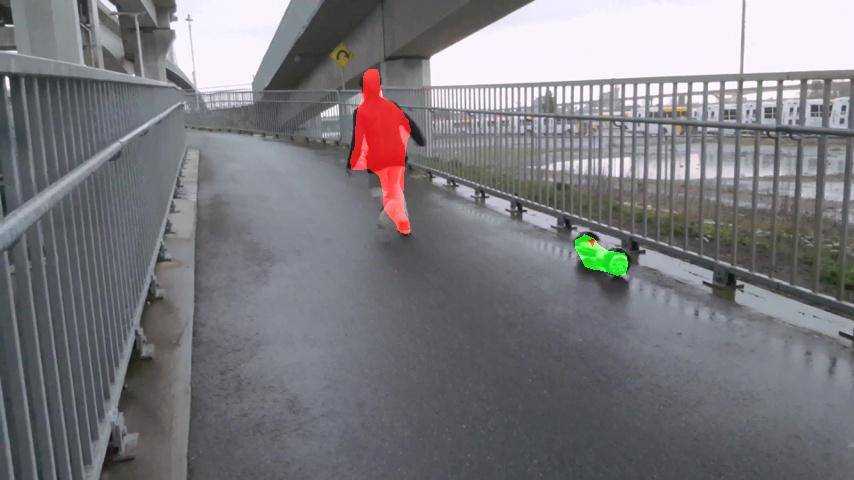}\tabularnewline
{\footnotesize{}\hspace{-1em}}\includegraphics[width=0.15\textwidth]{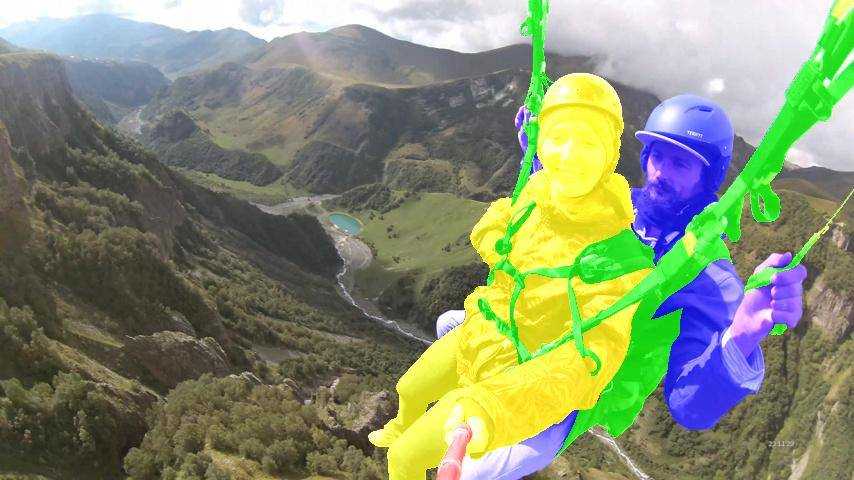} & {\footnotesize{}\hspace{-1em}}\includegraphics[width=0.15\textwidth]{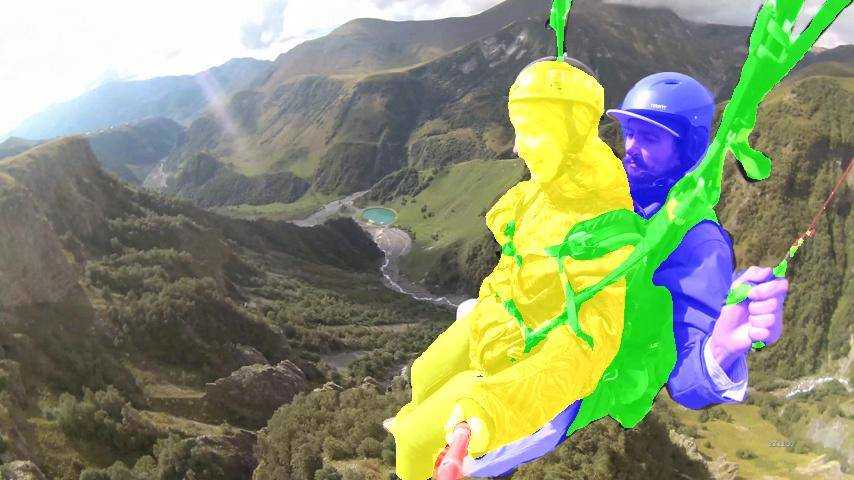} & {\footnotesize{}\hspace{-1em}}\includegraphics[width=0.15\textwidth]{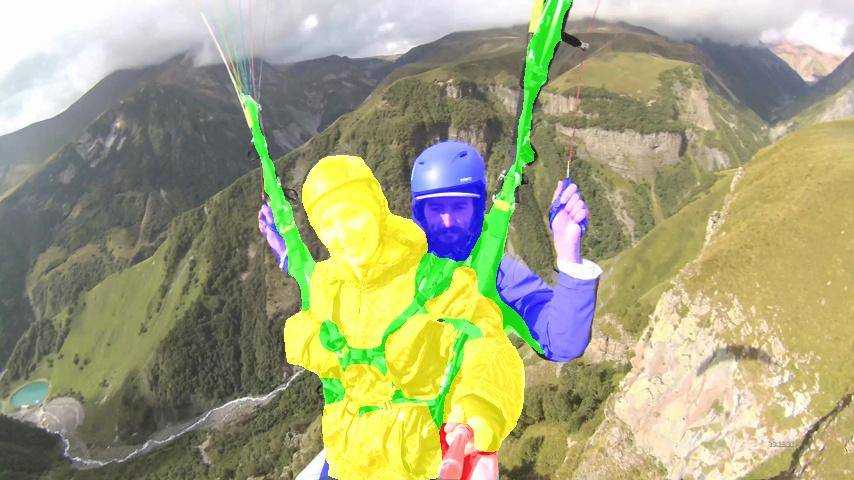} & {\footnotesize{}\hspace{-1em}}\includegraphics[width=0.15\textwidth]{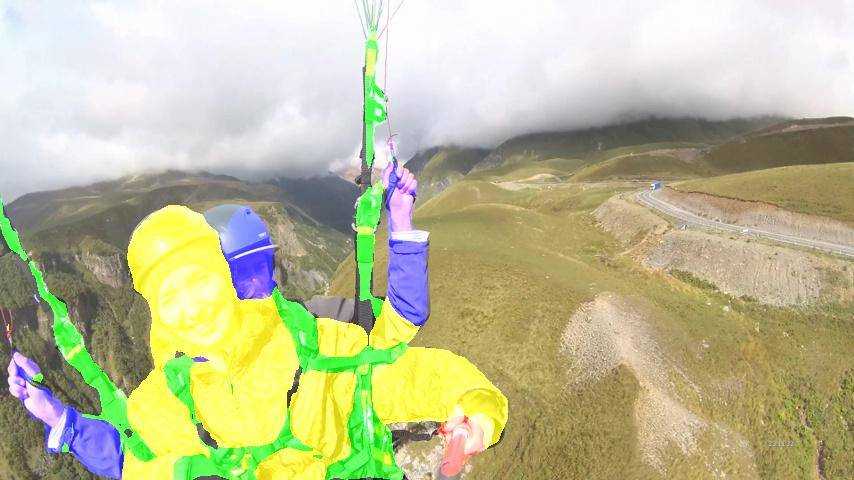} & {\footnotesize{}\hspace{-1em}}\includegraphics[width=0.15\textwidth]{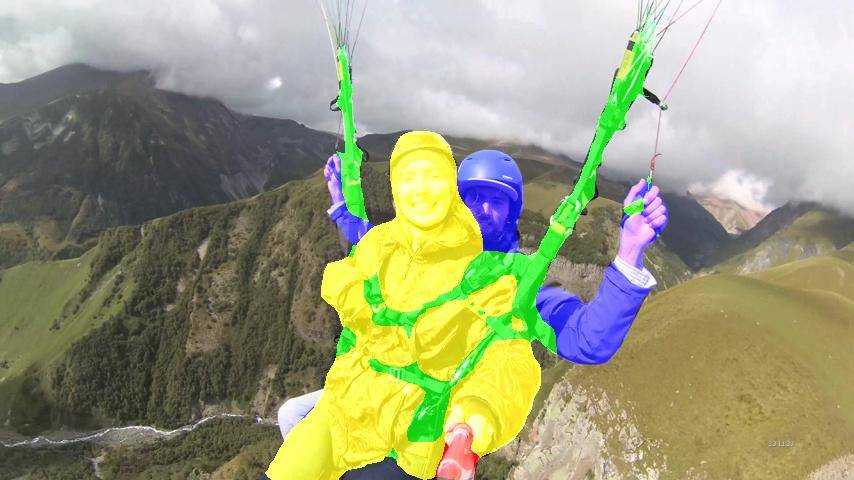} & {\footnotesize{}\hspace{-1em}}\includegraphics[width=0.15\textwidth]{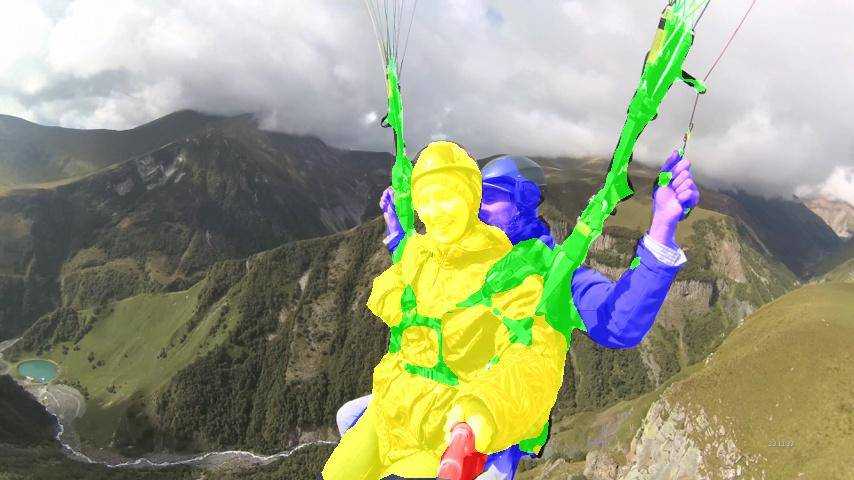}\tabularnewline
{\footnotesize{}\hspace{-1em}}\includegraphics[width=0.15\textwidth]{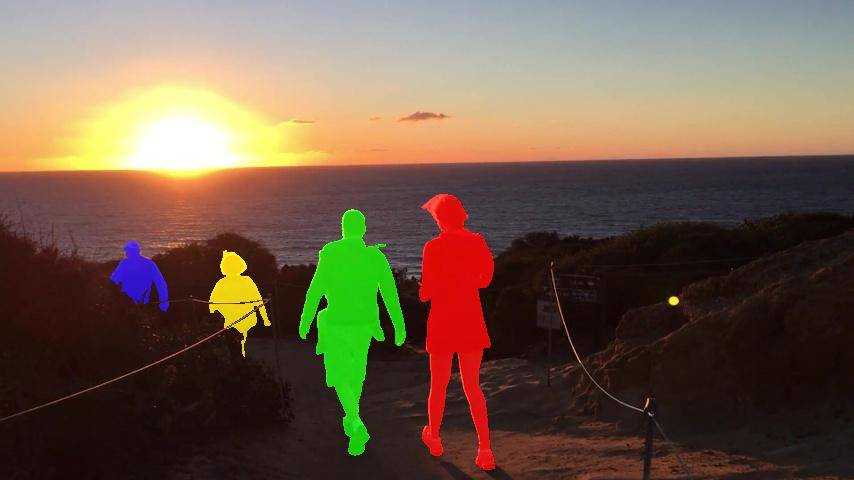} & {\footnotesize{}\hspace{-1em}}\includegraphics[width=0.15\textwidth]{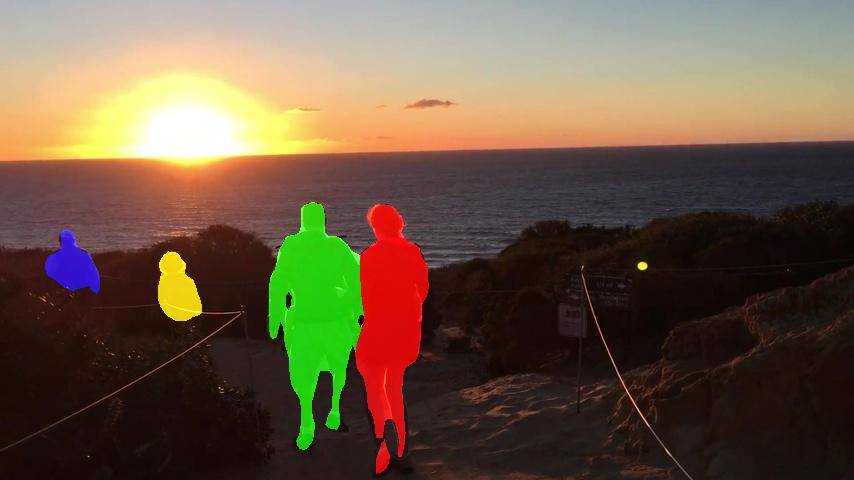} & {\footnotesize{}\hspace{-1em}}\includegraphics[width=0.15\textwidth]{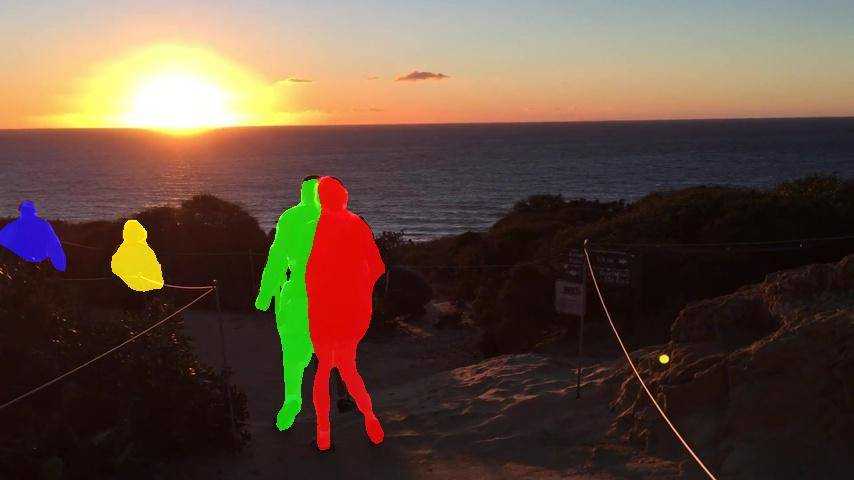} & {\footnotesize{}\hspace{-1em}}\includegraphics[width=0.15\textwidth]{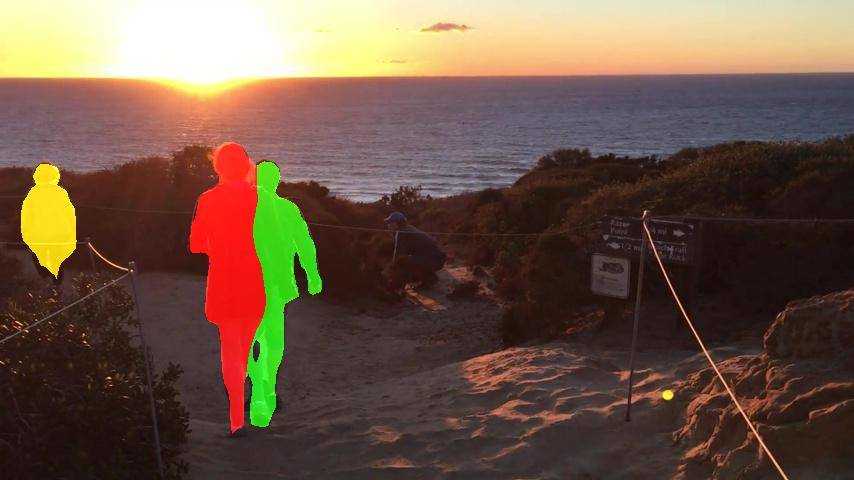} & {\footnotesize{}\hspace{-1em}}\includegraphics[width=0.15\textwidth]{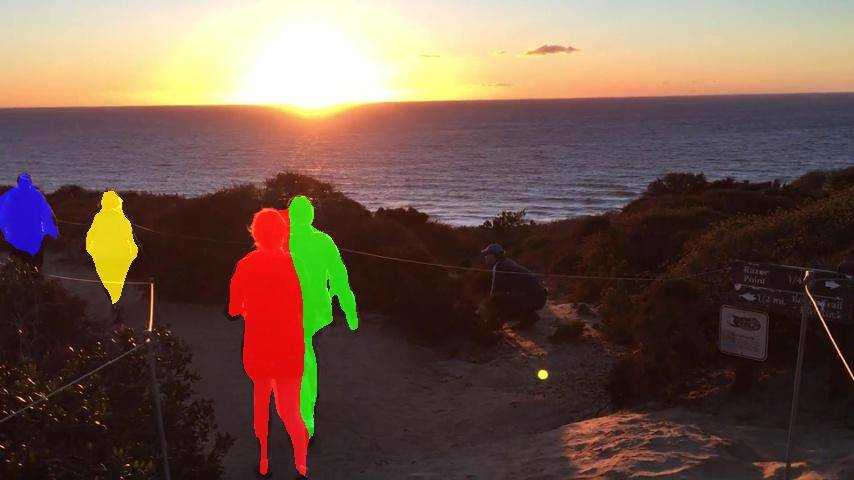} & {\footnotesize{}\hspace{-1em}}\includegraphics[width=0.15\textwidth]{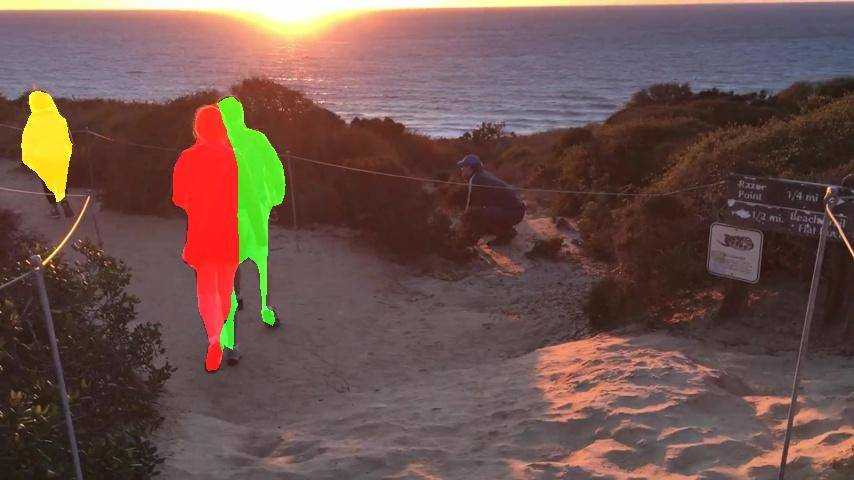}\tabularnewline
{\footnotesize{}\hspace{-1em}}\includegraphics[width=0.15\textwidth]{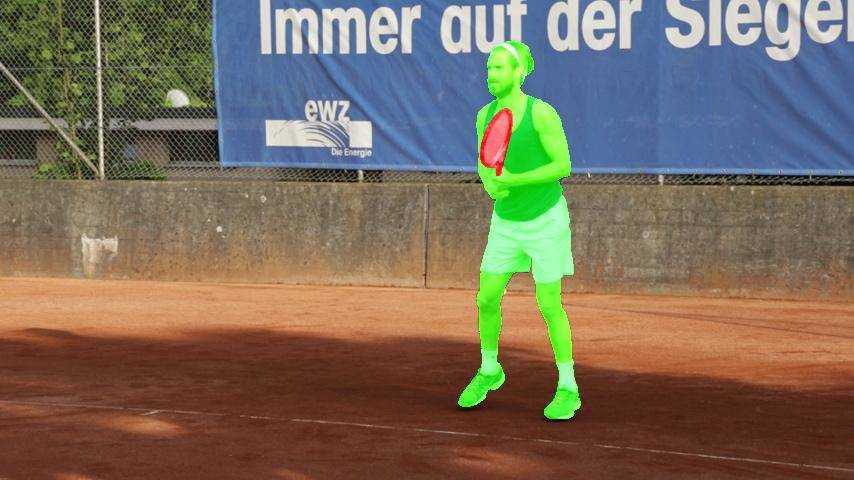} & {\footnotesize{}\hspace{-1em}}\includegraphics[width=0.15\textwidth]{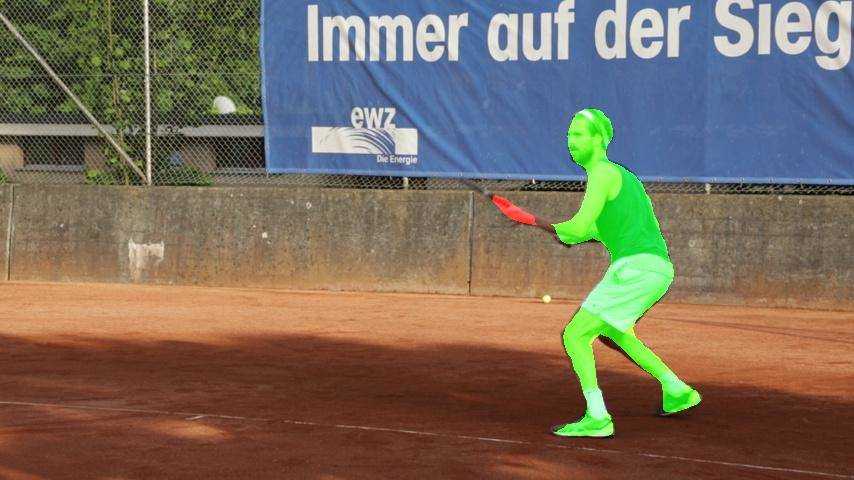} & {\footnotesize{}\hspace{-1em}}\includegraphics[width=0.15\textwidth]{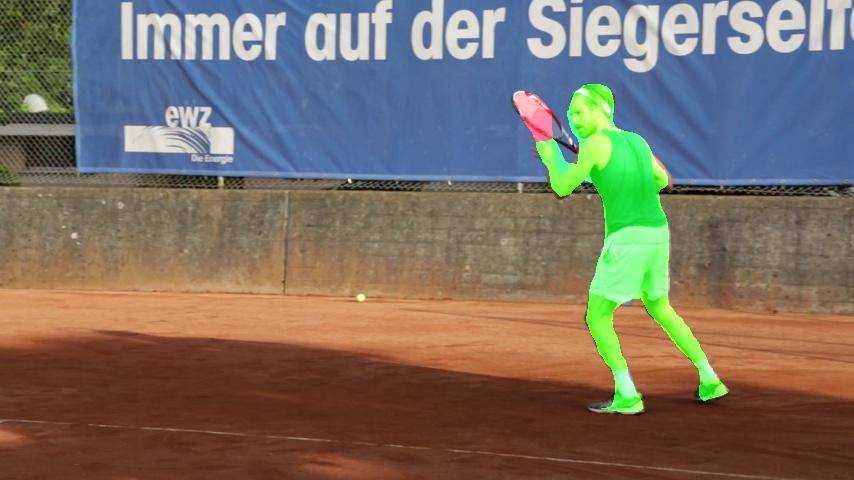} & {\footnotesize{}\hspace{-1em}}\includegraphics[width=0.15\textwidth]{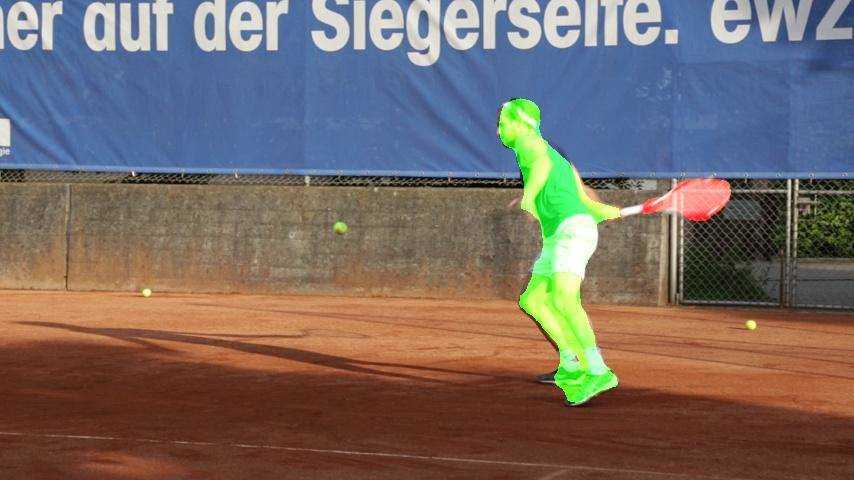} & {\footnotesize{}\hspace{-1em}}\includegraphics[width=0.15\textwidth]{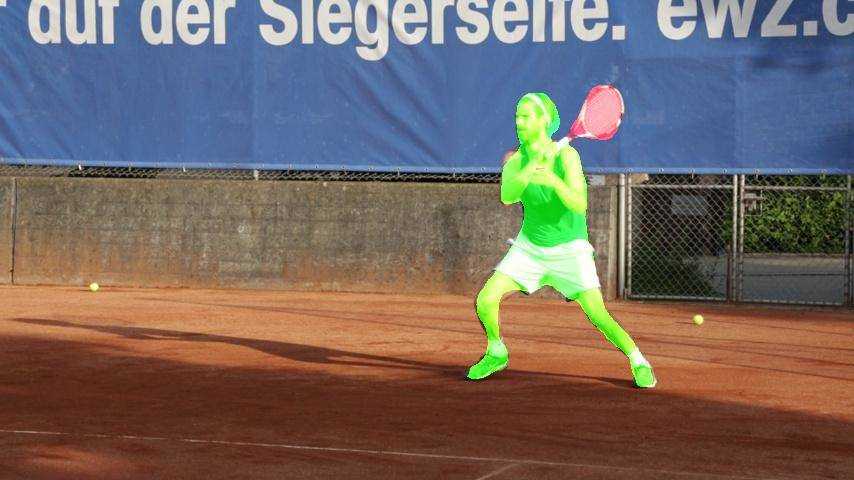} & {\footnotesize{}\hspace{-1em}}\includegraphics[width=0.15\textwidth]{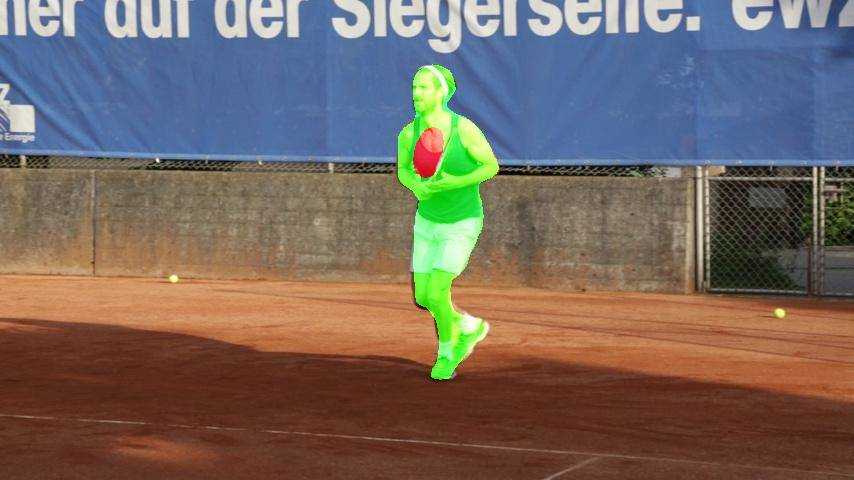}\tabularnewline
{\footnotesize{}\hspace{-1em}1st frame, GT segment} & {\footnotesize{}\hspace{-1em}$20\%$} & {\footnotesize{}\hspace{-1em}$40\%$} & {\footnotesize{}\hspace{-1em}$60\%$} & {\footnotesize{}\hspace{-1em}$80\%$} & {\footnotesize{}\hspace{-1em}$100\%$}\tabularnewline
\end{tabular}\hfill{}
\par\end{centering}
\caption{\label{fig:qualitative-results-mult}LucidTracker qualitative results
on $\text{DAVIS}_{\text{17}}$, test-dev set. Frames sampled along
the video duration (e.g. $50\%$: video middle point). The videos
are chosen with the highest mIoU measure. }
\vspace{0em}
\end{figure*}

\subsection{\label{subsec:Experimental-setup-mult}Experimental setup}

\paragraph{Dataset}

For the multiple object segmentation task we use the 2017 DAVIS Challenge on Video
Object Segmentation\footnote{http://davischallenge.org/challenge2017}
\cite{Pont-Tuset_arXiv_2017} ($\text{DAVIS}_{\text{17}}$). Compared
to $\text{DAVIS}_{\text{16}}$ this is a larger, more challenging
dataset, where the video sequences have multiple objects in the scene.
Videos that have more than one visible object in $\text{DAVIS}_{\text{16}}$
have been re-annotated (the objects were divided by semantics) and
the train and val sets were extended with more sequences. In addition,
two other test sets (test-dev and test-challenge) were introduced.
The complexity of the videos has increased with more distractors,
occlusions, fast motion, smaller objects, and fine structures. Overall,
$\text{DAVIS}_{\text{17}}$ consists of $150$ sequences, totalling
$10\,474$ annotated frames and $384$ objects.

We evaluate our method on two test sets, the test-dev and test-challenge
sets, each consists of $30$ video sequences, on average $\sim3$
objects per sequence, the length of the sequences is $\sim70$ frames.
For both test sets only the masks on the first frames are made public,
the evaluation is done via an evaluation server. 
Our experiments and ablation studies are done on the test-dev set. 

\paragraph{Evaluation metric}

The accuracy of multiple object segmentation is evaluated using the region
(J) and boundary (F) measures proposed by the organisers of the challenge.
The average of J and F measures is used as overall performance score (denoted as global mean in the tables).
Please refer to \cite{Pont-Tuset_arXiv_2017} for more details about
the evaluation protocol.

\paragraph{Training details}

All experiments in this section are done using the single stream architecture
discussed in sections \ref{subsec:Architecture} and \ref{subsubsec:convnet-arch}.
For training the models we use SGD with mini-batches of $10$ images
and a fixed learning policy with initial learning rate of $10^{-3}$.
The momentum and weight decay are set to $0.9$ and $5\cdot10^{-4}$,
respectively. All models are initialized with weights trained for
image classification on ImageNet \cite{Simonyan2015Iclr}. We then
train per-video for 40k iterations. 

\subsection{\label{subsec:Key-results-mult}Key results}

\begin{table*}
\setlength{\tabcolsep}{0.5em} 
\renewcommand{\arraystretch}{1.3}
\begin{centering}
\hspace*{\fill}%
\begin{tabular}{c|cc|ccc|ccc}
\multirow{3}{*}{{\footnotesize{}Method}} & \multicolumn{8}{c}{{\small{}$\text{DAVIS}_{\text{17}}$, test-dev set}}\tabularnewline
\cline{2-9} 
 & \multirow{2}{*}{{\footnotesize{}Rank}} & \multirow{2}{*}{{\footnotesize{}}%
\begin{tabular}{c}
{\footnotesize{}Global}\tabularnewline
{\footnotesize{}mean }\tabularnewline
\end{tabular}{\footnotesize{}$\uparrow$}} & \multicolumn{3}{c|}{{\footnotesize{}Region,}{\small{} \ensuremath{J} }} & \multicolumn{3}{c}{{\footnotesize{}Boundary,}{\small{} \ensuremath{F} }}\tabularnewline
\cline{4-9} 
 &  &  & {\footnotesize{}Mean $\uparrow$} & {\footnotesize{}Recall $\uparrow$} & {\footnotesize{}Decay $\downarrow$} & {\footnotesize{}Mean $\uparrow$} & {\footnotesize{}Recall $\uparrow$} & {\footnotesize{}Decay $\downarrow$}\tabularnewline
\hline 
\hline 
sidc & {\small{}10} & {\small{}45.8} & {\small{}43.9} & {\small{}51.5} & {\small{}34.3} & {\small{}47.8} & {\small{}53.6} & {\small{}36.9}\tabularnewline
YXLKJ & {\small{}9} & {\small{}49.6} & {\small{}46.1} & {\small{}49.1} & {\small{}22.7} & {\small{}53.0} & {\small{}56.5} & {\small{}22.3}\tabularnewline
haamooon \cite{DAVIS2017-4th} & {\small{}8} & {\small{}51.3} & {\small{}48.8} & {\small{}56.9} & \textbf{\small{}12.2} & {\small{}53.8} & {\small{}61.3} & \textbf{\small{}11.8}\tabularnewline
Fromandtozh \cite{DAVIS2017-7th} & {\small{}7} & {\small{}55.2} & {\small{}52.4} & {\small{}58.4} & {\small{}18.1} & {\small{}57.9} & {\small{}66.1} & {\small{}20.0}\tabularnewline
ilanv \cite{DAVIS2017-9th} & {\small{}6} & {\small{}55.8} & {\small{}51.9} & {\small{}55.7} & {\small{}17.6} & {\small{}59.8} & {\small{}65.8} & {\small{}18.9}\tabularnewline
voigtlaender \cite{DAVIS2017-5th} & {\small{}5} & {\small{}56.5} & {\small{}53.4} & {\small{}57.8} & {\small{}19.9} & {\small{}59.6} & {\small{}65.4} & {\small{}19.0}\tabularnewline
lalalafine123 & {\small{}4} & {\small{}57.4} & {\small{}54.5} & {\small{}61.3} & {\small{}24.4} & {\small{}60.2} & {\small{}68.8} & {\small{}24.6}\tabularnewline
wangzhe & {\small{}3} & {\small{}57.7} & {\small{}55.6} & {\small{}63.2} & {\small{}31.7} & {\small{}59.8} & {\small{}66.7} & {\small{}37.1}\tabularnewline
lixx \cite{DAVIS2017-1st} & {\small{}2} & {\small{}66.1} & \textbf{\small{}64.4} & {\small{}73.5} & {\small{}24.5} & {\small{}67.8} & {\small{}75.6} & {\small{}27.1}\tabularnewline
\hline 
\arrayrulecolor{black}\textbf{$\text{LucidTracker}$} & 1 & \textbf{\small{}66.6} & {\small{}63.4} & \textbf{\small{}73.9} & {\small{}19.5} & \textbf{\small{}69.9} & \textbf{\small{}80.1} & {\small{}19.4}\tabularnewline
\end{tabular}\hfill{}\medskip{}
\par\end{centering}
\caption{\label{tab:comparative-result-davis-mult}Comparison of video object segmentation
results on $\text{DAVIS}_{\text{17}}$, test-dev set. Our $\text{LucidTracker}$
shows top performance.}

\begin{centering}
\hspace*{\fill}%
\begin{tabular}{c|cc|ccc|ccc}
\multirow{3}{*}{{\footnotesize{}Method}} & \multicolumn{8}{c}{{\small{}$\text{DAVIS}_{\text{17}}$, test-challenge set}}\tabularnewline
\cline{2-9} 
 & \multirow{2}{*}{{\footnotesize{}Rank}} & \multirow{2}{*}{{\footnotesize{}}%
\begin{tabular}{c}
{\footnotesize{}Global}\tabularnewline
{\footnotesize{}mean }\tabularnewline
\end{tabular}{\footnotesize{}$\uparrow$}} & \multicolumn{3}{c|}{{\footnotesize{}Region,}{\small{} \ensuremath{J} }} & \multicolumn{3}{c}{{\footnotesize{}Boundary,}{\small{} \ensuremath{F} }}\tabularnewline
\cline{4-9} 
 &  &  & {\footnotesize{}Mean $\uparrow$} & {\footnotesize{}Recall $\uparrow$} & {\footnotesize{}Decay $\downarrow$} & {\footnotesize{}Mean $\uparrow$} & {\footnotesize{}Recall $\uparrow$} & {\footnotesize{}Decay $\downarrow$}\tabularnewline
\hline 
\hline 
zwrq0 & {\small{}10} & {\small{}53.6} & {\small{}50.5} & {\small{}54.9} & {\small{}28.0} & {\small{}56.7} & {\small{}63.5} & {\small{}30.4}\tabularnewline
Fromandtozh \cite{DAVIS2017-7th} & {\small{}9} & {\small{}53.9} & {\small{}50.7} & {\small{}54.9} & {\small{}32.5} & {\small{}57.1} & {\small{}63.2} & {\small{}33.7}\tabularnewline
wasidennis & {\small{}8} & {\small{}54.8} & {\small{}51.6} & {\small{}56.3} & {\small{}26.8} & {\small{}57.9} & {\small{}64.8} & {\small{}28.8}\tabularnewline
YXLKJ & {\small{}7} & {\small{}55.8} & {\small{}53.8} & {\small{}60.1} & {\small{}37.7} & {\small{}57.8} & {\small{}62.1} & {\small{}42.9}\tabularnewline
cjc \cite{DAVIS2017-6th} & {\small{}6} & {\small{}56.9} & {\small{}53.6} & {\small{}59.5} & {\small{}25.3} & {\small{}60.2} & {\small{}67.9} & {\small{}27.6}\tabularnewline
lalalafine123  & {\small{}6} & {\small{}56.9} & {\small{}54.8} & {\small{}60.7} & {\small{}34.4} & {\small{}59.1} & {\small{}66.7} & {\small{}36.1}\tabularnewline
voigtlaender \cite{DAVIS2017-5th} & {\small{}5} & {\small{}57.7} & {\small{}54.8} & {\small{}60.8} & {\small{}31.0} & {\small{}60.5} & {\small{}67.2} & {\small{}34.7}\tabularnewline
haamooon \cite{DAVIS2017-4th} & {\small{}4} & {\small{}61.5} & {\small{}59.8} & {\small{}71.0} & {\small{}21.9} & {\small{}63.2} & {\small{}74.6} & {\small{}23.7}\tabularnewline
vantam299 \cite{DAVIS2017-3rd} & {\small{}3} & {\small{}63.8} & {\small{}61.5} & {\small{}68.6} & \textbf{\small{}17.1} & {\small{}66.2} & {\small{}79.0} & \textbf{\small{}17.6}\tabularnewline
\hline 
\textbf{$\text{LucidTracker}$} & {\small{}2} & {\small{}67.8} & {\small{}65.1} & {\small{}72.5} & {\small{}27.7} & {\small{}70.6} & \textbf{\small{}79.8} & {\small{}30.2}\tabularnewline
\hline 
lixx \cite{DAVIS2017-1st} & 1 & \textbf{\small{}69.9} & \textbf{\small{}67.9} & \textbf{\small{}74.6} & {\small{}22.5} & \textbf{\small{}71.9} & {\small{}79.1} & {\small{}24.1}\tabularnewline
\end{tabular}\hfill{}\medskip{}
\par\end{centering}
\caption{\label{tab:comparative-result-davis-mult-test-chal}Comparison of
video object segmentation results on $\text{DAVIS}_{\text{17}}$, test-challenge
set. Our $\text{LucidTracker}$ shows competitive performance, holding
the second place in the competition.}
\end{table*}

Tables \ref{tab:comparative-result-davis-mult} and \ref{tab:comparative-result-davis-mult-test-chal}
presents the results of the 2017 DAVIS Challenge on test-dev and test-challenge
sets \cite{Davis2017Challenge}.

Our main results for the multi-object segmentation challenge are obtained
via an ensemble of four different models ($f_{\mathcal{I}}$, $f_{\mathcal{I{\scriptscriptstyle +}F}}$,
$f_{\mathcal{I{\scriptscriptstyle +}S}}$, $f_{\mathcal{I{\scriptscriptstyle +}F{\scriptscriptstyle +}\mathcal{S}}}$),
see Section \ref{subsec:Architecture}.

The proposed system, $\mathtt{Lucid}$\-$\mathtt{Tracker}$, provides
the best segmentation quality on the test-dev set and shows competitive
performance on the test-challenge set, holding the second place in
the competition. The full system is trained using the standard ImageNet
pre-training initialization, Pascal VOC12 semantic annotations for
the $\mathcal{\mathcal{S}}_{t}$ input ($\sim\negmedspace10k$ annotated
images), and one annotated frame per test video, $30$ frames total
on each test set. As discussed in Section \ref{subsec:Ablation-study-mult},
even without $\mathcal{\mathcal{S}}_{t}$ $\mathtt{Lucid}$\-$\mathtt{Tracker}$
obtains competitive results (less than $1$ percent point difference, see Table \ref{tab:ablation-challenge} for details).

The top entry $\mathtt{lixx}$ \cite{DAVIS2017-1st} uses a deeper
convnet model (ImageNet pre-trained ResNet), a similar segmentation architecture, trains it over external segmentation data (using
$\sim\negmedspace120k$ pixel-level annotated images from MS-COCO
and Pascal VOC for pre-training, and akin to {\cite{Caelles2017Cvpr}}
fine-tuning on the $\text{DAVIS}_{\text{17}}$ train and val sets,
$\sim\negmedspace10k$ annotated frames), and extends it with a box-level
object detector (trained over MS-COCO and Pascal VOC, $\sim\negmedspace500k$
bounding boxes) and a box-level object re-identification model trained
over $\sim\negmedspace60k$ box annotations (on both images and videos).
We argue that our system reaches comparable results with a significantly
lower amount of training data.

Figure \ref{fig:qualitative-results-mult} provides qualitative results
of $\mathtt{Lucid}$\-$\mathtt{Tracker}$ on the test-dev set. The
video results include successful handling of multiple objects, full
and partial occlusions, distractors, small objects, and out-of-view
scenarios.

\begin{table*}
	\setlength{\tabcolsep}{0.5em} 
	\renewcommand{\arraystretch}{1.3}
	\begin{centering}
		\bigskip{}
		\begin{tabular}{c|ccc|c|c|c|ccc|ccc}
			\multirow{3}{*}{{\small{}Variant}} & \multirow{3}{*}{{\scriptsize{}$\mathcal{I}$}} & \multirow{3}{*}{{\scriptsize{}$\mathcal{F}$}} & \multirow{3}{*}{{\scriptsize{}$\mathcal{S}$}} & \multirow{3}{*}{{\scriptsize{}ensemble}} & \multirow{3}{*}{{\scriptsize{}CRF tuning}}  & \multirow{3}{*}{{\scriptsize{}temp. coherency}} & \multicolumn{6}{c}{{\scriptsize{}$\text{DAVIS}_{\text{17}}$}}\tabularnewline
			&  &  &  &  &  &  &\multicolumn{3}{c|}{{\scriptsize{}test-dev}} & \multicolumn{3}{c}{{\scriptsize{}test-challenge}}\tabularnewline
			\cline{8-13} 
			&  &  &  &  &  & & {\scriptsize{}global mean} & {\scriptsize{}mIoU} & {\scriptsize{}mF} & {\scriptsize{}global mean} & {\scriptsize{}mIoU} & {\scriptsize{}mF}\tabularnewline
			\hline 
			\hline 
			\multirow{4}{*}{\textcolor{black}{\small{}}%
				\begin{tabular}{c}
					\textcolor{black}{\small{}$\text{LucidTracker}$}\tabularnewline
					(ensemble)\tabularnewline
			\end{tabular}} & \textcolor{black}{\scriptsize{}\Checkmark{}} & \textcolor{black}{\scriptsize{}\Checkmark{}} & \textcolor{black}{\scriptsize{}\Checkmark{}} & \textcolor{black}{\scriptsize{}\Checkmark{}} & \textcolor{black}{\scriptsize{}\Checkmark{}} & \textcolor{black}{\scriptsize{}\Checkmark{}}  & \textbf{\small{}66.6} & \textbf{\small{}63.4} & \textbf{\small{}69.9} & \textbf{\small{}67.8} & \textbf{\small{}65.1} & \textbf{\small{}70.6}\tabularnewline
			
			& \textcolor{black}{\scriptsize{}\Checkmark{}} & \textcolor{black}{\scriptsize{}\Checkmark{}} & \textcolor{black}{\scriptsize{}\Checkmark{}} & \textcolor{black}{\scriptsize{}\Checkmark{}} & \textcolor{black}{\scriptsize{}\Checkmark{}} & \textcolor{black}{\scriptsize{}\XSolidBrush{}}  & {\small{}65.2} & {\small{}61.5} & {\small{}69.0} & {\small{}67.0} & {\small{}64.3} & {\small{}69.7}\tabularnewline
			
			& \textcolor{black}{\scriptsize{}\Checkmark{}} & \textcolor{black}{\scriptsize{}\Checkmark{}} & \textcolor{black}{\scriptsize{}\Checkmark{}} & \textcolor{black}{\scriptsize{}\Checkmark{}} & \textcolor{black}{\scriptsize{}\XSolidBrush{}} & \textcolor{black}{\scriptsize{}\XSolidBrush{}} & {\small{}64.7} & {\small{}60.5} & {\small{}68.9} & 66.5 & 63.2 & 69.8\tabularnewline
			& \textcolor{black}{\scriptsize{}\Checkmark{}} & \textcolor{black}{\scriptsize{}\Checkmark{}} & \textcolor{black}{\scriptsize{}\XSolidBrush{}} & \textcolor{black}{\scriptsize{}\Checkmark{}} & \textcolor{black}{\scriptsize{}\Checkmark{}} & \textcolor{black}{\scriptsize{}\XSolidBrush{}} & {\small{}64.9} & {\small{}61.3} & {\small{}68.4} & - & - & -\tabularnewline
			& \textcolor{black}{\scriptsize{}\Checkmark{}} & \textcolor{black}{\scriptsize{}\Checkmark{}} & \textcolor{black}{\scriptsize{}\XSolidBrush{}} & \textcolor{black}{\scriptsize{}\Checkmark{}} & \textcolor{black}{\scriptsize{}\XSolidBrush{}} & \textcolor{black}{\scriptsize{}\XSolidBrush{}}  & {\small{}64.2} & {\small{}60.1} & {\small{}68.3} & - & - & -\tabularnewline
			\hline 
			\textcolor{black}{\small{}$\text{LucidTracker}$} & \textcolor{black}{\scriptsize{}\Checkmark{}} & \textcolor{black}{\scriptsize{}\Checkmark{}} & \textcolor{black}{\scriptsize{}\Checkmark{}} & \textcolor{black}{\scriptsize{}\XSolidBrush{}} & \textcolor{black}{\scriptsize{}\Checkmark{}} & \textcolor{black}{\scriptsize{}\XSolidBrush{}} & {\small{}62.9} & {\small{}59.1} & {\small{}66.6} & - & - & -\tabularnewline
			{\scriptsize{}$\mathcal{I}+\mathcal{F}+\mathcal{S}$} & \textcolor{black}{\scriptsize{}\Checkmark{}} & \textcolor{black}{\scriptsize{}\Checkmark{}} & \textcolor{black}{\scriptsize{}\Checkmark{}} & \textcolor{black}{\scriptsize{}\XSolidBrush{}} & \textcolor{black}{\scriptsize{}\XSolidBrush{}}& \textcolor{black}{\scriptsize{}\XSolidBrush{}}  & {\small{}62.0} & {\small{}57.7} & {\small{}62.2} & {\small{}64.0} & {\small{}60.7} & {\small{}67.3}\tabularnewline
			\hline 
			{\scriptsize{}$\mathcal{I}+\mathcal{F}$} & \textcolor{black}{\scriptsize{}\Checkmark{}} & \textcolor{black}{\scriptsize{}\Checkmark{}} & \textcolor{black}{\scriptsize{}\XSolidBrush{}} & \textcolor{black}{\scriptsize{}\XSolidBrush{}} & \textcolor{black}{\scriptsize{}\XSolidBrush{}} & \textcolor{black}{\scriptsize{}\XSolidBrush{}} & {\small{}61.3} & {\small{}56.8} & {\small{}65.8} & - & - & -\tabularnewline
			{\scriptsize{}$\mathcal{I}+\mathcal{S}$} & \textcolor{black}{\scriptsize{}\Checkmark{}} & \textcolor{black}{\scriptsize{}\XSolidBrush{}} & \textcolor{black}{\scriptsize{}\Checkmark{}} & \textcolor{black}{\scriptsize{}\XSolidBrush{}} & \textcolor{black}{\scriptsize{}\XSolidBrush{}} & \textcolor{black}{\scriptsize{}\XSolidBrush{}} & {\small{}61.1} & {\small{}56.9} & {\small{}65.3} & - & - & -\tabularnewline
			{\scriptsize{}$\mathcal{I}$} & \textcolor{black}{\scriptsize{}\Checkmark{}} & \textcolor{black}{\scriptsize{}\XSolidBrush{}} & \textcolor{black}{\scriptsize{}\XSolidBrush{}} & \textcolor{black}{\scriptsize{}\XSolidBrush{}} & \textcolor{black}{\scriptsize{}\XSolidBrush{}} & \textcolor{black}{\scriptsize{}\XSolidBrush{}}  & {\small{}59.8} & {\small{}63.1} & {\small{}63.9} & - & - & -\tabularnewline
		\end{tabular}\medskip{}
		\par\end{centering}
	\caption{\label{tab:ablation-challenge}Ablation study of different ingredients.
		$\text{DAVIS}_{\text{17}}$, test-dev and test challenge sets.}
	\vspace{0em}
\end{table*}

\paragraph{Conclusion}

We show that top results for multiple object segmentation can be achieved
via our approach that focuses on exploiting as much as possible the
available annotation on the first video frame, rather than relying
heavily on large external training data.

\subsection{\label{subsec:Ablation-study-mult}Ablation study}

Table \ref{tab:ablation-challenge} explores in more details how the
different ingredients contribute to our results.

We see that adding extra information (channels) to the system, either
optical flow magnitude or semantic segmentation, or both, does provide
$1\negmedspace\sim\negmedspace2$ percent point improvement. The results
show that leveraging semantic priors and motion information provides
a complementary signal to RGB image and both ingredients contribute
to the segmentation results.

Combining in ensemble four different models ($f_{\mathcal{I{\scriptscriptstyle +}F{\scriptscriptstyle +}\mathcal{S}}}+f_{\mathcal{I{\scriptscriptstyle +}F}}+f_{\mathcal{I{\scriptscriptstyle +}S}}+f_{\mathcal{I}}$
) allows to enhance the results even further, bringing $2.7$ percent point gain ($62.0$ vs. $64.7$ global mean). Excluding the models which use semantic information ($f_{\mathcal{I{\scriptscriptstyle +}F{\scriptscriptstyle +}\mathcal{S}}}$ and $f_{\mathcal{I{\scriptscriptstyle +}S}}$) from the ensemble results only in a minor drop in the performance ($64.2$ vs. $64.7$ global mean). This shows that the competitive results can be achieved even with the system trained only with one pixel-level mask annotation per video, without employing extra annotations from Pascal VOC12.

Our lucid dreams enable automatic CRF-tuning (see Section \ref{sec:Effect-of-CRF}) which allows to further improve the results ($64.7\negmedspace\rightarrow\negmedspace 65.2$ global mean).
Employing the proposed temporal coherency step (see Section \ref{subsec:Architecture}) during inference brings an additional performance gain ($65.2\negmedspace\rightarrow\negmedspace66.6$ global mean).

\paragraph{Conclusion}

The results show that both flow and semantic priors provide a complementary
signal to RGB image only. Despite its simplicity our ensemble strategy
provides additional gain and leads to competitive results. Notice
that even without the semantic segmentation signal $\mathcal{\mathcal{S}}_{t}$
our ensemble result is competitive.

\begin{figure*}[t]
	\begin{centering}
		\includegraphics[width=0.97\textwidth]{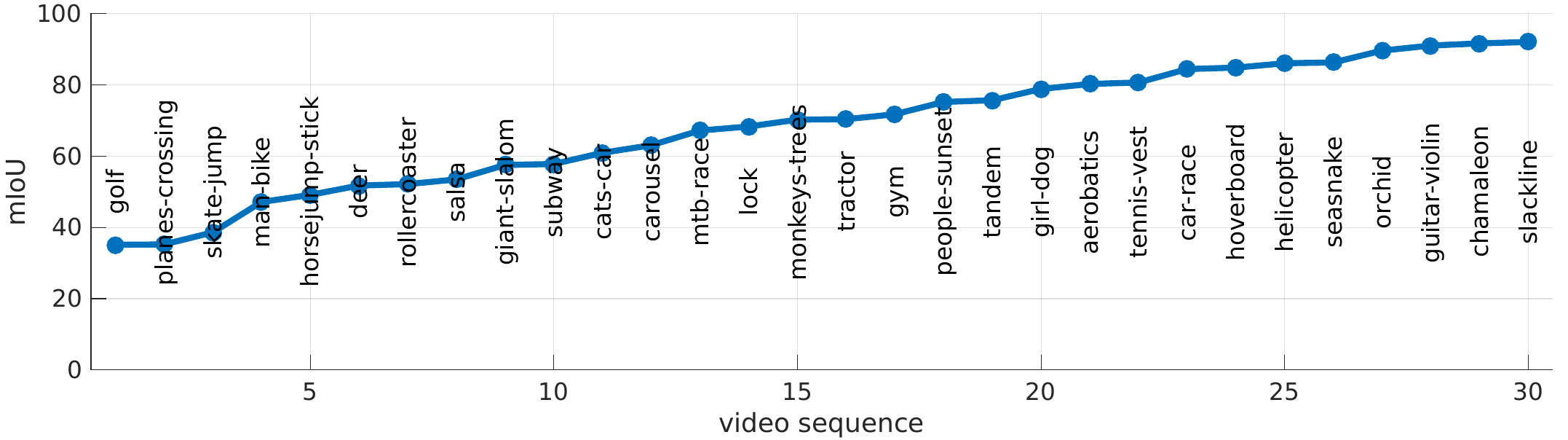}
		\par\end{centering}
	\caption{\label{fig:Per-sequence-results-davis17}Per-sequence results on $\text{DAVIS}_{\text{17}}$,
		test-dev set.}
\end{figure*}

\begin{figure*}
	\begin{centering}
		\setlength{\tabcolsep}{0em} 
		\renewcommand{\arraystretch}{0}
		\par\end{centering}
	\begin{centering}
		\hfill{}%
		\begin{tabular}{cccccc}
			{\footnotesize{}\hspace{-1em}}\includegraphics[width=0.15\textwidth]{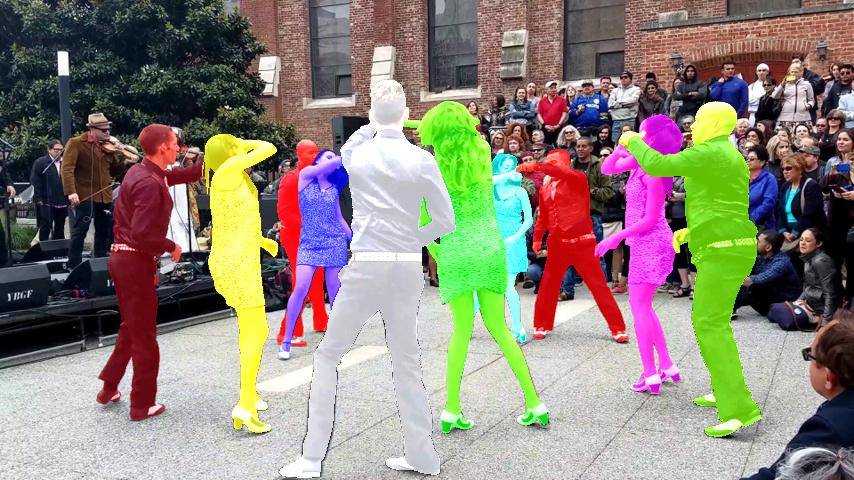} & {\footnotesize{}\hspace{-1em}}\includegraphics[width=0.15\textwidth]{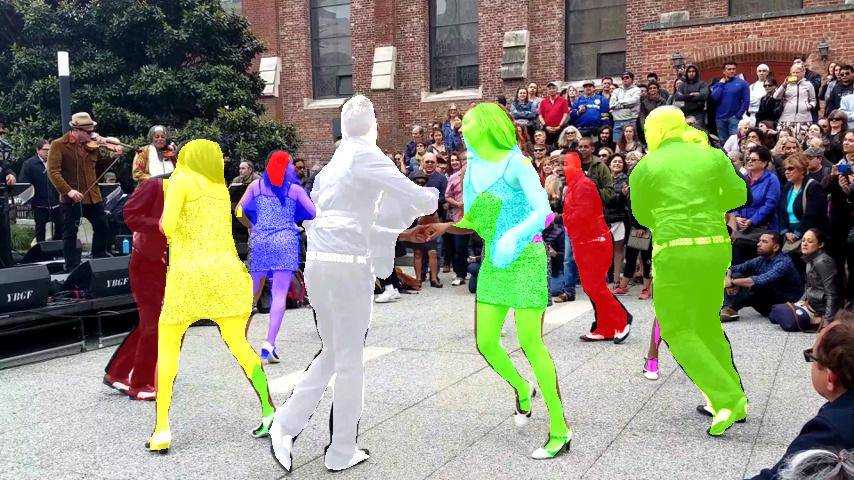} & {\footnotesize{}\hspace{-1em}}\includegraphics[width=0.15\textwidth]{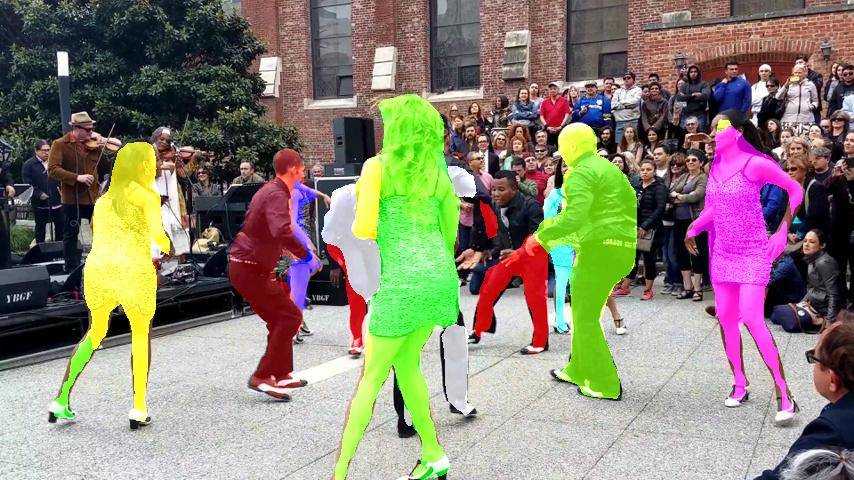} & {\footnotesize{}\hspace{-1em}}\includegraphics[width=0.15\textwidth]{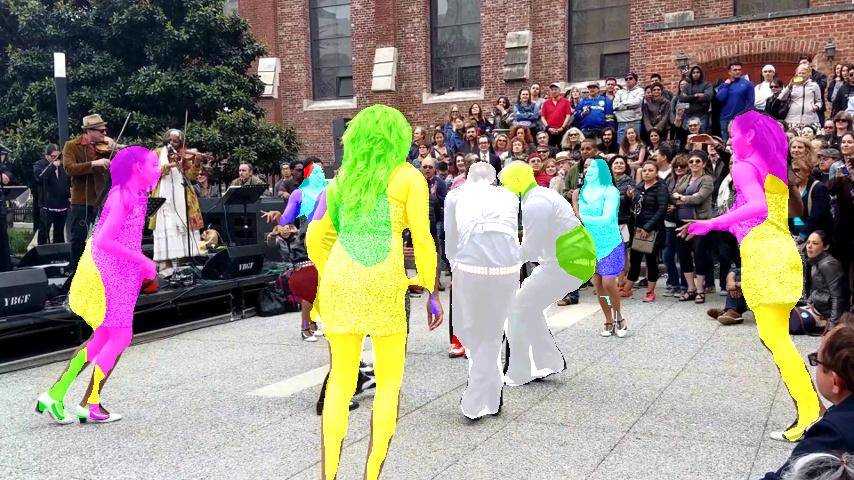} & {\footnotesize{}\hspace{-1em}}\includegraphics[width=0.15\textwidth]{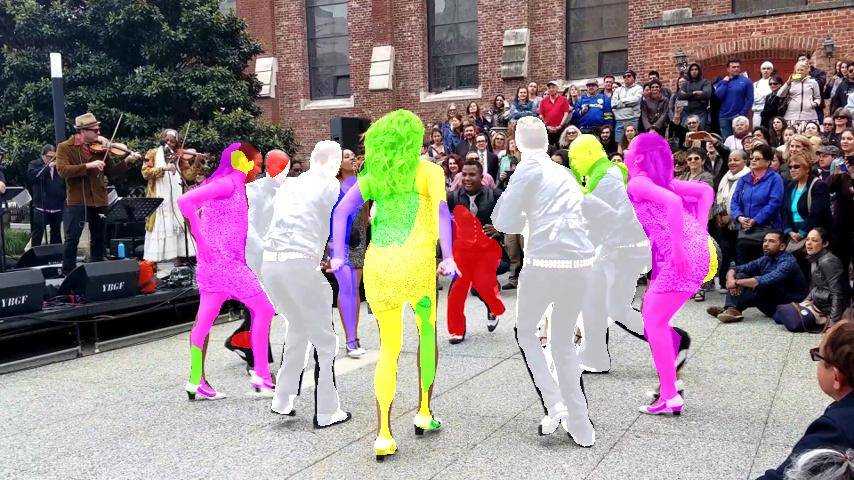} & {\footnotesize{}\hspace{-1em}}\includegraphics[width=0.15\textwidth]{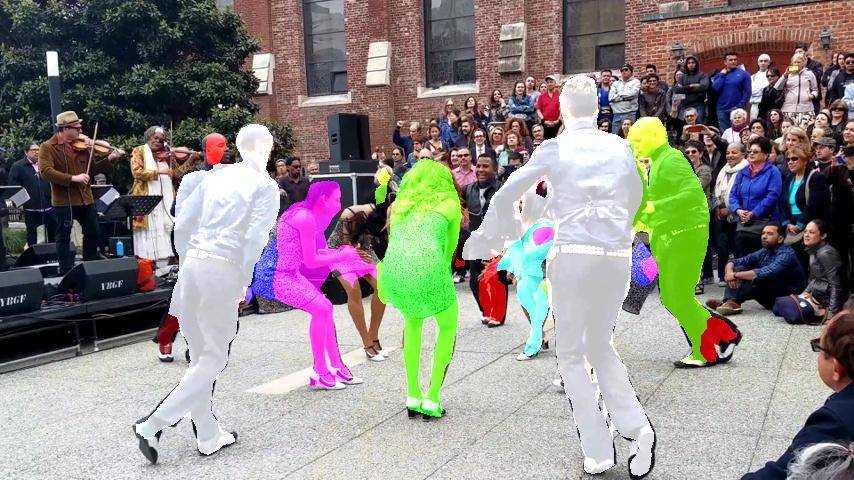}\tabularnewline
			{\footnotesize{}\hspace{-1em}}\includegraphics[width=0.15\textwidth]{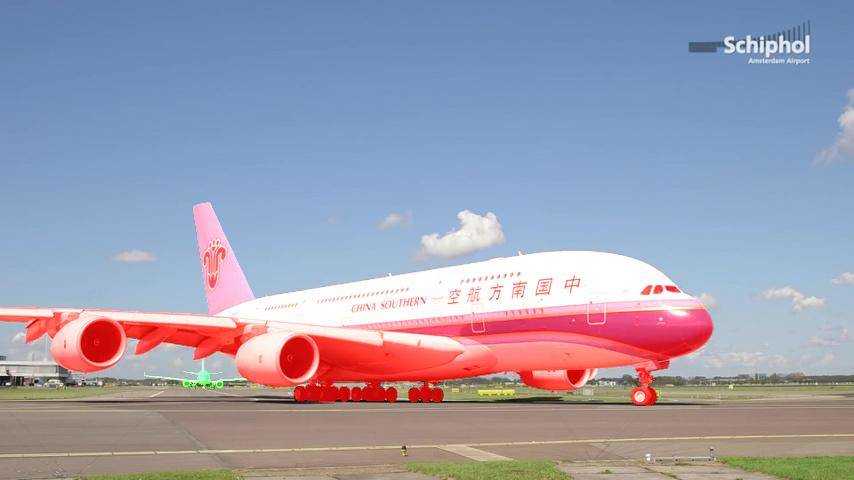} & {\footnotesize{}\hspace{-1em}}\includegraphics[width=0.15\textwidth]{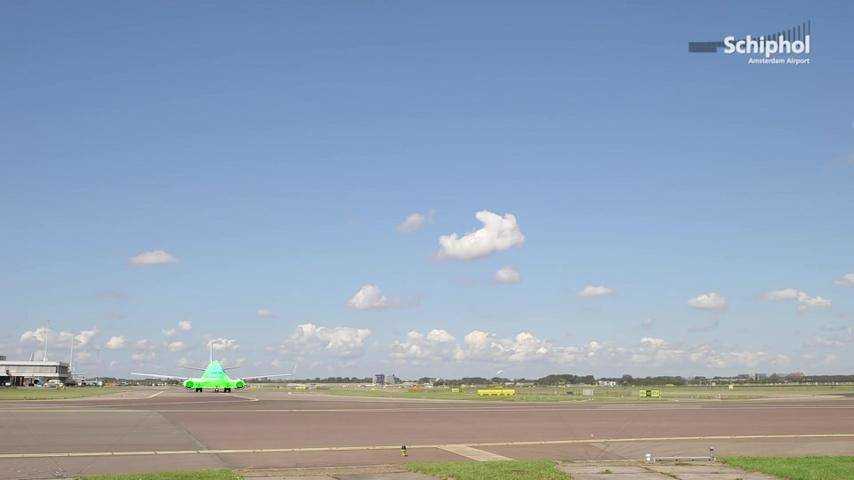} & {\footnotesize{}\hspace{-1em}}\includegraphics[width=0.15\textwidth]{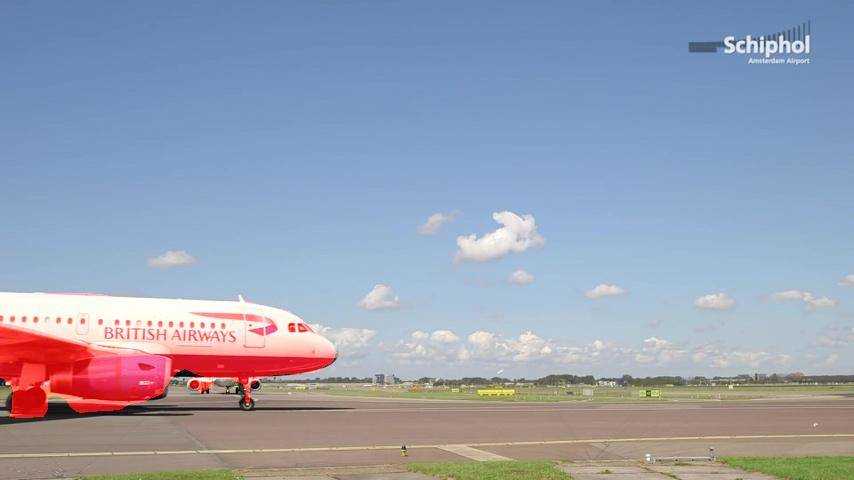} & {\footnotesize{}\hspace{-1em}}\includegraphics[width=0.15\textwidth]{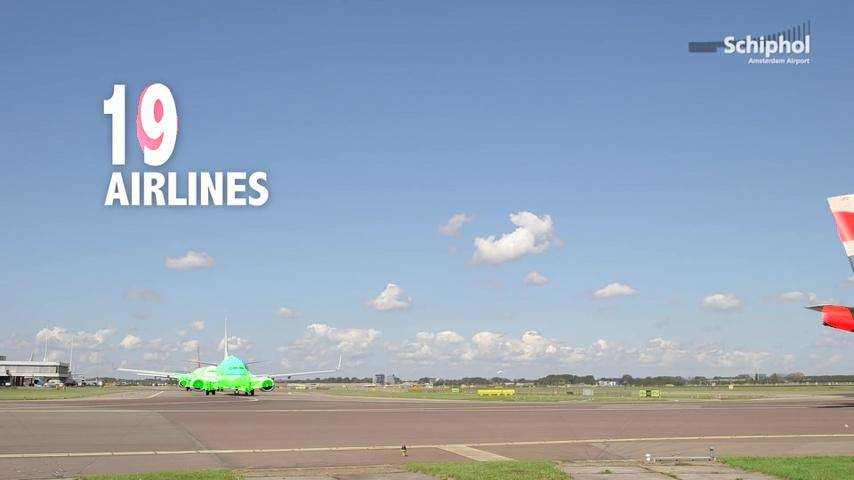} & {\footnotesize{}\hspace{-1em}}\includegraphics[width=0.15\textwidth]{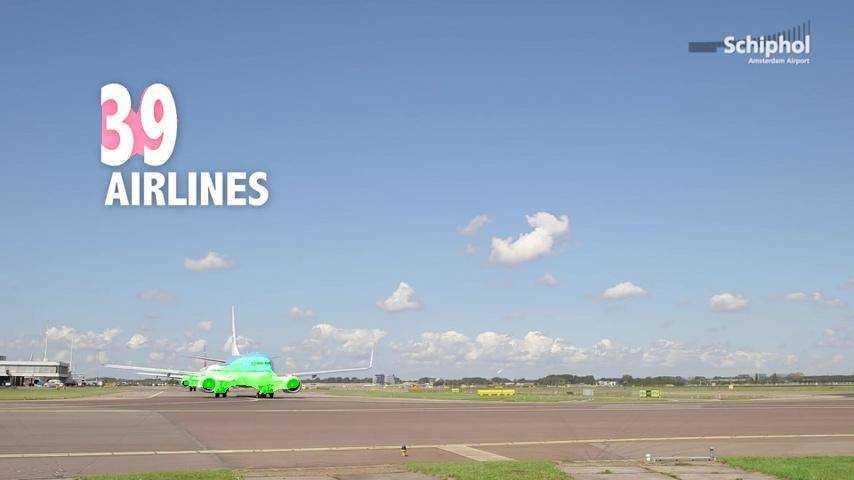} & {\footnotesize{}\hspace{-1em}}\includegraphics[width=0.15\textwidth]{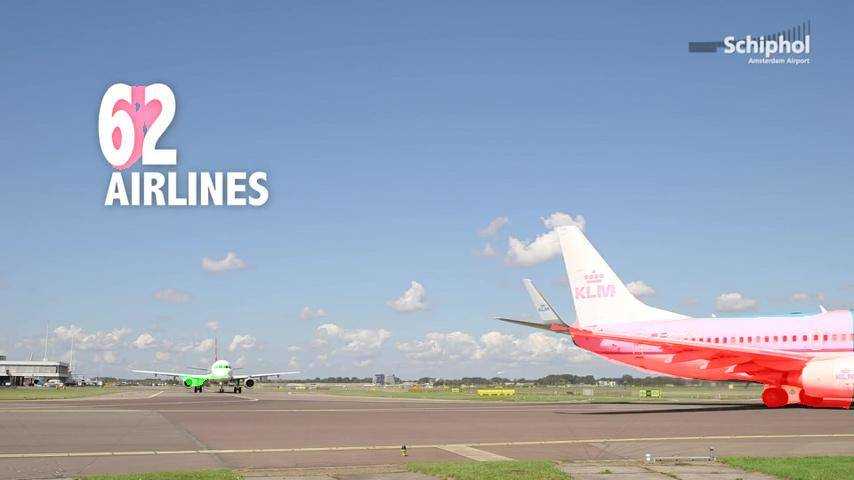}\tabularnewline
			{\footnotesize{}\hspace{-1em}1st frame, GT segment} & {\footnotesize{}\hspace{-1em}$20\%$} & {\footnotesize{}\hspace{-1em}$40\%$} & {\footnotesize{}\hspace{-1em}$60\%$} & {\footnotesize{}\hspace{-1em}$80\%$} & {\footnotesize{}\hspace{-1em}$100\%$}\tabularnewline
		\end{tabular}\hfill{}
		\par\end{centering}
	\medskip{}
	
	\caption{\label{fig:fail_cases-mult}LucidTracker failure cases on $\text{DAVIS}_{\text{17}}$,
		test-dev set. Frames sampled along the video duration (e.g. $50\%$:
		video middle point). We show 2 results mIoU over the video below 50. }
\end{figure*}

\subsection{\label{subsec:Error-analysis-mult}Error analysis}

We present the per-sequence results of $\mathtt{Lucid}$\-$\mathtt{Tracker}$
on $\text{DAVIS}_{\text{17}}$ in Figure \ref{fig:Per-sequence-results-davis17}
(per frame results not available from evaluation server). We observe
that this dataset is significantly more challenging than $\text{DAVIS}_{\text{16}}$
(compare to Figure \ref{fig:Per-sequence-results-davis16}), with
only $\nicefrac{1}{3}$ of the test videos above $80$ mIoU. This
shows that multiple object segmentation is a much more challenging task
than segmenting a single object.

The failure cases discussed in Section \ref{subsec:Error-analysis}
still apply to the multiple objects case. Additionally, on $\text{DAVIS}_{\text{17}}$
we observe a clear failure case when segmenting similar looking object
instances, where the object appearance is not discriminative to correctly
track the object, resulting in label switches or bleeding of the label
to other look-alike objects. Figure \ref{fig:fail_cases-mult} illustrates
this case. This issue could be mitigated by using object level instance
identification modules, like \cite{DAVIS2017-1st}, or by changing
the training loss of the model to more severely penalize identity
switches.

\paragraph{Conclusion}

In the multiple object case the $\mathtt{Lucid}$\-$\mathtt{Tracker}$
results remain robust ac\-ross different videos. The overall results
being lower than for the single object segmentation case, there is more
room for future improvement in the multiple object pixel-level segmentation
task.

\section{\label{sec:Conclusion}Conclusion}

We have described a new convnet-based approach for pixel-level object
segmentation in videos. In contrast to previous work, we show that top
results for single and multiple object segmentation can be achieved without
requiring external training datasets (neither annotated images nor
videos). Even more, our experiments indicate that it is not always
beneficial to use additional training data, synthesizing training
samples close to the test domain is more effective than adding more
training samples from related domains.

Our extensive analysis decomposed the ingredients that contribute
to our improved results, indicating that our new training strategy
and the way we leverage additional cues such as semantic and motion
priors are key.

Showing that training a convnet for video object segmentation can be done with
only few ($\sim\negmedspace100$) training samples changes the mindset
regarding how much general knowledge about objects is required to
approach this problem \cite{Khoreva2017CvprMaskTrack,Jain2017ArxivFusionSeg},
and more broadly how much training data is required to train large
convnets depending on the task at hand. 

We hope these new results will fuel the ongoing evolution of convnet
techniques for single and multiple object segmentation in videos.

\section*{Acknowledgements}

Eddy Ilg and Thomas Brox acknowledge funding by the DFG Grant BR 3815/7-1.

\FloatBarrier

{\small{}\bibliographystyle{ieee}
\bibliography{2017_iccv_segment_tracking}
}{\small \par}

\clearpage{}

\end{document}